\documentclass[letterpaper]{article} 
\usepackage{aaai2026}
\usepackage{times}  
\usepackage{helvet}  
\usepackage{courier}  
\usepackage[hyphens]{url}  
\usepackage{graphicx} 
\usepackage{natbib}  
\usepackage{caption} 
\usepackage{algorithm}
\usepackage{algorithmic}

\usepackage{xspace}
\newcommand{\ours}{FinMMDocR\xspace}

\usepackage{amsmath,amsfonts,bm}

\def\eqref#1{equation~\ref{#1}}

\def\1{\bm{1}}

\DeclareMathAlphabet{\mathsfit}{\encodingdefault}{\sfdefault}{m}{sl}
\SetMathAlphabet{\mathsfit}{bold}{\encodingdefault}{\sfdefault}{bx}{n}

\usepackage{subfigure}
\usepackage{xcolor} 
\usepackage{colortbl}
\usepackage{amsfonts} 
\usepackage{amssymb} 
\usepackage{pifont}
\usepackage{tabularx} 
\usepackage{arydshln} 
\usepackage{soul}
\usepackage{mathtools} 
\usepackage{courier}
\usepackage{adjustbox} 
\usepackage{multirow} 
\usepackage{paralist} 
\usepackage{booktabs} 
\usepackage{url}
\usepackage{enumitem}
\usepackage{tcolorbox}
\usepackage{tablefootnote}
\usepackage{bm}
\usepackage[textwidth=0.95in]{todonotes}
\definecolor{newblue}{RGB}{215,238,249}
\definecolor{my_green}{RGB}{40,154,121}
\definecolor{my_yellow}{RGB}{255,165,0}
\definecolor{my_red}{RGB}{176,46,46}
 
\newcommand{\errormark}{\textcolor{my_red}{\ding{56}}}
\newcommand{\eg}{\hbox{\emph{e.g.,}}\xspace}
\newcommand{\ie}{\hbox{\emph{i.e.,}}\xspace}

\usepackage{newfloat}
\usepackage{listings}
\DeclareCaptionStyle{ruled}{labelfont=normalfont,labelsep=colon,strut=off} 
\floatstyle{ruled}
\newfloat{listing}{tb}{lst}{}
\floatname{listing}{Listing}
\usepackage{makecell}
\usepackage{multirow}
\usepackage{colortbl}
\usepackage{dingbat}
\usepackage{xcolor}
\usepackage[table]{xcolor}
\usepackage{booktabs}
\usepackage{array}
\usepackage{pifont}
\usepackage{subcaption}
\usepackage{pdfpages}

\definecolor{darkgreen}{rgb}{0,0.5,0}

\title{\ours: Benchmarking Financial Multimodal Reasoning with\\Scenario Awareness, Document Understanding, and Multi-Step Computation}

\author{
Zichen Tang\textsuperscript{\rm 1},  
Haihong E\textsuperscript{\rm 1}\thanks{Corresponding author.},
\bf{Rongjin Li}\textsuperscript{\rm 1},
\bf{Jiacheng Liu}\textsuperscript{\rm 1},
\bf{Linwei Jia}\textsuperscript{\rm 1}, 
\bf{Zhuodi Hao}\textsuperscript{\rm 1},\\
\bf{Zhongjun Yang}\textsuperscript{\rm 1},
\bf{Yuanze Li}\textsuperscript{\rm 1},
\bf{Haolin Tian}\textsuperscript{\rm 1},
\bf{Xinyi Hu}\textsuperscript{\rm 1}, 
\bf{Peizhi Zhao}\textsuperscript{\rm 1},
\bf{Yuan Liu}\textsuperscript{\rm 1},\\
\bf{Zhengyu Wang}\textsuperscript{\rm 1}, 
\bf{Xianghe Wang}\textsuperscript{\rm 1},
\bf{Yiling Huang}\textsuperscript{\rm 1},
\bf{Xueyuan Lin}\textsuperscript{\rm 2},
\bf{Ruofei Bai}\textsuperscript{\rm 1},\\
\bf{Zijian Xie}\textsuperscript{\rm 1},
\bf{Qian Huang}\textsuperscript{\rm 1},
\bf{Ruining Cao}\textsuperscript{\rm 1},
\bf{Haocheng Gao\textsuperscript{\rm 1}}
}

\affiliations{
    \textsuperscript{\rm 1}Beijing University of Posts and Telecommunications\\
    \textsuperscript{\rm 2}Hithink RoyalFlush Information Network Co., Ltd.
    
}

\usepackage{bibentry}

\begin{document}

\definecolor{wkblue}{RGB}{210, 230, 250}
\definecolor{wkgreen}{RGB}{226,240,217}
\newcommand{\second}[1]{\cellcolor{wkblue}\underline{#1}}
\newcommand{\best}[1]{\cellcolor{wkgreen}\textbf{#1}}

\maketitle


\begin{abstract}
We introduce \textbf{FinMMDocR}, a novel bilingual multimodal benchmark for evaluating multimodal large language models (MLLMs) on real-world financial numerical reasoning. Compared to existing benchmarks, our work delivers three major advancements. (1) \textbf{Scenario Awareness}: 57.9\% of 1,200 expert-annotated problems incorporate 12 types of implicit financial scenarios (\eg{ Portfolio Management}), challenging models to perform expert-level reasoning based on assumptions; (2) \textbf{Document Understanding}: 837 Chinese/English documents spanning 9 types (\eg{ Company Research}) average 50.8 pages with rich visual elements, significantly surpassing existing benchmarks in both breadth and depth of financial documents; (3) \textbf{Multi-Step Computation}: Problems demand 11-step reasoning on average (5.3 extraction + 5.7 calculation steps), with 65.0\% requiring cross-page evidence (2.4 pages average). The best-performing MLLM achieves only 58.0\% accuracy, and different retrieval-augmented generation (RAG) methods show significant performance variations on this task. We expect FinMMDocR to drive improvements in MLLMs and reasoning-enhanced methods on complex multimodal reasoning tasks in real-world scenarios.
\end{abstract}

\newcommand{\homepage}{\raisebox{-1.5pt}{\includegraphics[height=1.2em]{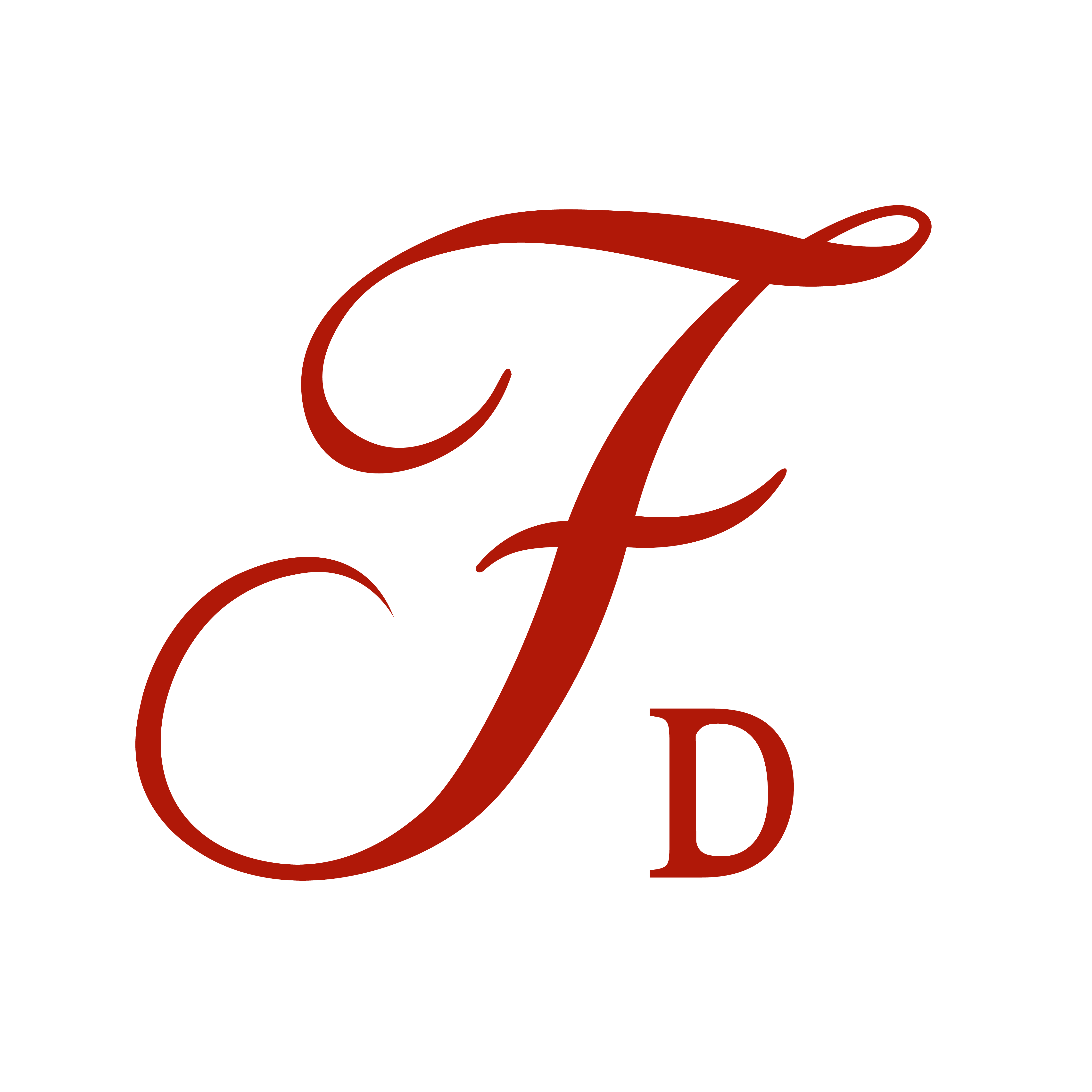}}}
\newcommand{\github}{\raisebox{-1.5pt}{\includegraphics[height=1em]{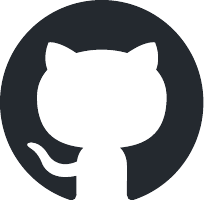}}}
\newcommand{\huggingface}{\raisebox{-1.5pt}{\includegraphics[height=1em]{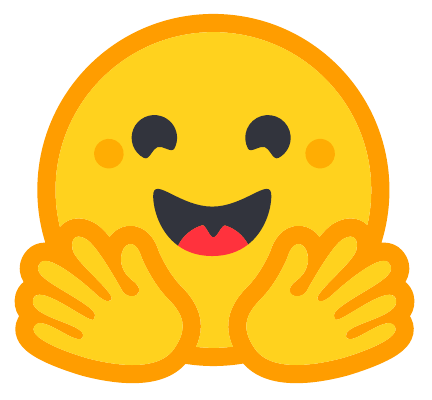}}}


\begin{links}
    \link{Project Resources}{https://bupt-reasoning-lab.github.io/FinMMDocR}
\end{links}

\section{Introduction}
\begin{table*}[htbp]
  \centering
  \small
  \setlength{\tabcolsep}{2pt}
  \begin{tabular}{l c cc ccc cccc}
    \toprule
    \multirow{2}{*}{\textbf{Benchmark}} &
    \multirow{2}{*}{\textbf{Modalities}} &
    \multicolumn{2}{c}{\textbf{Real-World Scenario}} &
    \multicolumn{3}{c}{\textbf{Visually-Rich Document}} &
    \multicolumn{4}{c}{\textbf{Multi-Step Computation}} \\
    \cmidrule(lr){3-4} \cmidrule(lr){5-7} \cmidrule(lr){8-11}
    & & Explicit (\%) & Implicit (\%) & \# Docs & \# Pages & \# Tokens (k) & Num. Rea. (\%) & \# Ext. & \# Cal. & Cross-Page (\%) \\
    \midrule
    \multicolumn{11}{l}{\textit{Financial QA}}\\
    CodeTAT-QA        & T & \errormark & \errormark & \errormark   & \errormark & \errormark & 100 & 2.1 & 1.0 & \errormark \\
    FinanceMath       & T &  47.5  &  39.0   &  \errormark   &  \errormark  &  \errormark & 100 & 3.3 & 2.5 &  \errormark \\
    FinanceReasoning  & T & 39.1  & 22.1   &  \errormark   &  \errormark  & \errormark & 100 & 2.9 & 2.2 &  \errormark \\
    MME-Finance       & T+I & \errormark & \errormark & \errormark &  \errormark & \errormark & 15 & 2.2 & 1.1 & \errormark \\
    FinMMR            & T+I & \errormark & \errormark &  \errormark  &  \errormark  & \errormark & 100 & 2.6 & 1.8 & \errormark \\
    DocMath-Eval$_{\text{CompLong}}$ & T+TD & 15.5 & 15.1 &  1,500  & 61.0 & 46.5 & 100 & 3.0 & 2.0 & 52.7 \\
    
    \midrule
    \multicolumn{11}{l}{\textit{Document QA}}\\
    SlideVQA          & T+MD & \errormark & \errormark & 2,619 &  20.0 &  2.0 & 35 &  $\leq$3 &  $\leq$3  & 13.9 \\
    MMLongBench-Doc   & T+MD & \errormark & \errormark &  135  &  47.5  & 21.2 & 6 &  $\leq$3  &  $\leq$3  & 33.7 \\
    LongDocURL        & T+MD &  \errormark & \errormark &  396 &  85.6   & 43.6 & 8 & 2.6 & 0.8 & 52.9 \\
    \midrule
    \textbf{FinMMDocR (ours)}    & T+MD & \textbf{33.7} & \textbf{57.9} &
                                   \textbf{837} & \textbf{50.8} & \textbf{38.8} & \textbf{100} &
                                   \textbf{5.3} & \textbf{5.7} & \textbf{65.0} \\
    \bottomrule
  \end{tabular}
  \caption{Comparison of FinMMDocR and related benchmarks.
		\textbf{T}: text; \textbf{I}: images; \textbf{TD}: text document; \textbf{MD}: multimodal document; \textbf{Explicit}: scenarios with directly given conditions; \textbf{Implicit}: scenarios requiring inferred assumptions; \textbf{Pages}: pages/doc; 
        \textbf{Tokens}: tokens/doc; \textbf{Num. Rea.}: numerical reasoning questions; \textbf{Ext.}: average extraction steps; \textbf{Cal.}: average calculation steps.}
  \label{tab:benchmark_comparison}
\end{table*}

Recently, multimodal large language models (MLLMs)~\citep{vit,qwen25vl} have advanced multimodal reasoning, excelling in visual commonsense reasoning~\citep{vcr,yu2024mmvet} and visual question answering~\citep{vqa,textvqa} end-to-end. Large multimodal reasoning models (LMRMs)~\citep{o4mini}, enhanced via reinforcement learning, show promise for complex real-world tasks. They demonstrate superior visual understanding and expert-level reasoning capabilities in domain-specific tasks, operating human-like~\citep{prtp}.

\begin{figure}[t]
	\centering
	\includegraphics[width=\columnwidth]{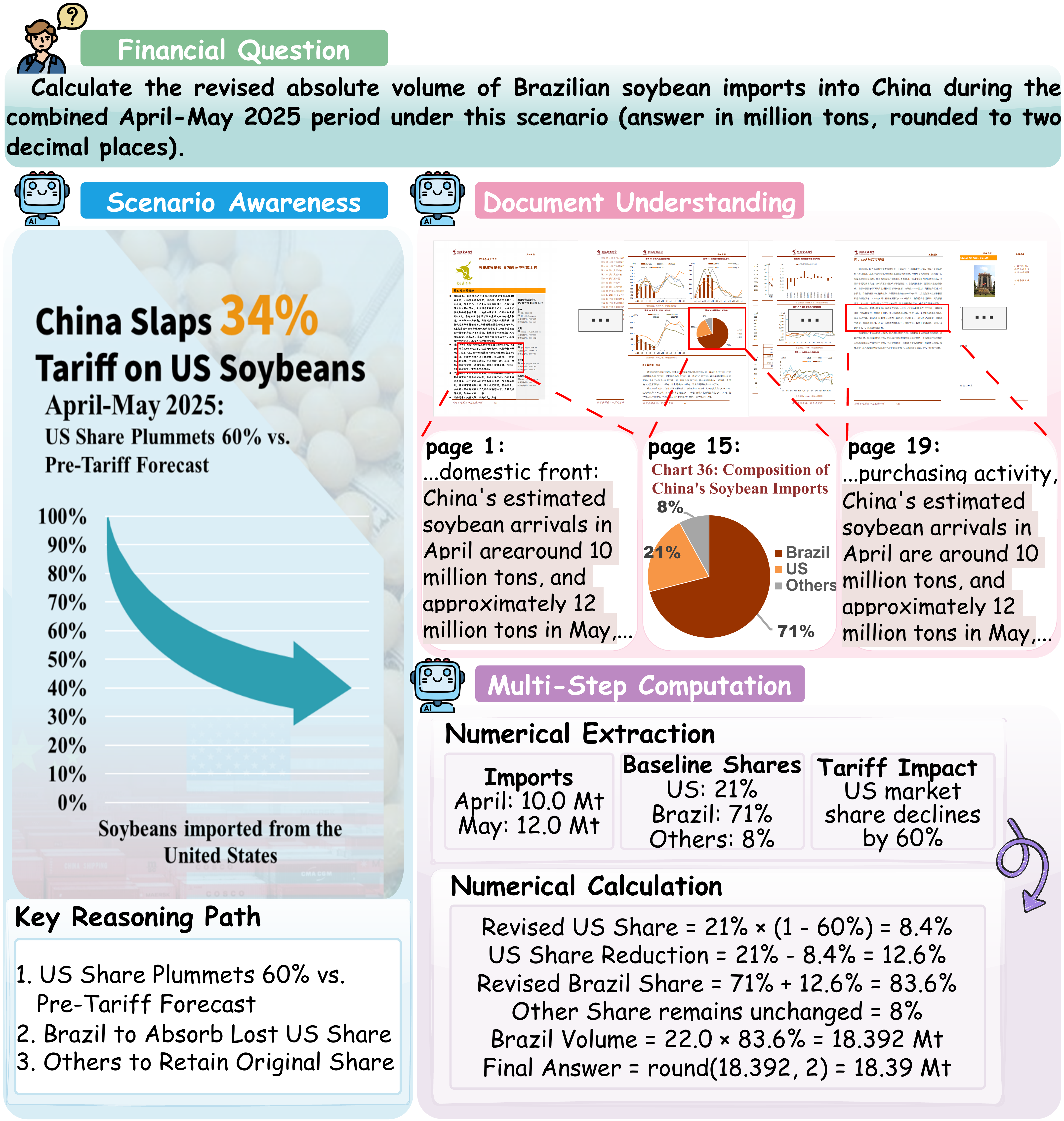} 

    \caption{An example of \ours, including a real-world scenario, a visually-rich document and a multi-step numerical reasoning question, demanding models to reason about China’s import volume shifts for Brazil vs. US soybeans based on evolving US-China tariff conflicts.} 
	\label{fig:first_example}
\end{figure}
Despite LMRMs' success, current domain-specific reasoning benchmarks remain confined to STEM disciplines~\citep{mathvista,math-vision}, often using abstract exam-style questions. They inadequately model the real-world tasks that experts routinely handle. As shown in Figure~\ref{fig:first_example}, financial analysts must integrate contextual knowledge to formulate necessary assumptions, then process visually dense financial documents to extract key information. This is followed by comprehensive analytical reasoning, often involving precise multi-step computations, to support high-stakes decision-making. Table~\ref{tab:benchmark_comparison} shows existing financial QA and document QA benchmarks' key limitations compared to such complex multimodal reasoning scenarios:

\setlist[itemize]{itemsep=0.1em, topsep=0.2em}
\begin{itemize}
    \item \textbf{Absence of Real-World Financial Scenario} 
    \emph{Financial analysts must analyze real-time financial environments to make professional judgments and plausible assumptions.} However, traditional benchmarks~\citep{bizbench,mme-finance,slidevqa,mmlongbench-doc,deng-etal-2025-longdocurl} only extract explicitly stated information.

    \item \textbf{Deficiency in Multimodal Document Understanding} 
    \emph{Financial analysts rely on extensive professional documents to extract key information and diverse indicators.} Some benchmarks~\citep{bizbench,financemath,financereasoning} use text-only inputs, while multimodal ones~\citep{finmme,mme-finance} contain sparse isolated charts or tables. Long-document benchmarks~\citep{mmlongbench-doc,deng-etal-2025-longdocurl} lack diverse financial documents and numerical reasoning tasks.

    \item \textbf{Neglect of Precise Multi-Step Computation} 
    \emph{Financial decision-making, unlike qualitative analysis, requires exact multi-step computations.} In this high-stakes domain~\citep{bizbench}, models must deliver numerically exact answers under strict criteria. Prior benchmarks~\citep{financemath,bizbench} ignore units, percentages, and decimals or allow 1.0\% error margins, diverging from real-world needs.
\end{itemize}

To fill this gap, we construct FinMMDocR, a more challenging and realistic financial multimodal reasoning benchmark featuring contextual awareness, document understanding, and multi-step computation. \ours consists of 1,200 numerical reasoning questions (1:1 Chinese-English), equipped with real-world scenarios, visually-rich financial documents, detailed evidence page annotations, golden Python solutions for problem-solving, and exact answers.

\setlist[itemize]{itemsep=0.1em, topsep=0.2em}
\begin{itemize}
    \item \textbf{Scenario Awareness} 57.9\% of questions incorporate carefully designed implicit financial scenarios from 12 categories (\eg{ Portfolio Management}), with an average of 1.9 scenarios per question, significantly surpassing existing datasets in density, richness, and complexity.

    \item \textbf{Document Understanding} FinMMDocR contains 837 financial long-documents covering 9 bilingual (Chinese/English) categories (\eg{ Financial Engineering, Futures \& Options}). These documents feature high information density (50.8 pages/doc and 38.8k tokens/doc) and professional visual elements (\eg{ candlestick charts}).

    \item \textbf{Multi-Step Computation} FinMMDocR averages 11 reasoning steps (5.3 extraction, 5.7 calculation), surpassing other financial reasoning tasks. It enforces strict evaluation (units, percentages, decimals) with 0.2\% error tolerance, matching real-world needs. 65.0\% of questions require cross-page reasoning (2.4 evidence pages each).
\end{itemize}

\begin{figure*}[t]
	\centering
	\includegraphics[width=0.94\textwidth]{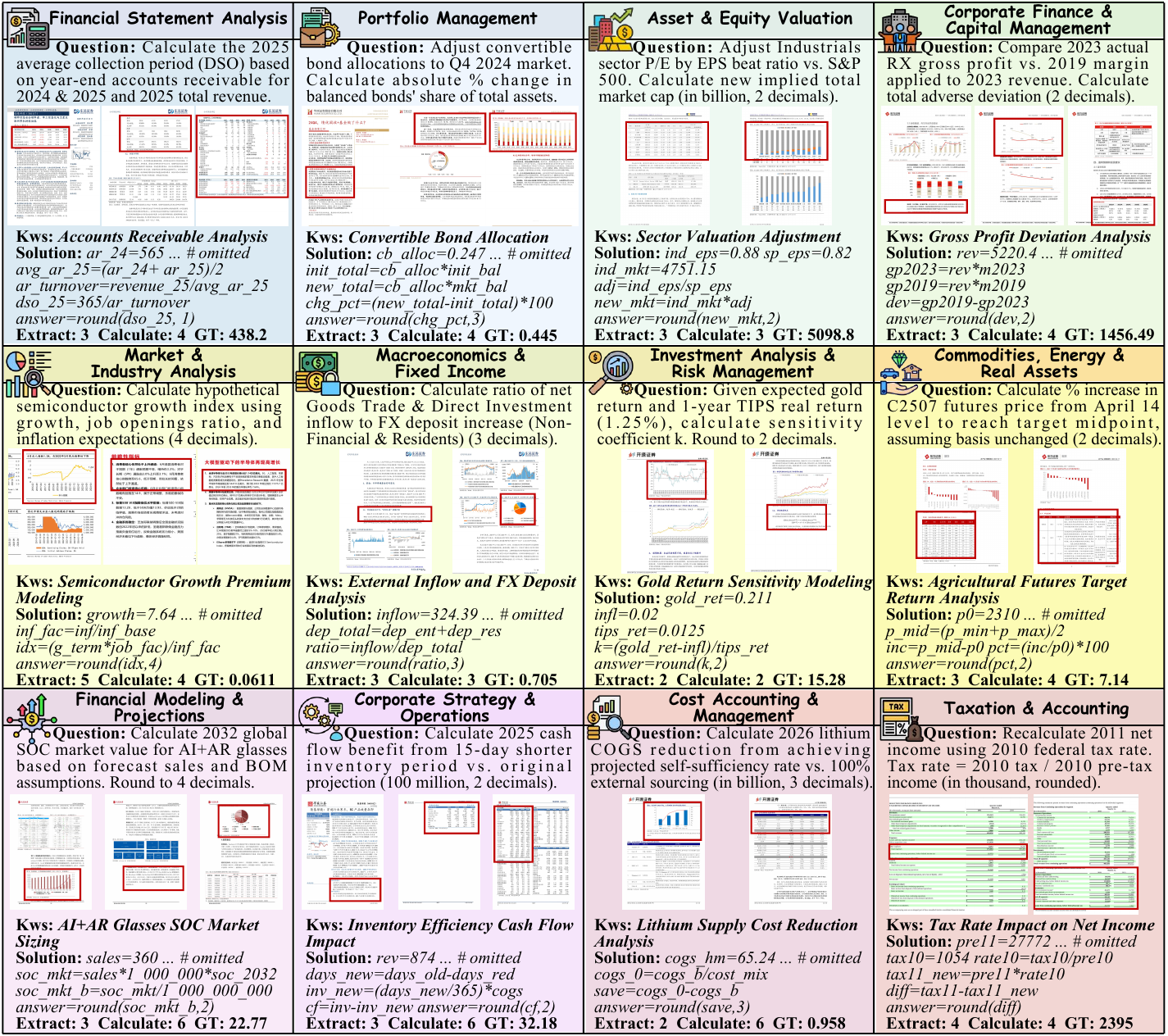}
	\caption{
		12 financial scenarios with \ours examples, covering 9 document categories and cross-page computations. Requires expert \emph{scenario awareness}, \emph{document understanding}, and \emph{multi-step computation}. \textbf{Kws}: keywords, \textbf{GT}: ground truth.}
	\label{fig:overall}
\end{figure*}

We evaluate 11 proprietary and open-source MLLMs with image inputs using Program-of-Thought (PoT)~\citep{pot}, along with 15 LLMs with text inputs using OCR. Beyond end-to-end reasoning, we also evaluate 6 embedding models and 5 agentic retrieval-augmented generation (Agentic RAG) frameworks~\citep{agenticrag}. The experimental results reveal three key findings:

\setlist[itemize]{itemsep=0.1em, topsep=0.2em}
\begin{itemize}
    \item \textbf{MLLMs Are Not Qualified Financial Experts for Multimodal Numerical Reasoning.} No model exceeds 60.0\% accuracy (OpenAI o4-mini-high: 58.0\%), with open-source models particularly struggling, while reasoning-enhanced models show consistent advantages.
    \item \textbf{The More Complex the Task, the Worse Models Perform.} Multimodal models show accuracy degradation in multi-scenario tasks and document understanding failures (78.0\% of errors), with extraction errors being the main bottleneck in PoT settings.
    \item \textbf{Vision Is Stronger Than Text, But Complex Agents Underperform Simple RAG.} Vision RAGs surpass text-only methods by utilizing critical document visual cues, yet longer pipelines introduce error propagation that degrades performance, while iterative Agentic RAGs suffer from prohibitive latency without corresponding accuracy improvements for practical deployment.
\end{itemize}

\section{Benchmark Construction}

\begin{figure*}[t]
	\centering
	\includegraphics[width=1\textwidth]{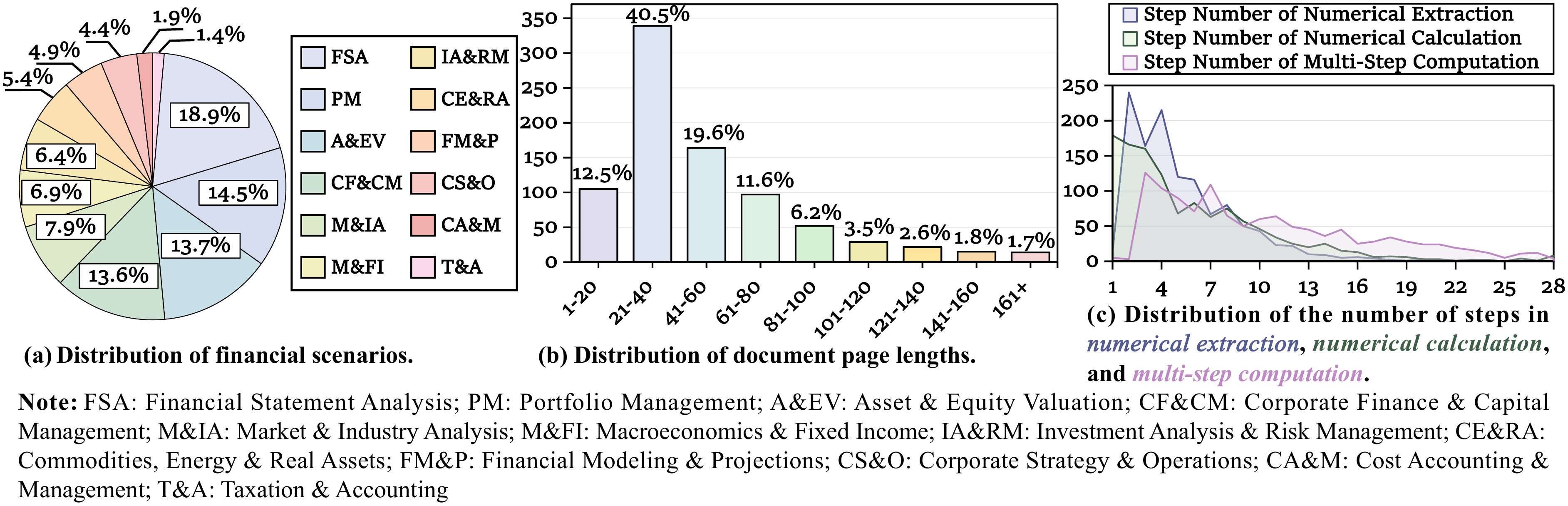}
	\caption{Distribution of \ours: financial scenarios, document page lengths, and reasoning steps per question.}
	\label{fig:distribution}
\end{figure*}

\begin{figure}[t]
	\centering
	\includegraphics[width=0.8\linewidth]{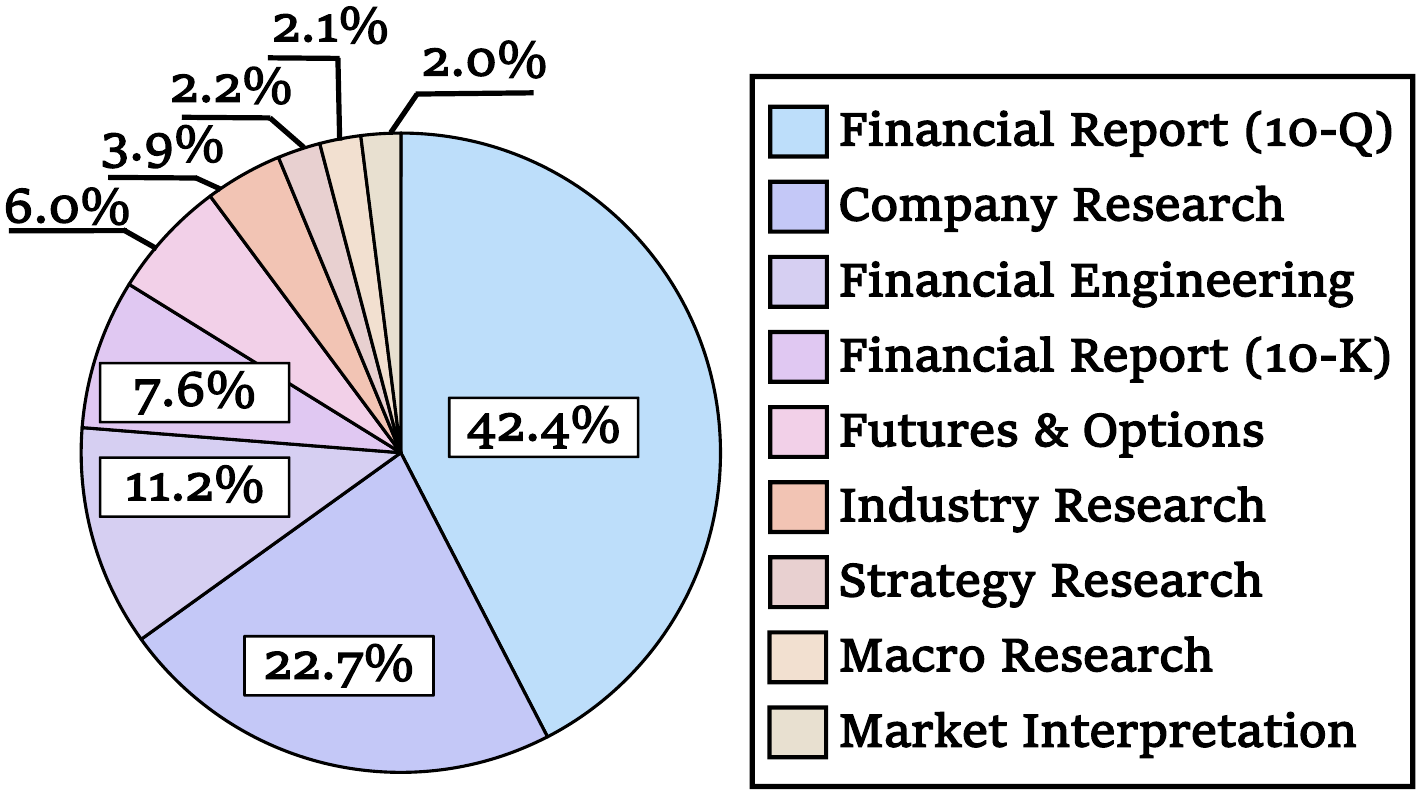}
	\caption{Distribution of \ours: financial document categories.}
	\label{fig:document_category}
\end{figure}

\subsection{Overview of FinMMDocR}
We introduce FinMMDocR, designed to evaluate the capability of MLLMs to perform complex numerical reasoning when presented with real-world financial scenarios and visually-rich financial documents. Following~\citep{docmath-eval}, each question is accompanied by a Python solution, a standard answer, and page numbers that indicate the locations of relevant visual elements. More examples are shown in Appendix A.



\subsection{Data Curation Process}
\paragraph{Updates to Public Dataset} We selected and re-annotated 600 English questions from the DocMath-Eval$_{\text{CompLong}}$ ~\citep{docmath-eval}, comprising all 300 samples from the \emph{testmini} subset and an additional 300 samples chosen from the \emph{test} subset based on diversity and complexity. For the latter, we manually completed previously unreleased solution programs, standard answers, and evidence pages. We retrieved the corresponding documents for all selected examples, rendered each page as an image, and removed original textual inputs to ensure a real multimodal reasoning setting.
\paragraph{Building a Novel Dataset from Scratch} We additionally created 600 entirely new Chinese questions. Specifically, we collected 385 Chinese research reports, acquired through authorized channels, covering diverse financial topics (\eg{ Company Research, Industry Research}). We manually constructed realistic financial scenarios based on document contents (\eg{ Financial Modeling \& Projections}), and further generated knowledge-intensive problems involving complex numerical reasoning along with corresponding Python solutions, with the assistance of two advanced MLLMs~\citep{gemini2.5propreview, claude3.7sonnet}. Documents included in \ours are exceptionally long, and problems require extracting information dispersed across various sections and modalities (\eg{ text, tables, and charts}). 

\paragraph{Data Quality Assurance} Our annotation team comprised 15 master's students majoring in finance and two CFA-certified experts. We implemented a rigorous annotation process to ensure benchmark quality. Specifically, we first fed each sample along with its multimodal document into Gemini 2.5 Pro Preview~\citep{gemini2.5propreview} and Claude 3.7 Sonnet~\citep{claude3.7sonnet}, the highest-performing MLLMs, to obtain two candidate annotations. Since the model's initial outputs contained numerous logical errors, calculation mistakes, and hallucinations, two annotators cross-reviewed the candidate annotations, selected one for adoption, and subsequently refined it.  In cases of disagreement, an additional expert was brought in for arbitration. The selected results underwent further verification and annotation by two annotators. From the initially generated 759 samples, 159 were discarded. Of the remaining 600 samples, 494 underwent modifications: 451 required evidence revision, 80 needed solution adjustment, and 36 had question reformulation. Details are provided in Appendix C.




\section{Benchmark Analysis}

\begin{table}[t]
	\centering
    \small
	\begin{adjustbox}{width=0.41\textwidth}
	\begin{tabular}{lc}
		\toprule
		\textbf{Property} & \textbf{Value}  \\
		\midrule
		\# Total Samples & 1,200 \\
		\# Total Document & 837 \\
		\midrule
		\# Financial Scenario (Avg.) & 1.9 \\
		\# Evidence Page (Avg.) & 2.4 \\
		\# Textual Extraction Step (Avg.) & 1.0 \\
	  \# Visual Extraction Step (Avg.) & 4.3 \\
        \# Extraction Step (Textual and Visual) (Avg.) & 5.3 \\
		\# Calculation Step (Avg.) & 5.7 \\
        \# Computation Step (Ext. and Cal.) (Avg.) & 11.0 \\
		
		\bottomrule
	\end{tabular}%
	\end{adjustbox}
	
	\caption{Basic statistics of \ours.
		}
	\label{tab:statistics}
\end{table}
Table~\ref{tab:statistics} shows \ours contains 1,200 samples evaluating MLLMs' capabilities across three key dimensions.

\noindent \textbf{Scenario Awareness} \emph{\ours introduces financial reasoning problems with unprecedented scenario density and depth.} 66.2\% of problems are scenario-driven across 12 categories (Figure~\ref{fig:distribution}(a)). Additionally, all problems feature 1.9 mixed scenarios on average, with 57.9\% requiring implicit scenario assumptions rather than given conditions.

\noindent \textbf{Document Understanding} \emph{Tasks in \ours require synthesizing information from multimodal domain-specific documents.} As shown in Figure~\ref{fig:distribution}(b) and Figure~\ref{fig:document_category}, 837 bilingual (Chinese/English) documents cover 9 categories, averaging 50.8 pages each with 2.4 evidence pages per task, and contain professional charts demanding domain expertise.

\noindent \textbf{Multi-Step Computation} \emph{FinMMDocR provides complex financial reasoning tasks requiring cross-page, multimodal, and multi-step reasoning.} As shown in Figure~\ref{fig:distribution}(c), each problem requires 11 sequential reasoning steps on average: 5.3 for multimodal numerical extraction (1.0 textual, 4.3 visual) and 5.7 for financial calculation synthesis.

Compared to prior financial QA and document QA benchmarks, FinMMDocR eliminates explicit conditions, limited modalities/types, and excessive focus on information extraction/logical reasoning, better evaluating MLLMs' complex numerical reasoning capabilities in real-world settings.


\begin{table*}[!ht]
\centering
\small
\setlength{\tabcolsep}{0pt}
{%
\renewcommand{\arraystretch}{1.0}

\begin{tabular*}{\textwidth}{@{\extracolsep{\fill}}lccccccccccc@{}}
\toprule

\multirow{2}{*}{\textbf{Model}} &
\multirow{2}{*}{\textbf{Size}} &
\multirow{2}{*}{\textbf{ACC}} &
\multirow{2}{*}{\textbf{Input \textit{Cfg.}}} &
\multicolumn{2}{c}{\textbf{Scenario}} &
\multicolumn{2}{c}{\textbf{\textit{Doc. Len.}}} &
\multicolumn{2}{c}{\textbf{Extract}} &
\multicolumn{2}{c}{\textbf{Compute}} \\

\cmidrule(lr){5-6} \cmidrule(lr){7-8} \cmidrule(lr){9-10} \cmidrule(lr){11-12}
& & & & w/ & w/o & $\le$30 & $\ge$31 & $\le$4 & $\ge$5 & $\le$4 & $\ge$5 \\
\midrule

\rowcolor{gray!10}
\multicolumn{12}{c}{\textbf{MLLM (Image Input)}} \\
\midrule
\multicolumn{12}{l}{\emph{Proprietary MLLMs}} \\
OpenAI o4-mini-high &  & \best{58.00} & 300@F & \best{55.72} & \best{62.34} & \best{57.02} & \best{58.95} & \best{63.92} & \best{51.50} & \best{63.36} & \best{52.05} \\
Doubao-1.5-thinking-pro &  & \second{38.17} & U@F & \second{39.50} & 35.41 & \second{43.99} & \second{32.51} & 40.35 & \second{35.93} & 39.15 & \second{37.25} \\
Claude 3.7 Sonnet (Thinking) &  & 37.00 & 50@1920 & 35.60 & \second{39.40} & 41.96 & 32.18 & \second{40.66} & 32.92 & \second{39.31} & 34.40 \\
Doubao-1.5-vision-pro &  & 29.25 & U@F & 28.81 & 30.17 & 32.99 & 25.62 & 32.91 & 25.13 & 31.92 & 26.20\\
Gemini 2.5 Pro Preview &  & 27.42 & 300@F & 27.92 & 26.43 & 26.40 & 28.41 & 32.91 & 21.24 & 31.45 & 22.82 \\
GPT-4o &  & 17.17 & 50@1920 & 12.20 & 27.18 & 13.54 & 20.69 & 26.42 & 6.90 & 25.79 & 7.49 \\
Grok 2 Vision &  & 2.17 & 15@1920 & 2.64 & 1.25 & 1.18 & 3.12 & 3.16 & 1.06 & 3.14 & 1.07 \\

\multicolumn{12}{l}{\emph{Open-source MLLMs}} \\
Qwen2.5-VL 72B & 72B & 12.92 & 50@F & 10.57 & 17.71 & 14.04 & 11.82 & 18.35 & 6.90 & 18.24 & 6.95\\
Llama 4 Maverick & 400A17B & 2.67 & 300@F & 3.65 & 0.75 & 1.86 & 3.45 & 3.96 & 1.24 & 4.09 & 1.07 \\
Mistral Small 3.1 & 24B & 1.08 & 15@3840 & 1.51 & 0.25 & 0.51 & 1.64 & 1.58 & 0.53 & 1.42 & 0.71 \\
Gemma 3 27B & 27B & 0.67 & 15@3840 & 1.01 & 0.00 & 0.17 & 1.15 & 0.95 & 0.35 & 0.94 & 0.36 \\
\midrule

\rowcolor{gray!10}
\multicolumn{12}{c}{\textbf{OCR\,+\,LLM (Text Input)}} \\
\midrule
\multicolumn{12}{l}{\emph{Proprietary LLMs}} \\
Gemini 2.5 Pro Preview &  & \best{53.83} & N & \best{55.22} & \best{51.12} & \best{56.01} & \best{51.72} & \best{56.80} & \best{50.62} & \best{54.09} & \best{53.65} \\
Claude 3.7 Sonnet (Thinking) &  & \second{48.58} & N & 48.68 & \second{48.38} & 50.42 & \second{46.80} & \second{51.90} & 44.96 & \second{49.69} & 47.42 \\
OpenAI o4-mini-high &  & 47.92 & 200k & \second{50.94} & 41.90 & \second{51.27} & 44.66 & 49.53 & \second{46.19} & 47.64 & \second{48.31} \\
Doubao-1.5-thinking-pro &  & 42.67 & 96k & 43.52 & 40.90 & 44.33 & 41.05 & 46.99 & 37.88 & 44.65 & 40.46 \\
Grok 3 &  & 41.00 & 128k & 40.13 & 42.64 & 41.29 & 40.72 & 44.62 & 36.99 & 43.87 & 37.79 \\
Doubao-1.5-vision-pro &  & 32.75 & 128k & 31.70 & 34.66 & 30.46 & 34.98 & 39.40 & 25.49 & 38.36 & 26.56 \\
GPT-4o &  & 22.17 & 128k & 19.25 & 28.18 & 20.14 & 24.14 & 28.96 & 14.69 & 28.93 & 14.62 \\

\multicolumn{12}{l}{\emph{Open-source LLMs}} \\
DeepSeek-R1 & 671A37B & 40.00 & 64k & 41.51 & 37.16 & 42.13 & 37.93 & 44.46 & 35.22 & 42.61 & 37.25 \\
DeepSeek-V3 & 671A37B & 32.67 & 128k & 30.57 & 36.66 & 30.46 & 34.81 & 40.03 & 24.42 & 39.47 & 24.96 \\
Llama 4 Maverick & 400A17B & 29.08 & N & 27.30 & 32.42 & 29.61 & 28.57 & 33.23 & 24.42 & 32.55 & 25.13 \\
Qwen3 & 235A22B & 25.08 & 128k & 21.26 & 32.67 & 22.00 & 28.08 & 34.18 & 15.04 & 33.33 & 15.86 \\
Mistral Small 3.1 & 24B & 15.83 & 128k & 12.45 & 22.44 & 14.72 & 16.91 & 21.68 & 9.38 & 22.33 & 8.56 \\
Qwen2.5-VL 72B & 72B & 15.00 & 128k & 12.96 & 18.95 & 16.75 & 13.30 & 19.62 & 9.91 & 19.81 & 9.63 \\
Llama 3.3 70B & 70B & 12.17 & 128k & 9.43 & 17.71 & 9.14 & 15.11 & 18.51 & 5.13 & 19.18 & 4.28 \\
Gemma 3 27B & 27B & 5.75 & 128k & 5.41 & 6.48 & 4.91 & 6.57 & 8.39 & 2.83 & 8.65 & 2.50 \\
\bottomrule
\end{tabular*}
}
\caption{Model performance across input configurations. \textbf{Size}: for MoE models, total params and total activated are divided by ``A''; \textbf{ACC}: accuracy; \textbf{Input \textit{Cfg.}}: \textbf{U@F} = unmerged at full resolution, \textbf{X@Y} = merge \textbf{X} images (\eg{ 300}), \textbf{Y} = long edge pixels (\eg{ 1920}), \textbf{N} = No cut-off; \textbf{Scenario}: \textbf{w/} = with contextual scenarios, \textbf{w/o} = without; \textbf{\textit{Doc. Len.}}: document length.}
\label{tab:main_result_revised}
\end{table*}

\section{Experiments}
\subsection{Experiments Setting}
\paragraph{Models} 
Following~\citep{mmlongbench-doc, deng-etal-2025-longdocurl}, we assessed the comprehension capabilities of MLLMs by feeding images directly into models and inputting text extracted by Tesseract OCR engine~\citep{tesseract}. We evaluated 26 different configurations (11 for image input, 15 for text input) on both proprietary and open-source models.

\paragraph{Input Paradigm}
We designed various configurations to accommodate differences across MLLMs. We tested merging 300, 50, or 15 pages into a single input, alongside an unmerged strategy, while each setting was further tested under three resolution levels (\ie{ full resolution, long side 3840/1920 pixels}). A fallback strategy that prioritizes preserving page count was applied when models fail to respond in most cases. For text input, we set multiple cut-off lengths to ensure compatibility. Details are provided in Appendix D. 

\paragraph{Evaluation Methods}
We adopt PoT prompts~\citep{pot}, which mitigate numerical errors~\citep{financemath,docmath-eval}, and assess accuracy under a tolerance of 0.2\%.


\begin{figure*}[t]
	\centering
	\includegraphics[width=0.72\textwidth]{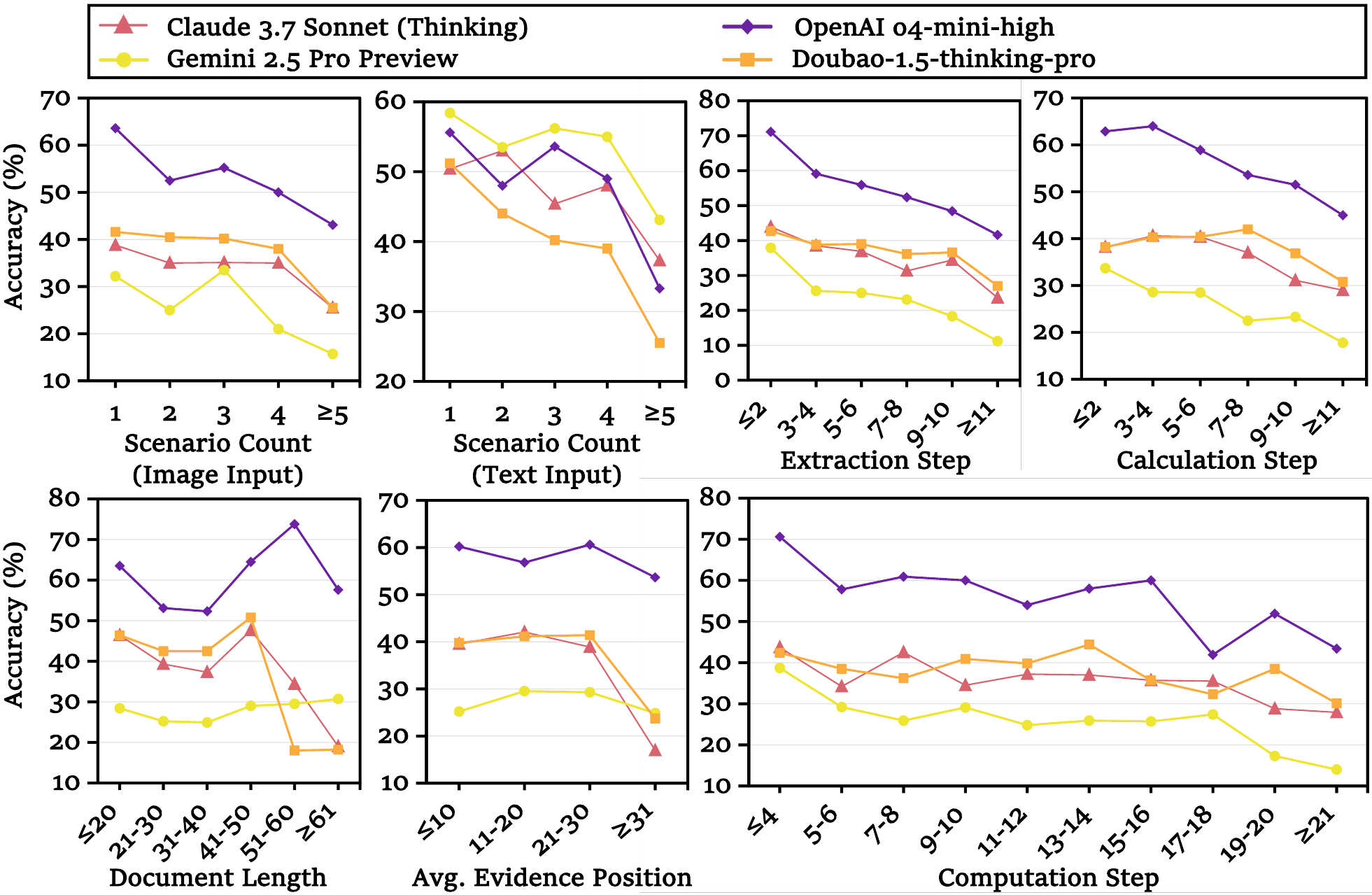} 
	\caption{Fine-grained results based on \textbf{(top left)} scenario count, \textbf{(bottom left)} document length, \textbf{(bottom middle)} average evidence position, and \textbf{(right)} the number of steps in numerical extraction, numerical calculation, and overall computation.} 
	\label{fig:fine-grained}
\end{figure*}
\subsection{Main Results}
Table~\ref{tab:main_result_revised} presents the results across all models. Our main findings are summarized as follows:

\noindent\textbf{\textit{Overall performance across models remains unsatisfactory.}}
None of the models achieved accuracy above the 60\% threshold in any of the settings. Within MLLMs, even the SoTA model OpenAI o4-mini-high reached only 58\% accuracy. Many models struggled with handling large-scale inputs, both visual and textual. Moreover, open-source models consistently underperformed proprietary models.

\noindent\textbf{\textit{Reasoning-enhanced models consistently outperform those without.}}  
Across both input settings, reasoning-enhanced models achieved substantially higher accuracy. Among proprietary models, the top three performers were all reasoning-enhanced. Notably, DeepSeek-R1~\citep{deepseekr1}, the only open-source large reasoning model (LRM) in the evaluation, achieved the highest accuracy (40.0\%) within its group.

\noindent\textbf{\textit{MLLMs face significant bottlenecks in processing long multimodal inputs. }}  
While MMLongBenchDoc~\citep{mmlongbench-doc} acknowledges the potential information loss introduced by OCR, most MLLMs still perform worse than OCR+LLM models on \ours, highlighting the bottlenecks MLLMs face when handling image input directly. Specifically, OpenAI o4-mini-high is the only model whose image input performance exceeded its text counterpart, indicating its superior multimodal reasoning capabilities.

\noindent\textbf{\textit{Models exhibit substantial disparities in visual understanding.}}  
In the OCR+LLM group, the accuracy gap among the top four proprietary models was under 12 points. However, this gap was notably larger in MLLMs (nearly 30 points between OpenAI o4-mini-high and Doubao-1.5-vision-pro). This indicates that visual understanding varies much more significantly across MLLMs, compared to relatively stable language understanding.

\subsection{Fine-Grained Analysis}
Table~\ref{tab:main_result_revised} and Figure~\ref{fig:fine-grained} also present the fine-grained results on the further analysis. Detailed results are provided in Appendix E. The key findings are as follows: 

\noindent\textit{\textbf{Current models struggle with multi-scenario tasks.}} All exhibit a notable decline in accuracy as the number of scenarios increases. This likely stems from the increased complexity of scenario combinations, requiring more assumptions and associations, thereby better evaluating models' stable reasoning capabilities in complex environments.

\noindent\textit{\textbf{Strong document understanding plays a critical role.}} OpenAI o4-mini-high and Gemini 2.5 Pro Preview maintain stable performance across varying document lengths, likely due to their robust contextual comprehension, while the other two models drop substantially. A similar trend is observed in Figure~\ref{fig:fine-grained} (bottom middle), where the average index position of evidence positively correlates with document length.

\noindent\textit{\textbf{Information extraction, rather than numerical calculation, has a greater impact on model performance in the PoT setting.}} Accuracy declines progressively with increasing computation steps, following similar patterns to both extraction and calculation performance. Given that calculation typically depends on prior extraction, we hypothesize that this step-dependent accuracy reduction is primarily driven by extraction errors, which aligns with both the PoT's advantage and subsequent error analysis.

\subsection{Error Analysis}
We randomly sampled 100 failure cases from OpenAI o4-mini-high. Each instance may exhibit multiple error types, which we categorize into four categories. Detailed examples and analysis are provided in Appendix F.
\setlist[itemize]{itemsep=0.1em, topsep=0.2em}
\begin{itemize}
    \item \textbf{Scenario Awareness Error (33/100):} Misinterpretation of task intent, contextual constraints, or key parameters, resulting in flawed reasoning paths.
    \item \textbf{Document Understanding Error (78/100)}: Failure to accurately locate or extract critical information from complex multimodal documents.
    \item \textbf{Knowledge Reasoning Error (44/100)}: Incorrect formula selection or invalid reasoning structures.
    \item \textbf{Numerical Calculation Error (5/100)}: Mistakes in calculation despite correct formulas, often due to precision loss, rounding, or intermediate step errors.
\end{itemize}

\begin{figure}[t]
	\centering
	\includegraphics[width=0.99\linewidth]{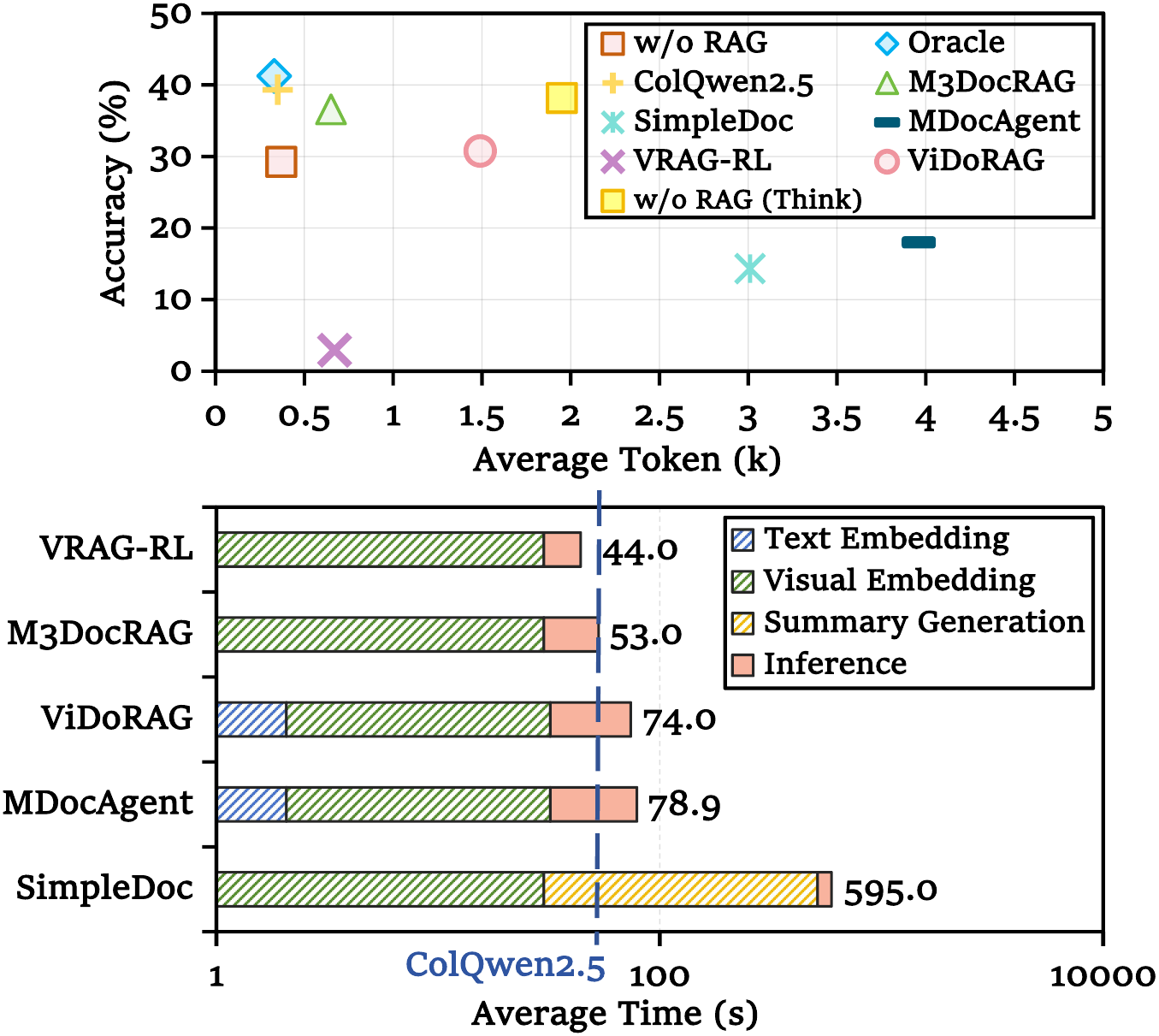}
	\caption{
		\textbf{(Top)} Accuracy and token consumption comparison of RAG methods. \textbf{(Bottom)} Runtime composition comparison of Agentic RAGs vs. ColQwen2.5.}
	\label{exp:acc_token_runtime}
\end{figure}

\subsection{RAG Analysis}
We evaluated 6 embedding models~\citep{contriever,bgem3,visrag,colpali} and 5 Agentic RAGs~\citep{cho2024m3docragmultimodalretrievalneed,wang-etal-2025-vidorag,mdocagent,jain-etal-2025-simpledoc,vrag-rl}. All Agentic RAGs employed ColQwen2.5 for retrieval and Doubao-1.5-vision-pro for generation. Methods with visual embeddings consistently outperformed text-only approaches, and ColQwen2.5 achieving the best performance. Agentic RAGs underperformed ColQwen2.5, despite consuming more tokens and time, as shown in Figure~\ref{exp:acc_token_runtime}. Detailed analysis is provided in Appendix H. The key findings are as follows:


\noindent\textbf{\textit{Agents based solely on semantic retrieval fall short in handling FinMMDocR's complex reasoning demands.}} SimpleDoc and MDocAgent attempt to enhance semantic representation through multimodal embeddings. However, they often miss the pages containing intermediate variables that are not explicitly stated in the question, resulting in incomplete information retrieval. ViDoRAG partially addresses this issue through an iterative workflow, simulating limited reasoning. Despite lower overall accuracy, it achieves more complete retrieval and reasoning coverage on most of the questions where both models and ColQwen2.5 failed.

\noindent\textbf{\textit{Agentic RAGs rely on predefined workflows and fall short of reasoning-enhanced models.}} ViDoRAG exhibits more numerical errors, like invalid significant figures, likely due to test-based output randomness and context-induced forgetting. Additionally, current frameworks heavily depend on upstream outputs that are rarely questioned or revised downstream, preventing error recovery.

\noindent\textbf{\textit{The effectiveness of visually focused strategies remains to be explored.}} VRAG-RL performed poorly on FinMMDocR, though understandable given the task difficulty. We attribute this to its small base model (7B), and the benefit of scaling up with reinforcement learning remains to be verified.



\section{Related Work}

Inspired by real-world financial analysis tasks, financial multimodal reasoning demands models to comprehend financial contexts, extract key data from visually dense multimodal financial documents, and perform precise numerical calculations to support multi-step reasoning. However, existing financial QA benchmarks and long-document VQA benchmarks fail to authentically model this task, exhibiting significant gaps. Benchmarks like FinQA~\citep{finqa}, TAT-QA~\citep{tatqa}, and ConvFinQA~\citep{convfinqa} only require simple information extraction and arithmetic operations under explicit conditions, while FinanceReasoning~\citep{financereasoning}, FinanceMath~\citep{financemath}, DocMath-Eval~\cite{docmath-eval}, and FinCode~\citep{bizbench} incorporate limited contexts with text-only inputs. FinMMR~\citep{finmmr}, FinMME~\citep{finmme}, and MME-Finance~\citep{mme-finance} evaluate models’ reasoning capabilities on single or few images. LongDocURL~\citep{deng-etal-2025-longdocurl} and MMLongBench-Doc~\citep{mmlongbench-doc} focus on generic multimodal long-document QA, where merely 6\% and 8\% of tasks involve financial numerical reasoning, further constrained by the scarcity and diversity of domain-specific documents.

MLLMs~\citep{doubao1.5visionpro, gpt4o, grok2vision, qwen25vl, llama4maverick, mistralsmall3.1, gemma3} and LMRMs~\citep{o4mini, doubao1.5thinkingpro, claude3.7sonnet, gemini2.5propreview} offer promising solutions for end-to-end financial multimodal reasoning, leveraging expanded context windows and enhanced reasoning capacities. Concurrently, RAG methods have alleviated models’ long-document processing burdens, retrieving relevant pages via semantic similarity between queries and pages. Following text-based RAGs (\eg{ BM25, Contriever~\citep{contriever}, BGE-M3~\citep{bgem3}}), vision RAGs like VisRAG~\citep{visrag}, ColPali~\citep{colpali}, and ColQwen2.5~\citep{colpali} have improved multimodal retrieval performance. Agentic RAG frameworks such as M3DocRAG~\citep{cho2024m3docragmultimodalretrievalneed}, ViDoRAG~\citep{wang-etal-2025-vidorag}, MDocAgent~\citep{mdocagent}, SimpleDoc~\citep{jain-etal-2025-simpledoc}, and VRAG-RL~\citep{vrag-rl} employ multi-agent collaboration for flexible reasoning.


\section{Conclusion}
We introduce \ours, a financial multimodal reasoning benchmark for evaluating MLLMs' professional document understanding and precise multi-step computation in real-world financial scenarios, alongside comprehensive assessments of diverse RAG methods in this complex setting. Extensive experiments reveal significant performance gaps between MLLMs and human experts, with no model exceeding 60\% accuracy. While RAG shows promise for information retrieval and reducing visual burdens, fundamental improvements in models' reasoning capabilities and RAG efficiency remain critical future directions. We hope this work establishes foundations for advancing domain-specific multimodal reasoning.

\clearpage
\section*{Acknowledgments}
This work is supported by the National Natural Science Foundation of China (Grant Nos. 62176026, 62473271), the Beijing Natural Science Foundation (Grant Nos. QY25345, QY25338), the Fundamental Research Funds for the Beijing University of Posts and Telecommunications (Grant No. 2025AI4S03), the BUPT Innovation and Entrepreneurship Support Program (Grant Nos. 2025-YC-A033, 2025-YC-A042), and data support from Hithink RoyalFlush Information Network Co., Ltd. This work is also supported by the Engineering Research Center of Information Networks, Ministry of Education, China. We would also like to thank the anonymous reviewers and area chairs for constructive discussions and feedback.

\small

\begin{thebibliography}{45}
\providecommand{\natexlab}[1]{#1}

\bibitem[{AI(2025)}]{mistralsmall3.1}
AI, M. 2025.
\newblock Mistral Small 3.1.
\newblock \url{https://mistral.ai/news/mistral-small-3-1}.
\newblock Accessed: 2025-03-17.

\bibitem[{AI@Meta(2025)}]{llama4maverick}
AI@Meta. 2025.
\newblock The Llama 4 herd: The beginning of a new era of natively multimodal AI innovation.
\newblock \url{https://ai.meta.com/blog/llama-4-multimodal-intelligence/}.
\newblock Accessed: 2025-04-05.

\bibitem[{Anthropic(2025)}]{claude3.7sonnet}
Anthropic. 2025.
\newblock Claude 3.7 Sonnet and Claude Code.
\newblock \url{https://www.anthropic.com/news/claude-3-7-sonnet}.
\newblock Accessed: 2025-02-25.

\bibitem[{Bai et~al.(2025)Bai, Chen, Liu, Wang, Ge, Song, Dang, Wang, Wang, Tang, Zhong, Zhu, Yang, Li, Wan, Wang, Ding, Fu, Xu, Ye, Zhang, Xie, Cheng, Zhang, Yang, Xu, and Lin}]{qwen25vl}
Bai, S.; Chen, K.; Liu, X.; Wang, J.; Ge, W.; Song, S.; Dang, K.; Wang, P.; Wang, S.; Tang, J.; Zhong, H.; Zhu, Y.; Yang, M.; Li, Z.; Wan, J.; Wang, P.; Ding, W.; Fu, Z.; Xu, Y.; Ye, J.; Zhang, X.; Xie, T.; Cheng, Z.; Zhang, H.; Yang, Z.; Xu, H.; and Lin, J. 2025.
\newblock Qwen2.5-VL Technical Report.
\newblock arXiv:2502.13923.

\bibitem[{ByteDance(2025{\natexlab{a}})}]{doubao1.5thinkingpro}
ByteDance. 2025{\natexlab{a}}.
\newblock Doubao-1.5-thinking-pro Model Card.
\newblock \url{https://console.volcengine.com/ark/region:ark+cn-beijing/model/detail?Id=doubao-1-5-thinking-pro}.
\newblock Accessed: 2025-04-15.

\bibitem[{ByteDance(2025{\natexlab{b}})}]{doubao1.5visionpro}
ByteDance. 2025{\natexlab{b}}.
\newblock Doubao-1.5-vision-pro Model Card.
\newblock \url{https://console.volcengine.com/ark/region:ark+cn-beijing/model/detail?Id=doubao-1-5-vision-pro}.
\newblock Accessed: 2025-03-28.

\bibitem[{Chen et~al.(2024)Chen, Xiao, Zhang, Luo, Lian, and Liu}]{bgem3}
Chen, J.; Xiao, S.; Zhang, P.; Luo, K.; Lian, D.; and Liu, Z. 2024.
\newblock {M}3-Embedding: Multi-Linguality, Multi-Functionality, Multi-Granularity Text Embeddings Through Self-Knowledge Distillation.
\newblock In Ku, L.-W.; Martins, A.; and Srikumar, V., eds., \emph{Findings of the Association for Computational Linguistics: ACL 2024}, 2318--2335. Bangkok, Thailand: Association for Computational Linguistics.

\bibitem[{Chen et~al.(2023)Chen, Ma, Wang, and Cohen}]{pot}
Chen, W.; Ma, X.; Wang, X.; and Cohen, W.~W. 2023.
\newblock Program of Thoughts Prompting: Disentangling Computation from Reasoning for Numerical Reasoning Tasks.
\newblock \emph{Transactions on Machine Learning Research}.

\bibitem[{Chen et~al.(2021)Chen, Chen, Smiley, Shah, Borova, Langdon, Moussa, Beane, Huang, Routledge, and Wang}]{finqa}
Chen, Z.; Chen, W.; Smiley, C.; Shah, S.; Borova, I.; Langdon, D.; Moussa, R.; Beane, M.; Huang, T.-H.; Routledge, B.; and Wang, W.~Y. 2021.
\newblock {F}in{QA}: A Dataset of Numerical Reasoning over Financial Data.
\newblock In Moens, M.-F.; Huang, X.; Specia, L.; and Yih, S. W.-t., eds., \emph{Proceedings of the 2021 Conference on Empirical Methods in Natural Language Processing}, 3697--3711. Online and Punta Cana, Dominican Republic: Association for Computational Linguistics.

\bibitem[{Chen et~al.(2022)Chen, Li, Smiley, Ma, Shah, and Wang}]{convfinqa}
Chen, Z.; Li, S.; Smiley, C.; Ma, Z.; Shah, S.; and Wang, W.~Y. 2022.
\newblock {C}onv{F}in{QA}: Exploring the Chain of Numerical Reasoning in Conversational Finance Question Answering.
\newblock In Goldberg, Y.; Kozareva, Z.; and Zhang, Y., eds., \emph{Proceedings of the 2022 Conference on Empirical Methods in Natural Language Processing}, 6279--6292. Abu Dhabi, United Arab Emirates: Association for Computational Linguistics.

\bibitem[{Cho et~al.(2025)Cho, Mahata, Irsoy, He, and Bansal}]{cho2024m3docragmultimodalretrievalneed}
Cho, J.; Mahata, D.; Irsoy, O.; He, Y.; and Bansal, M. 2025.
\newblock M3DocVQA: Multi-modal Multi-page Multi-document Understanding.
\newblock In \emph{Proceedings of the IEEE/CVF International Conference on Computer Vision (ICCV) Workshops}, 6178--6188.

\bibitem[{DeepMind(2025)}]{gemini2.5propreview}
DeepMind, G. 2025.
\newblock Build rich, interactive web apps with an updated Gemini 2.5 Pro.
\newblock \url{https://blog.google/products/gemini/gemini-2-5-pro-updates/}.
\newblock Accessed: 2025-05-06.

\bibitem[{Deng et~al.(2025)Deng, Yuan, Bu, Wang, Li, Xu, Li, Gao, Song, Zheng, and Liu}]{deng-etal-2025-longdocurl}
Deng, C.; Yuan, J.; Bu, P.; Wang, P.; Li, Z.-Z.; Xu, J.; Li, X.-H.; Gao, Y.; Song, J.; Zheng, B.; and Liu, C.-L. 2025.
\newblock {L}ong{D}oc{URL}: a Comprehensive Multimodal Long Document Benchmark Integrating Understanding, Reasoning, and Locating.
\newblock In Che, W.; Nabende, J.; Shutova, E.; and Pilehvar, M.~T., eds., \emph{Proceedings of the 63rd Annual Meeting of the Association for Computational Linguistics (Volume 1: Long Papers)}, 1135--1159. Vienna, Austria: Association for Computational Linguistics.
\newblock ISBN 979-8-89176-251-0.

\bibitem[{Faysse et~al.(2025)Faysse, Sibille, Wu, Omrani, Viaud, HUDELOT, and Colombo}]{colpali}
Faysse, M.; Sibille, H.; Wu, T.; Omrani, B.; Viaud, G.; HUDELOT, C.; and Colombo, P. 2025.
\newblock ColPali: Efficient Document Retrieval with Vision Language Models.
\newblock In \emph{The Thirteenth International Conference on Learning Representations}.

\bibitem[{Gan et~al.(2025)Gan, Zhang, Li, Wu, Lin, Liu, Wu, Fu, Xu, Zhang, and Dai}]{mme-finance}
Gan, Z.; Zhang, D.; Li, H.; Wu, Y.; Lin, X.; Liu, J.; Wu, H.; Fu, C.; Xu, Z.; Zhang, R.; and Dai, Y. 2025.
\newblock MME-Finance: A Multimodal Finance Benchmark for Expert-level Understanding and Reasoning.
\newblock In \emph{Proceedings of the 33rd ACM International Conference on Multimedia}, MM '25, 12867–12874. New York, NY, USA: Association for Computing Machinery.
\newblock ISBN 9798400720352.

\bibitem[{Goyal et~al.(2017)Goyal, Khot, Summers-Stay, Batra, and Parikh}]{vqa}
Goyal, Y.; Khot, T.; Summers-Stay, D.; Batra, D.; and Parikh, D. 2017.
\newblock Making the v in VQA Matter: Elevating the Role of Image Understanding in Visual Question Answering.
\newblock In \emph{Proceedings of the IEEE Conference on Computer Vision and Pattern Recognition (CVPR)}.

\bibitem[{Guo et~al.(2025)Guo, Yang, Zhang, Song, Wang, Zhu, Xu, Zhang, Ma, Bi et~al.}]{deepseekr1}
Guo, D.; Yang, D.; Zhang, H.; Song, J.; Wang, P.; Zhu, Q.; Xu, R.; Zhang, R.; Ma, S.; Bi, X.; et~al. 2025.
\newblock Deepseek-r1 incentivizes reasoning in llms through reinforcement learning.
\newblock \emph{Nature}, 645(8081): 633--638.

\bibitem[{Han et~al.(2025)Han, Xia, Zhang, Sun, Li, Zhu, and Yao}]{mdocagent}
Han, S.; Xia, P.; Zhang, R.; Sun, T.; Li, Y.; Zhu, H.; and Yao, H. 2025.
\newblock MDocAgent: A Multi-Modal Multi-Agent Framework for Document Understanding.
\newblock arXiv:2503.13964.

\bibitem[{Izacard et~al.(2022)Izacard, Caron, Hosseini, Riedel, Bojanowski, Joulin, and Grave}]{contriever}
Izacard, G.; Caron, M.; Hosseini, L.; Riedel, S.; Bojanowski, P.; Joulin, A.; and Grave, E. 2022.
\newblock Unsupervised Dense Information Retrieval with Contrastive Learning.
\newblock \emph{Transactions on Machine Learning Research}.

\bibitem[{Jain et~al.(2025)Jain, Wu, Zeng, Liu, Dai, Shao, Wu, and Wang}]{jain-etal-2025-simpledoc}
Jain, C.; Wu, Y.; Zeng, Y.; Liu, J.; Dai, S.; Shao, Z.; Wu, Q.; and Wang, H. 2025.
\newblock {S}imple{D}oc: {M}ulti{-}{M}odal Document Understanding with {D}ual{-}{C}ue Page Retrieval and Iterative Refinement.
\newblock In Christodoulopoulos, C.; Chakraborty, T.; Rose, C.; and Peng, V., eds., \emph{Proceedings of the 2025 Conference on Empirical Methods in Natural Language Processing}, 28398--28415. Suzhou, China: Association for Computational Linguistics.
\newblock ISBN 979-8-89176-332-6.

\bibitem[{Krumdick et~al.(2024)Krumdick, Koncel-Kedziorski, Lai, Reddy, Lovering, and Tanner}]{bizbench}
Krumdick, M.; Koncel-Kedziorski, R.; Lai, V.~D.; Reddy, V.; Lovering, C.; and Tanner, C. 2024.
\newblock {B}iz{B}ench: A Quantitative Reasoning Benchmark for Business and Finance.
\newblock In Ku, L.-W.; Martins, A.; and Srikumar, V., eds., \emph{Proceedings of the 62nd Annual Meeting of the Association for Computational Linguistics (Volume 1: Long Papers)}, 8309--8332. Bangkok, Thailand: Association for Computational Linguistics.

\bibitem[{Li et~al.(2025)Li, Liu, Li, Zhang, Xu, Chen, Shi, Jiang, Wang, Wang, Huang, Zhao, Jiang, Hong, Wang, Tian, Huai, Luo, Luo, Zhang, Hu, and Zhang}]{prtp}
Li, Y.; Liu, Z.; Li, Z.; Zhang, X.; Xu, Z.; Chen, X.; Shi, H.; Jiang, S.; Wang, X.; Wang, J.; Huang, S.; Zhao, X.; Jiang, B.; Hong, L.; Wang, L.; Tian, Z.; Huai, B.; Luo, W.; Luo, W.; Zhang, Z.; Hu, B.; and Zhang, M. 2025.
\newblock Perception, Reason, Think, and Plan: A Survey on Large Multimodal Reasoning Models.
\newblock arXiv:2505.04921.

\bibitem[{Liu et~al.(2023)Liu, Li, Wu, and Lee}]{vit}
Liu, H.; Li, C.; Wu, Q.; and Lee, Y.~J. 2023.
\newblock Visual Instruction Tuning.
\newblock In Oh, A.; Naumann, T.; Globerson, A.; Saenko, K.; Hardt, M.; and Levine, S., eds., \emph{Advances in Neural Information Processing Systems}, volume~36, 34892--34916. Curran Associates, Inc.

\bibitem[{Lu et~al.(2024)Lu, Bansal, Xia, Liu, Li, Hajishirzi, Cheng, Chang, Galley, and Gao}]{mathvista}
Lu, P.; Bansal, H.; Xia, T.; Liu, J.; Li, C.; Hajishirzi, H.; Cheng, H.; Chang, K.-W.; Galley, M.; and Gao, J. 2024.
\newblock MathVista: Evaluating Mathematical Reasoning of Foundation Models in Visual Contexts.
\newblock In \emph{The Twelfth International Conference on Learning Representations}.

\bibitem[{Luo et~al.(2025)Luo, Kou, Yang, Luo, Huang, Xiao, Peng, Liu, Ji, Liu, Han, Zhang, and Guo}]{finmme}
Luo, J.; Kou, Z.; Yang, L.; Luo, X.; Huang, J.; Xiao, Z.; Peng, J.; Liu, C.; Ji, J.; Liu, X.; Han, S.; Zhang, M.; and Guo, Y. 2025.
\newblock {F}in{MME}: Benchmark Dataset for Financial Multi-Modal Reasoning Evaluation.
\newblock In Che, W.; Nabende, J.; Shutova, E.; and Pilehvar, M.~T., eds., \emph{Proceedings of the 63rd Annual Meeting of the Association for Computational Linguistics (Volume 1: Long Papers)}, 29465--29489. Vienna, Austria: Association for Computational Linguistics.
\newblock ISBN 979-8-89176-251-0.

\bibitem[{Ma et~al.(2024)Ma, Zang, Chen, Chen, Jiao, Li, Lu, Liu, Ma, Dong, Zhang, Pan, Jiang, Wang, Cao, and Sun}]{mmlongbench-doc}
Ma, Y.; Zang, Y.; Chen, L.; Chen, M.; Jiao, Y.; Li, X.; Lu, X.; Liu, Z.; Ma, Y.; Dong, X.; Zhang, P.; Pan, L.; Jiang, Y.-G.; Wang, J.; Cao, Y.; and Sun, A. 2024.
\newblock MMLONGBENCH-DOC: Benchmarking Long-context Document Understanding with Visualizations.
\newblock In Globerson, A.; Mackey, L.; Belgrave, D.; Fan, A.; Paquet, U.; Tomczak, J.; and Zhang, C., eds., \emph{Advances in Neural Information Processing Systems}, volume~37, 95963--96010. Curran Associates, Inc.

\bibitem[{OpenAI(2024)}]{gpt4o}
OpenAI. 2024.
\newblock Hello GPT-4o.
\newblock \url{https://openai.com/index/hello-gpt-4o/}.
\newblock Accessed: 2024-05-13.

\bibitem[{OpenAI(2025)}]{o4mini}
OpenAI. 2025.
\newblock OpenAI o3 and o4-mini System Card.
\newblock \url{https://openai.com/index/o3-o4-mini-system-card/}.
\newblock Accessed: 2025-04-16.

\bibitem[{Singh et~al.(2025)Singh, Ehtesham, Kumar, and Khoei}]{agenticrag}
Singh, A.; Ehtesham, A.; Kumar, S.; and Khoei, T.~T. 2025.
\newblock Agentic Retrieval-Augmented Generation: A Survey on Agentic RAG.
\newblock arXiv:2501.09136.

\bibitem[{Singh et~al.(2019)Singh, Natarajan, Shah, Jiang, Chen, Batra, Parikh, and Rohrbach}]{textvqa}
Singh, A.; Natarajan, V.; Shah, M.; Jiang, Y.; Chen, X.; Batra, D.; Parikh, D.; and Rohrbach, M. 2019.
\newblock Towards VQA Models That Can Read.
\newblock In \emph{Proceedings of the IEEE/CVF Conference on Computer Vision and Pattern Recognition (CVPR)}.

\bibitem[{Smith(2007)}]{tesseract}
Smith, R. 2007.
\newblock An Overview of the Tesseract OCR Engine.
\newblock In \emph{Ninth International Conference on Document Analysis and Recognition (ICDAR 2007)}, volume~2, 629--633.

\bibitem[{Tanaka et~al.(2023)Tanaka, Nishida, Nishida, Hasegawa, Saito, and Saito}]{slidevqa}
Tanaka, R.; Nishida, K.; Nishida, K.; Hasegawa, T.; Saito, I.; and Saito, K. 2023.
\newblock SlideVQA: A Dataset for Document Visual Question Answering on Multiple Images.
\newblock \emph{Proceedings of the AAAI Conference on Artificial Intelligence}, 37(11): 13636--13645.

\bibitem[{Tang et~al.(2025{\natexlab{a}})Tang, E, Liu, Yang, Li, Rong, He, Hao, Hu, Ji, Ma, Ji, Zhang, Ma, Zheng, Liu, Huang, Hu, Huang, Xie, and Peng}]{finmmr}
Tang, Z.; E, H.; Liu, J.; Yang, Z.; Li, R.; Rong, Z.; He, H.; Hao, Z.; Hu, X.; Ji, K.; Ma, Z.; Ji, M.; Zhang, J.; Ma, C.; Zheng, Q.; Liu, Y.; Huang, Y.; Hu, X.; Huang, Q.; Xie, Z.; and Peng, S. 2025{\natexlab{a}}.
\newblock FinMMR: Make Financial Numerical Reasoning More Multimodal, Comprehensive, and Challenging.
\newblock In \emph{Proceedings of the IEEE/CVF International Conference on Computer Vision (ICCV)}, 3245--3257.

\bibitem[{Tang et~al.(2025{\natexlab{b}})Tang, E, Ma, He, Liu, Yang, Rong, Li, Ji, Huang, Hu, Liu, and Zheng}]{financereasoning}
Tang, Z.; E, H.; Ma, Z.; He, H.; Liu, J.; Yang, Z.; Rong, Z.; Li, R.; Ji, K.; Huang, Q.; Hu, X.; Liu, Y.; and Zheng, Q. 2025{\natexlab{b}}.
\newblock {F}inance{R}easoning: Benchmarking Financial Numerical Reasoning More Credible, Comprehensive and Challenging.
\newblock In Che, W.; Nabende, J.; Shutova, E.; and Pilehvar, M.~T., eds., \emph{Proceedings of the 63rd Annual Meeting of the Association for Computational Linguistics (Volume 1: Long Papers)}, 15721--15749. Vienna, Austria: Association for Computational Linguistics.
\newblock ISBN 979-8-89176-251-0.

\bibitem[{Team et~al.(2025)Team, Kamath, Ferret, Pathak, Vieillard, Merhej, Perrin, Matejovicova, Ramé, Rivière et~al.}]{gemma3}
Team, G.; Kamath, A.; Ferret, J.; Pathak, S.; Vieillard, N.; Merhej, R.; Perrin, S.; Matejovicova, T.; Ramé, A.; Rivière, M.; et~al. 2025.
\newblock Gemma 3 Technical Report.
\newblock arXiv:2503.19786.

\bibitem[{Wang et~al.(2024)Wang, Pan, Shi, Lu, Ren, Zhou, Zhan, and Li}]{math-vision}
Wang, K.; Pan, J.; Shi, W.; Lu, Z.; Ren, H.; Zhou, A.; Zhan, M.; and Li, H. 2024.
\newblock Measuring Multimodal Mathematical Reasoning with MATH-Vision Dataset.
\newblock In Globerson, A.; Mackey, L.; Belgrave, D.; Fan, A.; Paquet, U.; Tomczak, J.; and Zhang, C., eds., \emph{Advances in Neural Information Processing Systems}, volume~37, 95095--95169. Curran Associates, Inc.

\bibitem[{Wang et~al.(2025{\natexlab{a}})Wang, Ding, Chen, Wu, Wang, Xie, and Zhao}]{wang-etal-2025-vidorag}
Wang, Q.; Ding, R.; Chen, Z.; Wu, W.; Wang, S.; Xie, P.; and Zhao, F. 2025{\natexlab{a}}.
\newblock {V}i{D}o{RAG}: Visual Document Retrieval-Augmented Generation via Dynamic Iterative Reasoning Agents.
\newblock In Christodoulopoulos, C.; Chakraborty, T.; Rose, C.; and Peng, V., eds., \emph{Proceedings of the 2025 Conference on Empirical Methods in Natural Language Processing}, 9124--9145. Suzhou, China: Association for Computational Linguistics.
\newblock ISBN 979-8-89176-332-6.

\bibitem[{Wang et~al.(2025{\natexlab{b}})Wang, Ding, Zeng, Chen, Chen, Wang, Xie, Huang, and Zhao}]{vrag-rl}
Wang, Q.; Ding, R.; Zeng, Y.; Chen, Z.; Chen, L.; Wang, S.; Xie, P.; Huang, F.; and Zhao, F. 2025{\natexlab{b}}.
\newblock {VRAG}-{RL}: Empower Vision-Perception-Based {RAG} for Visually Rich Information Understanding via Iterative Reasoning with Reinforcement Learning.
\newblock In \emph{The Thirty-ninth Annual Conference on Neural Information Processing Systems}.

\bibitem[{xAI(2024)}]{grok2vision}
xAI. 2024.
\newblock Grok 2 Vision Model Card.
\newblock \url{https://docs.x.ai/docs/models/grok-2-vision-1212}.
\newblock Accessed: 2024-12-12.

\bibitem[{Yu et~al.(2025)Yu, Tang, Xu, Cui, Ran, Yan, Liu, Wang, Han, Liu, and Sun}]{visrag}
Yu, S.; Tang, C.; Xu, B.; Cui, J.; Ran, J.; Yan, Y.; Liu, Z.; Wang, S.; Han, X.; Liu, Z.; and Sun, M. 2025.
\newblock Vis{RAG}: Vision-based Retrieval-augmented Generation on Multi-modality Documents.
\newblock In \emph{The Thirteenth International Conference on Learning Representations}.

\bibitem[{Yu et~al.(2024)Yu, Yang, Li, Wang, Lin, Liu, Wang, and Wang}]{yu2024mmvet}
Yu, W.; Yang, Z.; Li, L.; Wang, J.; Lin, K.; Liu, Z.; Wang, X.; and Wang, L. 2024.
\newblock {MM}-Vet: Evaluating Large Multimodal Models for Integrated Capabilities.
\newblock In \emph{Forty-first International Conference on Machine Learning}.

\bibitem[{Zellers et~al.(2019)Zellers, Bisk, Farhadi, and Choi}]{vcr}
Zellers, R.; Bisk, Y.; Farhadi, A.; and Choi, Y. 2019.
\newblock From Recognition to Cognition: Visual Commonsense Reasoning.
\newblock In \emph{Proceedings of the IEEE/CVF Conference on Computer Vision and Pattern Recognition (CVPR)}.

\bibitem[{Zhao et~al.(2024{\natexlab{a}})Zhao, Liu, Long, Zhang, Zhao, and Cohan}]{financemath}
Zhao, Y.; Liu, H.; Long, Y.; Zhang, R.; Zhao, C.; and Cohan, A. 2024{\natexlab{a}}.
\newblock FinanceMATH: Knowledge-Intensive Math Reasoning in Finance Domains.
\newblock In Ku, L.-W.; Martins, A.; and Srikumar, V., eds., \emph{Proceedings of the 62nd Annual Meeting of the Association for Computational Linguistics (Volume 1: Long Papers)}, 12841--12858. Bangkok, Thailand: Association for Computational Linguistics.

\bibitem[{Zhao et~al.(2024{\natexlab{b}})Zhao, Long, Liu, Kamoi, Nan, Chen, Liu, Tang, Zhang, and Cohan}]{docmath-eval}
Zhao, Y.; Long, Y.; Liu, H.; Kamoi, R.; Nan, L.; Chen, L.; Liu, Y.; Tang, X.; Zhang, R.; and Cohan, A. 2024{\natexlab{b}}.
\newblock {D}oc{M}ath-Eval: Evaluating Math Reasoning Capabilities of {LLM}s in Understanding Long and Specialized Documents.
\newblock In Ku, L.-W.; Martins, A.; and Srikumar, V., eds., \emph{Proceedings of the 62nd Annual Meeting of the Association for Computational Linguistics (Volume 1: Long Papers)}, 16103--16120. Bangkok, Thailand: Association for Computational Linguistics.

\bibitem[{Zhu et~al.(2021)Zhu, Lei, Huang, Wang, Zhang, Lv, Feng, and Chua}]{tatqa}
Zhu, F.; Lei, W.; Huang, Y.; Wang, C.; Zhang, S.; Lv, J.; Feng, F.; and Chua, T.-S. 2021.
\newblock {TAT}-{QA}: A Question Answering Benchmark on a Hybrid of Tabular and Textual Content in Finance.
\newblock In Zong, C.; Xia, F.; Li, W.; and Navigli, R., eds., \emph{Proceedings of the 59th Annual Meeting of the Association for Computational Linguistics and the 11th International Joint Conference on Natural Language Processing (Volume 1: Long Papers)}, 3277--3287. Online: Association for Computational Linguistics.

\end{thebibliography}

\input{sections/ReproducibilityChecklist}

\clearpage              


\section*{Supplementary material}

# Technical Appendix for FinMMDocR

**Contents**





# A   Examples from FinMMDocR Categorized by Scenario

## A.1   Example 1

**Financial Statement Analysis**

**Question**: Based on the detailed financial forecast tables provided at the end of the report, analyze the company's working capital management efficiency related to its customer collections for the fiscal year 2025. Using the year-end balances of (Accounts Receivable) for 2024 and 2025 to calculate the average balance for 2025, and the corresponding (Total Operating Revenue) for 2025, calculate the implied average collection period for receivables during 2025 (assume 365 days in a year, round to one decimal place, unit: days).
**Keyword:** Accounts Receivable Analysis
**Python Solution:**

```python
import numpy as np
def solution():
    # Define variables with their values
    ar_2024 = 565 # million yuan, Financial Forecasts p. 21
    ar_2025 = 372 # million yuan, Financial Forecasts p. 21
    revenue_2025 = 390.23 # million yuan, Financial Forecasts p. 21
    # Calculate Average Accounts Receivable for 2025
    avg_ar_2025 = (ar_2024 + ar_2025) / 2
    # Calculate Accounts Receivable Turnover
    # Avoid division by zero if avg_ar_2025 happens to be zero
    if avg_ar_2025 == 0:
        ar_turnover = 0
    else:
        ar_turnover = revenue_2025 / avg_ar_2025
    # Calculate Days Sales Outstanding (DSO)
    # Avoid division by zero if ar_turnover is zero
    if ar_turnover == 0:
        dso_2025 = np.inf # Or handle as appropriate, e.g., 0 or NaN
    else:
        dso_2025 = 365 / ar_turnover
    # Round the final result
    answer = round(dso_2025, 1)
    # Return final result
    return answer
```

**Extract: 3 Calculate: 4 Answer: 438.2**

## A.2 Example 2

**Portfolio Management**

**Question**: Consider a hypothetical portfolio representing the combined assets of the top 40 performing Fixed Income Plus funds mentioned in the report analysis section (2.2. Asset Allocation of Top-Performing Funds). This portfolio initially mirrors the average asset allocation described for these funds, particularly their stated average allocation percentage to convertible bonds and the breakdown of that convertible bond allocation by type (equity-biased, balanced, debt-biased). Suppose the portfolio manager decides to realign the proportions of the different types of convertible bonds within their existing total convertible bond allocation to precisely match the overall market's convertible bond type proportions as reported at the end of Q4 2024. Calculate the resulting absolute change in the percentage allocation to balanced-type convertible bonds, expressed as a percentage of the *total portfolio assets. (Round the answer to three decimal places).
**Keyword:** Convertible Bond Allocation
**Python Solution:**

```
def solution():
    total_cb_alloc_top40 = 0.247
    initial_prop_balanced_top40 = 0.442
    # initial_prop_equity_top40 = 0.100
    # initial_prop_debt_top40 = 0.459 # Sums to 1.001, use as is
    # Data from Figure 12 / Page 13 text (Market Q4 CB type
        proportions)
    market_q4_prop_balanced = 0.460
    # market_q4_prop_equity = 0.130
    # market_q4_prop_debt = 0.410 # Sums to 1.000
    # Calculate initial total portfolio allocation to balanced CBs
    initial_total_alloc_balanced = total_cb_alloc_top40 * 
        initial_prop_balanced_top40
    # Calculate new total portfolio allocation to balanced CBs after
        rebalancing type proportions
    new_total_alloc_balanced = total_cb_alloc_top40 *
        market_q4_prop_balanced
    # Calculate the absolute change in percentage points
    change_percentage_points = (new_total_alloc_balanced -
        initial_total_alloc_balanced) * 100
    # Round to three decimal places
    answer = round(change_percentage_points, 3)
    return answer
```

**Extract: 3 Calculate: 4 Answer: 0.445**

## A.3 Example 3

> ### Asset & Equity Valuation
>
> [Figures from report: Table 11 on page 9 showing sector valuation data; Figure 18 on page 13 showing EPS beat/meet/miss distribution across S&P 500 sectors]
>
> **Question**: Consider the Industrials sector. Assume that the market adjusts this sector's Price-to-Earnings (P/E) ratio relative to its current level based on how its Q3 2023 EPS beat rate compares to the overall S&P 500 average EPS beat rate reported. Specifically, assume the P/E ratio adjustment factor is equal to the ratio of the Industrials sector's EPS beat percentage to the S&P 500 average EPS beat percentage. Using the current Total Market Cap and P/E data for the Industrials sector from the report, calculate the sector's *new* implied Total Market Cap after this hypothetical P/E adjustment. Provide the answer in billions of USD, rounded to two decimal places.
>
> **Keyword:** Sector Valuation Adjustment
>
> **Python Solution:**
>
> ```
> def solution():
>     # Define variables with their values from the report
>     industrials_eps_beat_rate = 0.88 # Figure 18, page 13 (88%)
>     sp500_eps_beat_rate = 0.82    # Figure 18, page 13 (82%)
>     industrials_market_cap_billion_usd = 4751.15 # Table 11, page 9
>     # Note: P/E ratio is not strictly needed if using the direct
>         adjustment factor method
>     # industrials_pe_ratio = 21.45 # Table 11, page 9
>     # Calculate the adjustment factor
>     adjustment_factor = industrials_eps_beat_rate /
>         sp500_eps_beat_rate
>     # Calculate the new implied market cap
>     new_market_cap = industrials_market_cap_billion_usd *
>         adjustment_factor
>     # Round final result to two decimal places
>     answer = round(new_market_cap, 2)
>     # Return final result
>     return answer
> ```
>
> **Extract:3 Calculate:3 Answer:5098.8**

## A.4 Example 4

**Corporate Finance & Capital Management**

**Question**: Analyze the financial performance impact on Hua Run San Jiu's prescription drug (RX) business division stemming from external market pressures and policy changes between 2019 and 2023. Quantify the total adverse gross profit deviation for this specific division in the full year 2023. Calculate this deviation by comparing the actual reported gross profit for the division in 2023 against a hypothetical scenario where the division had been able to maintain the gross margin percentage it achieved in 2019, applied to its actual 2023 revenue. (Round to two decimal places, unit: Million Yuan).

**Keyword:** Gross Profit Deviation Analysis

**Python Solution:**

```
def solution():
    # Define variables with their values
    # RX Segment Data 2023A (from page 29 forecast table)
    rx_revenue_2023a = 5220.4
    rx_actual_margin_2023a = 0.521
    # RX Segment Data 2019 (Hypothetical Baseline from Figure 7, page
        8)
    rx_hypothetical_margin_2019 = 0.80
    # Calculate Actual RX Gross Profit 2023
    rx_actual_gp_2023a = rx_revenue_2023a * rx_actual_margin_2023a
    # Calculate Hypothetical RX Gross Profit 2023
    rx_hypothetical_gp_2023a = rx_revenue_2023a *
        rx_hypothetical_margin_2019
    # Calculate the Adverse Deviation
    adverse_deviation = rx_hypothetical_gp_2023a - rx_actual_gp_2023a
    # Round to two decimal places
    answer = round(adverse_deviation, 2)
    # Return final result
    return answer
```

**Extract:3 Calculate:4 Answer:1456.49**

## A.5 Example 5

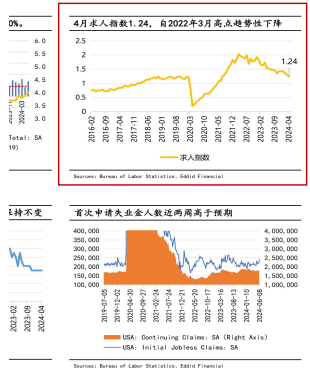

Market & Industry Analysis

**Question**: Construct a Hypothetical Semiconductor Growth Premium Index based on the following assumptions: The index is calculated as '(Base Growth Rate / 100) * (Job Openings Ratio Factor) / (Inflation Expectation Factor)'. The Base Growth Rate is the forecasted compound annual growth rate for the global semiconductor market from 2024 to 2033 mentioned in the report. The Job Openings Ratio Factor is derived from the latest reported (Job Openings to Unemployed Persons ratio) value mentioned in the report, calculated as (Reported / 1.0), using 1.0 as the baseline normal ratio. The Inflation Expectation Factor is derived from the latest reported long-term (5-year) consumer inflation expectation mentioned, calculated as (Reported 5-year Inflation Expectation Rate / 2.0), assuming 2.0 is the target/neutral long-term inflation rate. Calculate the value of this index based on the data provided in the report (round to four decimal places).

**Keyword:** Semiconductor Growth Premium Modeling

**Python Solution:**

```
def solution():
    # Data Extraction
    base_growth_rate_percent = 7.64 # Page 12, text paragraph 1 (Semi
        CAGR 2024-2033)
    reported_job_openings_ratio = 1.24
    reported_inflation_expectation_5yr_percent = 3.1 # Page 8, text
        point 1
    # Assumptions from question
    baseline_job_ratio = 1.0
    target_inflation_rate_percent = 2.0
    # Calculations
    base_growth_rate_term = base_growth_rate_percent / 100.0
    job_openings_ratio_factor = reported_job_openings_ratio /
        baseline_job_ratio
    # Ensure inflation expectation factor uses rates as percentages or
         decimals consistently
    inflation_expectation_factor =
        reported_inflation_expectation_5yr_percent /
        target_inflation_rate_percent
    # Calculate the index
    index_value = (base_growth_rate_term * job_openings_ratio_factor)
        / inflation_expectation_factor
    # Return final result rounded to four decimal places
    return round(index_value, 4)
```

**Extract:5 Calculate:4 Answer:0.0611**

17

## A.6 Example 6

**Macroeconomics & Fixed Income**

**Question**: Considering the period from December 2024 to January 2025, calculate the ratio of the combined net inflow from Goods Trade and Direct Investment to the combined increase in foreign exchange deposits held by Non-Financial Enterprises and Residents. Report the ratio rounded to three decimal places.

**Keyword:** External Inflow and FX Deposit Analysis

**Python Solution:**

```
def solution():
    # Define variables with their values (in billions USD)
    goods_di_net_inflow = 324.39
    non_fin_ent_deposit_increase = 303.79
    resident_deposit_increase = 156.51
    # Do financial calculation
    combined_deposit_increase = non_fin_ent_deposit_increase +
        resident_deposit_increase
    ratio = goods_di_net_inflow / combined_deposit_increase
    # Round final result to three decimal places
    answer = round(ratio, 3)
    # Return final result
    return answer
```

**Extract:3 Calculate:3 Answer:0.705**

## A.7 Example 7

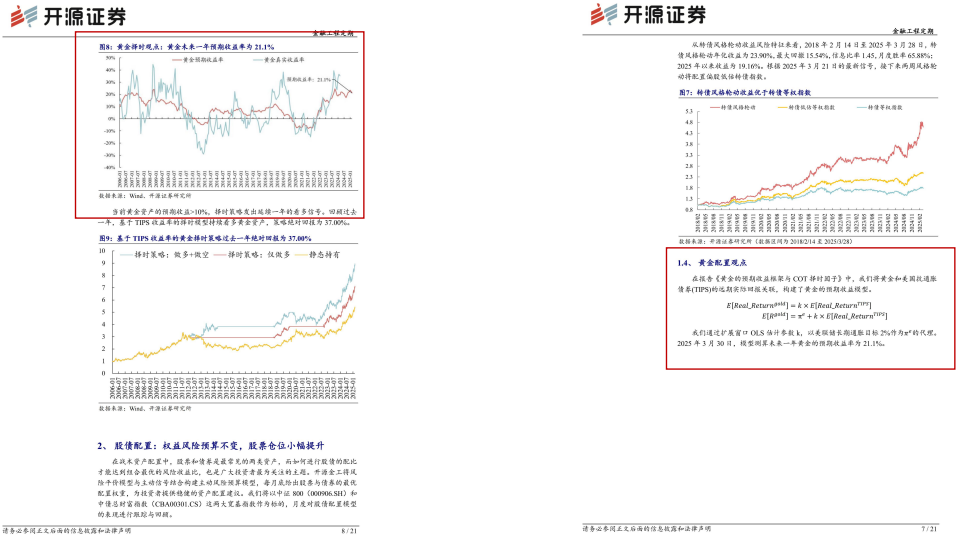

Investment Analysis & Risk Management

**Question**: The report presents a model for the expected return of gold, linking it to expected inflation and the expected real return of US Treasury Inflation-Protected Securities (TIPS). Based on the model's structure described and the specific forecast provided for gold's expected return over the next year as of March 30, 2025, assume that the market's concurrent expectation for the one-year real return on TIPS ($E[\text{Real\_Return}^{\text{TIPS}}]$) was exactly 1.25%. Calculate the implied sensitivity coefficient 'k' used in the model under these conditions. (Round the final answer to two decimal places. The unit is a dimensionless coefficient).
**Keyword:** Gold Return Sensitivity Modeling
**Python Solution:**

```
import numpy as np
def solution():
    # Define variables with their values
    # Expected Gold Return (Page 7, text for Fig 8)
    expected_gold_return = 21.1 / 100 # Convert percentage to decimal
    # Expected Inflation (Proxy used in report, Page 7, Section 1.4)
    expected_inflation = 2.0 / 100 # Convert percentage to decimal
    # Expected TIPS Real Return (Hypothetical value from question)
    expected_tips_real_return = 1.25 / 100 # Convert percentage to
        decimal
    # Formula: E[R^gold] = pi^e + k * E[Real_Return^TIPS]
    # Rearrange to solve for k: k = (E[R^gold] - pi^e) / E[Real_Return
        ^TIPS]
    # Calculate k
    k = (expected_gold_return - expected_inflation) /
        expected_tips_real_return
    # Round final result
    answer = np.round(k, 2)
    # Return final result
    return answer
```

**Extract:2 Calculate:2 Answer:15.28**

19

## A.8 Example 8

**Question**: Consider the recommendation to long the Corn C2507 contract. Suppose that by the time the deep processing enterprise corn inventory-to-consumption ratio declines to 3.0 weeks (hypothetically, based on the trend context shown in Figure 6), the C2507 futures price reaches the exact midpoint of the profit target range provided in the report's single-leg strategy section. Assuming the basis rate for C2507 remains unchanged from its value on April 14th, calculate the percentage increase in the C2507 futures price from its April 14th level required to achieve this target midpoint. Express the answer as a percentage, rounded to two decimal places.

**Keyword:** Agricultural Futures Target Return Analysis

**Python Solution:**

```
def solution():
    # Define variables with their values
    # Initial C2507 futures price from Table on page 14
    initial_futures_price = 2310
    # Profit target range from Table on page 5
    target_min = 2450
    target_max = 2500
    # Calculate the midpoint of the target range
    target_midpoint_price = (target_min + target_max) / 2
    # Calculate the absolute increase in price
    absolute_increase = target_midpoint_price - initial_futures_price
    # Calculate the percentage increase
    percentage_increase = (absolute_increase / initial_futures_price)
        * 100
    # Round the final result to two decimal places
    answer = round(percentage_increase, 2)
    # Return final result
    return answer
```

**Extract:3 Calculate:4 Answer:7.14**

## A.9 Example 9

### Cost Accounting & Management

**Question**: The report projects an increase in the self-sufficiency rate for lithium concentrate for Yahua Group. Assume that for the year 2026, the actual self-sufficiency rate reaches the level projected in the report's relevant figure for that year. Further assume that the unit cost of internally sourced lithium raw materials (feeding into the Cost of Goods Sold for the Lithium segment) is 40% lower than the unit cost of externally purchased lithium raw materials implicitly reflected in the overall 2026 forecast. Calculate the estimated total reduction in the Lithium segment's Cost of Goods Sold for 2026 attributable to achieving this projected self-sufficiency rate, compared to a hypothetical scenario where the self-sufficiency rate was zero (i.e., all lithium raw materials were sourced externally at the higher implicit market cost).

**Keyword:** Lithium Supply Cost Reduction Analysis
**Python Solution:**

```python
def solution():
    # Data Extraction
    # Lithium Segment COGS 2026E from Table 11 (pg 24), unit: hundred
        million yuan
    lithium_cogs_2026e_hm_yuan = 65.24
    lithium_cogs_2026e_b_yuan = lithium_cogs_2026e_hm_yuan / 10.0
    self_sufficiency_rate_2026e = 0.32
    # Assumption: Cost internal = 0.6 * Cost external
    internal_cost_factor = 0.60
    # Calculate the blended cost factor relative to external cost
    blended_cost_factor = (self_sufficiency_rate_2026e *
        internal_cost_factor) + (1 - self_sufficiency_rate_2026e)
    # So, Forecasted COGS = Hypothetical COGS * blended_cost_factor
    if blended_cost_factor == 0:
        hypothetical_cogs_b_yuan = 0
    else:
        hypothetical_cogs_b_yuan = lithium_cogs_2026e_b_yuan /
            blended_cost_factor
    # Calculate Cost Reduction
    cost_reduction_b_yuan = hypothetical_cogs_b_yuan -
        lithium_cogs_2026e_b_yuan
    # Rounding to three decimal places
    answer = round(cost_reduction_b_yuan, 3)
    # Return final result
    return answer
```

**Extract:3 Calculate:6 Answer:22.77**

## A.10 Example 10

<div style="border: 1px solid; padding: 10px;">

<div style="text-align: center; background: #4a7ab8; color: white;">Taxation & Accounting</div>

**Question**: What would be the resulting effect on net income if the federal income tax expense for 2011 was computed based on 2010's federal income tax rate? This rate is assumed to be the ratio of federal income tax expense to income from continuing operations before federal income tax for 2010, and is given in thousands of US dollars. Answer in thousands. Answer to the nearest integer.

**Keyword:** Tax Rate Impact on Net Income

**Python Solution:**

```python
def solution():
    # Define variables name and value based on the given context
    pre_tax_income_2011 = 27772 # in thousands
    federal_income_tax_2011 = 6223 # in thousands
    pre_tax_income_2010 = 7647 # in thousands
    federal_income_tax_2010 = 1054 # in thousands
    # Calculate the effective tax rate for 2010
    tax_rate_2010 = federal_income_tax_2010 / pre_tax_income_2010
    # Calculate what the federal income tax would be in 2011 using
        2010's rate
    hypothetical_tax_2011 = pre_tax_income_2011 * tax_rate_2010
    # Calculate the difference in net income
    tax_difference = federal_income_tax_2011 - hypothetical_tax_2011
    net_income_difference = tax_difference # Higher tax means lower
        net income
    # return answer rounded to nearest thousand
    return round(net_income_difference)
```

**Extract:3 Calculate:6 Answer:32.18**

</div>

## A.11 Example 11

> **Financial Modeling & Projections**
>
> [Figures from 开源证券 report showing lithium segment projections and company business segmentation/profit forecasts]
>
> **Question**: Assume that the global market for AI+AR intelligent glasses evolves according to the sales volume and penetration rate trajectory forecasted in the report. Suppose the average hardware Bill of Materials (BOM) cost for these glasses in 2032 stabilizes at a level 15% higher than the Ray-Ban Meta's reported hardware cost, due to enhanced features. Furthermore, assume the proportional cost contribution of the System-on-Chip (SOC) within the total hardware BOM remains identical to that depicted for the Ray-Ban Meta. Calculate the projected total global market value specifically for the SOC components used in AI+AR intelligent glasses manufactured in the year 2032. Provide the answer in billions of US dollars, rounded to two decimal places.
>
> **Keyword:** AI+AR Glasses SOC Market Sizing
>
> **Python Solution:**
>
> ```python
> def solution():
>     # Define variables with their values
>     # Source: Figure 28, page 23
>     sales_volume_2032_millions = 360 # in millions of units
>     # Source: Text paragraph below Figure 30 on page 24
>     ray_ban_meta_hardware_cost_usd = 164
>     # Source: Figure 31, page 25 (Pie Chart: Ray Ban Meta )
>     soc_cost_percentage = 33.54 / 100
>     # Assumption from question: 15% higher BOM cost in 2032
>     cost_increase_factor = 1.15
>     # Calculate assumed average BOM cost in 2032
>     avg_bom_cost_2032_usd = ray_ban_meta_hardware_cost_usd * 
>         cost_increase_factor
>     # Calculate SOC cost per unit in 2032
>     soc_cost_per_unit_2032_usd = avg_bom_cost_2032_usd * 
>         soc_cost_percentage
>     # Calculate total SOC market value in 2032 (in USD)
>     total_soc_market_value_2032_usd = (sales_volume_2032_millions * 1
>         _000_000) * soc_cost_per_unit_2032_usd
>     total_soc_market_value_2032_billion_usd = 
>         total_soc_market_value_2032_usd / 1_000_000_000
>     return round(total_soc_market_value_2032_billion_usd, 2)
> ```
>
> **Extract:2 Calculate:6 Answer:0.958**

## A.12 Example 12

**Corporate Strategy & Operations**

**Question**: Evaluate the potential cash flow impact from improved inventory management in 2025 based on the report's forecasts. Assume that due to efficiencies gained, particularly from the new BC product lines, the company manages to reduce its inventory holding period (calculated as year-end inventory divided by the forecasted cost of goods sold for the year, then multiplied by 365) by 15 days compared to the holding period implied by the original 2025 projections found in the financial statements. Calculate the resulting positive contribution to cash flow specifically from this reduction in year-end inventory investment during 2025 (round to two decimal places, unit: 100 million yuan).

**Keyword:** Inventory Efficiency Cash Flow Impact

**Python Solution:**

```
def solution():
    # Define variables with their values (in billion Yuan, %)
    revenue_2025 = 874 # billion Yuan
    gross_margin_2025 = 0.104 # 10.4%
    inventory_ye_2025_forecast = 95.1 # billion Yuan
    # Assumption
    inventory_days_reduction = 15 # days
    # Calculate Forecasted COGS for 2025
    cogs_2025 = revenue_2025 * (1 - gross_margin_2025)
    # Calculate Implied Inventory Days from Forecast Data
    implied_days_2025 = (inventory_ye_2025_forecast / cogs_2025) * 365
    # Calculate Target Inventory Days
    target_days_2025 = implied_days_2025 - inventory_days_reduction
    # Calculate Target Year-End Inventory
    target_inventory_ye_2025 = (target_days_2025 / 365) * cogs_2025
    # Calculate Cash Flow Impact (Reduction in Inventory)
    cash_flow_impact = inventory_ye_2025_forecast - 
        target_inventory_ye_2025
    return round(cash_flow_impact, 2)
```

**Extract:4 Calculate:4 Answer:2395**

# B  Examples from Existing Benchmarks

## B.1  Example from FinanceMath

> **One Example from FinanceMath**
>
> **Question_ID**: validation-137
> **Question**: According to the analyst's data within the context of the capital asset pricing model, if the anticipated return for Share B is 11.4% and the risk-free rate is 3%, what is the projected return for the market?

**Context Modalities**: single table(text)

| Security | Expected Standard Deviation(%) | Beta |
|---|---|---|
| Security 1 | 25 | 1.50 |
| Security 2 | 15 | 1.40 |
| Security 3 | 20 | 1.60 |

**NO Multi-Modal Documents Context**

**Real-world Scenario**:
1. Share B is 11.4% and the risk-free rate is 3% **(explicit)**

**Few Explicit Scenarios**
**No Implicit Scenarios**

**Multi-step Computation**:
Extract: 3
Calculate: 1

**Few Extractions and Simple Calculations**

## B.2 Example from FinanceReasoning

> **One Example from FinanceReasoning**
>
> **Question_ID**: test84
> **Question**: If the exchange rate for the Euro (EUR) in London stands at GBP/EUR 0.8878, what would probably be the exchange rate for the British pound (GBP) in Frankfurt (EUR/GBP)? Answer to three decimal places.

**Context Modalities**: no context
### NO Multi-Modal Documents Context

**Real-world Scenario**:
1. If the exchange rate for the Euro (EUR) in London stands at GBP/EUR 0.8878 **(explicit)**
### Few Explicit Scenarios
### No Implicit Scenarios

**Multi-step Computation**:
Extract: 1
Calculate: 1
### Few Extractions and Simple Calculations

## B.3 Example from MME-Finance

> **One Example from MME-Finance**
>
> **Question_ID**: 20
> **Question**: What is the difference between the 60-day moving average and the 5-day moving average on the last day's in the chart.

**Context Modalities**: one image

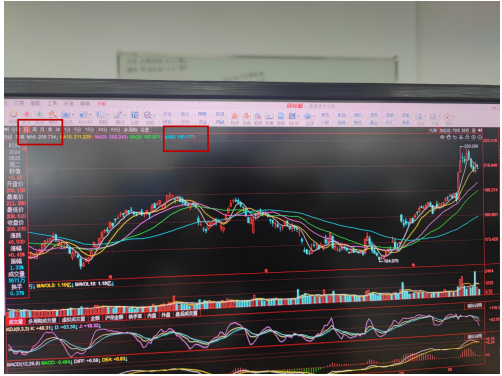

**NO Multi-Modal Documents Context**

**Real-world Scenario**:

**No Scenarios**

**Multi-step Computation**:
Extract: 2
Calculate: 1

**Few Extractions and Simple Calculations**

## B.4 Example from FinMMR

**One Example from FinMMR**

**Question_ID**: easy-test-18
**Question**: What is the total amount of Corporate notes and bonds of 2010 Fair Value, and Net sales of 2011 ?

**Context Modalities**: tow tables(images)

|                      | 2013    | 2012    | 2011    |
|----------------------|---------|---------|---------|
| Net sales            | $8,367  | $8,846  | $9,381  |
| Operating profit     | 759     | 808     | 874     |
| Operating margins    | 9.1%    | 9.1%    | 9.3%    |
| Backlog at year-end  | 8,300   | 8,700   | 9,300   |

|                                       | 2011       |         |                  |            | 2010       |         |                  |            |
|---------------------------------------|------------|---------|------------------|------------|------------|---------|------------------|------------|
|                                       |            |         | Gross Unrealized |            |            |         | Gross Unrealized |            |
|                                       | Fair Value | Amortized Cost | Gains | Losses | Fair Value | Amortized Cost | Gains | Losses |
| Corporate notes and bonds             | $2,032     | $2,024  | $16              | $-8        | $1,133     | $1,124  | $12              | $-3        |
| Commercial paper                      | 1,029      | 1,029   | —                | —          | 1,046      | 1,046   | —                | —          |
| U.S. government and agency securities | 1,021      | 1,018   | 3                | —          | 500        | 501     | 1                | -2         |
| Municipal securities                  | —          | —       | —                | —          | 361        | 359     | 4                | -2         |
| Asset-backed securities               | 292        | 292     | 1                | -1         | 171        | 170     | 1                | —          |
| Mortgage-backed securities            | 223        | 223     | 1                | -1         | 112        | 108     | 5                | -1         |
| Foreign government bonds              | 72         | 72      | —                | —          | 10         | 10      | —                | —          |
| Other debt securities                 | 3          | 1       | 2                | —          | 3          | 1       | 2                | —          |
| Equity securities                     | 397        | 383     | 14               | —          | 321        | 295     | 34               | -8         |
|                                       | $5,069     | $5,042  | $37              | $-10       | $3,657     | $3,614  | $59              | $-16       |

**NO Multi-Modal Documents Context**

**Real-world Scenario**:

**No Scenarios**

**Multi-step Computation**:
Extract: 2
Calculate: 1

**Few Extractions and Simple Calculations**

## B.5 Example from DocMath-Eval

> **One Example from DocMath-Eval**
>
> **Question_ID**: complong-testmini-30
> **Question**: What is the percentage of total offering cost on the total amount raised in the IPO *if the total offering cost is $14,528,328 and each unit sold is $10?*

**Context Modalities**: **texts**
1. Offering costs consist of legal, accounting and other costs incurred through the balance sheet date that are directly related to the Initial Public Offering. Offering costs amounting to $14,528,328 were charged to shareholders' equity upon the completion of the Initial Public Offering.
2. Pursuant to the Initial Public Offering on July 20, 2020, the Company sold 25,300,000 Units, which includes the full exercise by the underwriter of its option to purchase an additional 3,300,000 Units, at a purchase price of $10.00 per Unit. Each Unit consists of one Class A ordinary share and one-half of one redeemable warrant ("Public Warrant"). Each whole Public Warrant entitles the holder to purchase one Class A ordinary share at an exercise price of $11.50 per whole share (see Note 7).

**NO Multi-Modal Documents Context**

**Real-world Scenario**:
1. if the total offering cost is $14,528,328 and each unit sold is $10 **(explicit)**

**Few Explicit Scenarios**
**No Implicit Scenarios**

**Multi-step Computation**:
Extract: 3
Calculate: 2

**Few Extractions and Simple Calculations**

## B.6 Example from SlideVQA

| One Example from SlideVQA |
|---|

**Question_ID**: 1
**Question**: How much difference in INR is there between the average order value of CY2013 and that of CY2012?

**Context Modalities**: Multi-Modal Documents

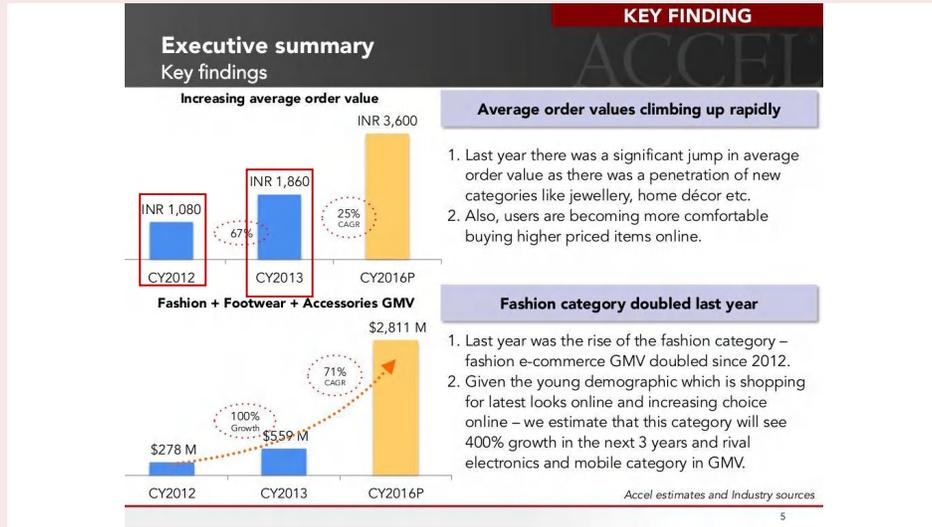

**Multi-Modal Documents Context But Not Cross-Page**

**Real-world Scenario**:

**No Scenarios**

**Multi-step Computation**:
Extract: 2
Calculate: 1

**Few Extractions and Simple Calculations**

## B.7 Example from MMLongBench-Doc

> **One Example from MMLongBench-Doc**
>
> **Question_ID**:
> **Question**: How much higher was the proposed dividend paid (Rupees in lacs) in 2002 compared to 2001?

**Context Modalities**: **Multi-Modal Documents**

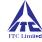

**Multi-Modal Documents Context But Not Cross-Page**

**Real-world Scenario**:

**No Scenarios**

**Multi-step Computation**:
Extract: 2
Calculate: 1

**Few Extractions and Simple Calculations**

## B.8 Example from LongDocURL

> **One Example from LongDocURL**
>
> **Question_ID**: free_gemini15_pro_4061601_47_71_8
> **Question**: What was the total fair value of options that vested in 2016, 2015, and 2014, in millions of Canadian dollars?

**Context Modalities**: **Multi-Modal Documents**

The following table summarizes additional stock option information:

| year ended December 31, (millions of Canadian $, unless otherwise noted) | 2016 | 2015 | 2014 |
|---|---|---|---|
| Total intrinsic value of options exercised | 31 | 10 | 21 |
| Fair value of options that have vested | 126 | 91 | 95 |
| Total options vested | 2.1 million | 2.0 million | 1.7 million |

As at December 31, 2016, the aggregate intrinsic value of the total options exercisable was $86 million and the total intrinsic value of options outstanding was $130 million.

### 21. PREFERRED SHARES

In March 2014, TCPL redeemed all of the 4 million outstanding Series Y preferred shares at a redemption price of $50 per share for a gross payment of $200 million.

### 22. OTHER COMPREHENSIVE (LOSS)/INCOME AND ACCUMULATED OTHER COMPREHENSIVE LOSS

Components of Other comprehensive (loss)/income, including the portion attributable to non-controlling interests and related tax effects, are as follows:

| year ended December 31, 2016 (millions of Canadian $) | Before Tax Amount | Income Tax Recovery/ (Expense) | Net of Tax Amount |
|---|---|---|---|
| Foreign currency translation gains on net investment in foreign operations | 3 | — | 3 |
| Change in fair value of net investment hedges | (14) | 4 | (10) |
| Change in fair value of cash flow hedges | 44 | (14) | 30 |
| Reclassification to net income of gains and losses on cash flow hedges | 71 | (29) | 42 |
| Unrealized actuarial gains and losses on pension and other post-retirement benefit plans | (38) | 12 | (26) |
| Reclassification to net income of actuarial loss on pension and other post-retirement benefit plans | 22 | (6) | 16 |
| Other comprehensive loss on equity investments | (117) | 30 | (87) |
| **Other Comprehensive Loss** | **(29)** | **(3)** | **(32)** |

| year ended December 31, 2015 (millions of Canadian $) | Before Tax Amount | Income Tax Recovery/ (Expense) | Net of Tax Amount |
|---|---|---|---|
| Foreign currency translation gains on net investment in foreign operations | 798 | 15 | 813 |
| Change in fair value of net investment hedges | (505) | 133 | (372) |
| Change in fair value of cash flow hedges | (92) | 35 | (57) |
| Reclassification to net income of gains and losses on cash flow hedges | 144 | (56) | 88 |
| Unrealized actuarial gains and losses on pension and other post-retirement benefit plans | 74 | (23) | 51 |
| Reclassification to net income of actuarial loss and prior service costs on pension and other post-retirement benefit plans | 41 | (9) | 32 |
| Other comprehensive income on equity investments | 62 | (15) | 47 |
| **Other Comprehensive Income** | **522** | **80** | **602** |

155 TCPL Consolidated financial statements 2016

**Multi-Modal Documents Context But Not Cross-Page**

**Real-world Scenario**:

**No Scenarios**

**Multi-step Computation**:
Extract: 3
Calculate: 1

**Few Extractions and Simple Calculations**

# C Benchmark Annotation and Construction

## C.1 Prompts for Question Generation by MLLMs

> **Question Generation Instruction**
>
> ```
> You will receive a financial research report. Based on the content of
>     this report, design 3 English graduate-level questions that are
>     as complex as possible. The difficulty of each question should
>     derive from the following three aspects:
>
> 1. **Numerical Calculation Complexity**: The calculation process must
>      involve multiple steps and should not be solvable with just a few
>      simple calculations.
> 2. **Conceptual Understanding**: Each question must be set in a
>     financial context and should assess the understanding and
>     application of financial terminology and concepts.
> 3. **Data Extraction Difficulty**: The numerical data required to
>     solve the problem must be retrieved from multiple parts of the
>     report. It should not all be found on a single page, within a
>     single chart, or in one paragraph. You are encouraged to
> extract clear data from the chart for problem solving or extract data
>      from the later part of the document for problem solving (to
>     ensure difficulty). Without fabricating inaccurate data from
>     charts without clear data, data sources should include tables,
>     images, and charts as much as possible.
>
> Each question must have **only one numerical answer**. The output
>     must be a **plain number**-**no units, no percent signs**. The
>     question must specify the required units and number of
>     significant digits.
>
> You are allowed to **create reasonable assumptions/hypothetical
>     scenarios** or through other means to enhance complexity. For
>     example, you may introduce cost estimation scenarios by manually
>     setting additional values, or create forecasting questions with
>     assumptions like linear trends, etc.
> Extracting data from research reports is part of the difficulty of
>     the topic. Therefore, in the **question text**, you **must not
>     mention** the specific formulas being tested or the sources/
>     numbers used for the calculations.
> In addition, provide a **detailed solution for each question**, which
>      must include:
> - An explanation that clearly states **where the data came from** (e.
>     g., page number, table/chart, or paragraph reference in the
>     report).
> - A **Python code snippet** that solves the problem, following the
>     format below:
> ```

> ## Question Generation Instruction (Continued)
>
> ```python
> def solution():
>     # Define variables with their values
>     revenue = 600000
>     avg_account_receivable = 50000
>
>     # Do financial calculation
>     receivables_turnover = revenue / avg_account_receivable
>     answer = 365 / receivables_turnover
>
>     # Return final result
>     return answer
> ```
>
>
> ### Example of a High-Quality Question:
>
> Assume that in 2026, VLLC (Weilan Lithium Core) continues to operate based on the forecasted revenue and gross margin data provided in the report. However, a new business structure emerges within the battery segment: the Backup Battery Unit (BBU) accounts for 30% of this segment's revenue, and its gross margin is 10 percentage points higher than the overall battery segment's gross margin provided in the report. The gross margin for the power tool battery subsegment remains unchanged. All other business segments (LED, metal logistics, and ''Others'') maintain the forecasted revenue and gross margin levels from the report. Under these assumptions, calculate the company's total gross profit in 2026** (round to two decimal places, unit: 100 million yuan).

## C.2 Human Annotator Guidelines

The annotation process applies to two sources of questions: (1) 600 newly constructed questions based on Chinese financial documents, and (2) 600 revised questions selected from DocMath-Eval$_{\text{CompLong}}$. All questions follow the same rigorous annotation protocol. Annotators are expected to conduct multi-stage verification and correction to ensure question quality, factual validity, and formal consistency. The process includes:

- **Correctness validation**: Each question must be logically solvable, and the provided Python solution must execute successfully and yield the correct numerical answer. Annotators should fix any solvable issues in either the question wording or the code, and discard only unrepairable cases.
- **Evidence verification**: All numerical values used in the question and solution must either be verifiably extractable from the source document or clearly justified as scenario-based assumptions. Annotators must manually identify all evidence sources, specifying both the document structure (*e.g.*, table, text, chart) and the page number. If multiple data points originate from the same page, that page must be listed repeatedly in the `evidence` field to reflect each distinct usage.
- **Complexity filtering**: Questions that require only trivial calculations (*e.g.*, simple averages over 2–3 numbers) or produce non-numeric answers must be removed or revised to meet the reasoning depth required.
- **Formal consistency**: For all questions, annotators must ensure clarity in expected answer format, including (1) explicit unit specification (*e.g.*, percent, thousands, millions, billions), (2) consistent rounding rules—defaulting to two decimal places when unspecified, and (3) clarity in use of positive/negative signs and numerical ranges.

This unified annotation procedure ensures consistency across both original and revised question sources, and enables the construction of a high-quality benchmark for evaluating multimodal numerical reasoning in real-world financial scenarios.

## C.3 Annotation Statistics

We began with a total of 759 candidate questions. During the annotation phase, 159 questions were discarded due to irreparable issues such as logical inconsistency, unverifiable evidence, or insufficient reasoning complexity.

Of the 600 questions that remained after initial filtering, 494 (82.3%) required manual revision to ensure correctness, evidential traceability, and formal clarity. The distribution of these modifications is as follows:

- **451 questions** had their `evidence` fields revised to accurately align each numerical value with its original source, including precise references to page numbers, tables, charts, or text segments.
- **80 questions** involved corrections to the `solution` code, typically addressing issues such as incorrect formulas, misallocated variables, or computational errors.
- **36 questions** required edits to the `question` text itself, primarily to clarify assumptions, improve phrasing, or enforce consistency in answer format specifications.

## C.4 Examples of Annotation

### C.4.1 Case 1: Discard Directly

> **Example1:0226-2**
>
> [Financial statement forecast table from 民生证券 (Minsheng Securities) report on 钒钛股份 (000629)/金属, showing 利润表, 资产负债表, 现金流量表, and 主要财务指标 data for 2024A, 2025E, 2026E, 2027E]
>
> **Question**: Based on the financial statement forecasts provided in the report, calculate the company's projected Cash Conversion Cycle (CCC) for the fiscal year 2025. Use year-end balances for 2024 and 2025 from the balance sheet forecasts to compute the average balance sheet figures required for the calculation. Assume a 365-day year for ratio calculations (round to one decimal place, unit: days).
> **Evidence Pages**: [13]
> **Ground Truth**: −6.0

**Before**:
evidence:
```
"table": [13],
```

**Discard/Modify**: **Discard**

**Analysis**: In the python_solution of this question, in order to obtain the preset answer of -6.0 days, the accounts payable data for 2025 was artificially adjusted from the original 907 million yuan in the report to 1932 million yuan. This adjustment lacks a reasonable basis and violates the principle of calculation based on original data. Moreover, the adjusted accounts payable has an increase of as high as 235% compared with the 2024 data, which is logically unreasonable. At the same time, the self-calculated DSO is 4.96 days, which is 1.51 days different from the accounts receivable turnover days of 3.45 days in the report, and the DIO is 25.42 days, which is 0.35 days different from the inventory turnover days of 25.07 days in the report. These objectively existing differences have not been reasonably explained, such as failing to indicate whether they are caused by different averaging methods or denominator selections, making the calculation basis unclear. Due to the significant logical errors in the solution, we have chosen to delete this question.

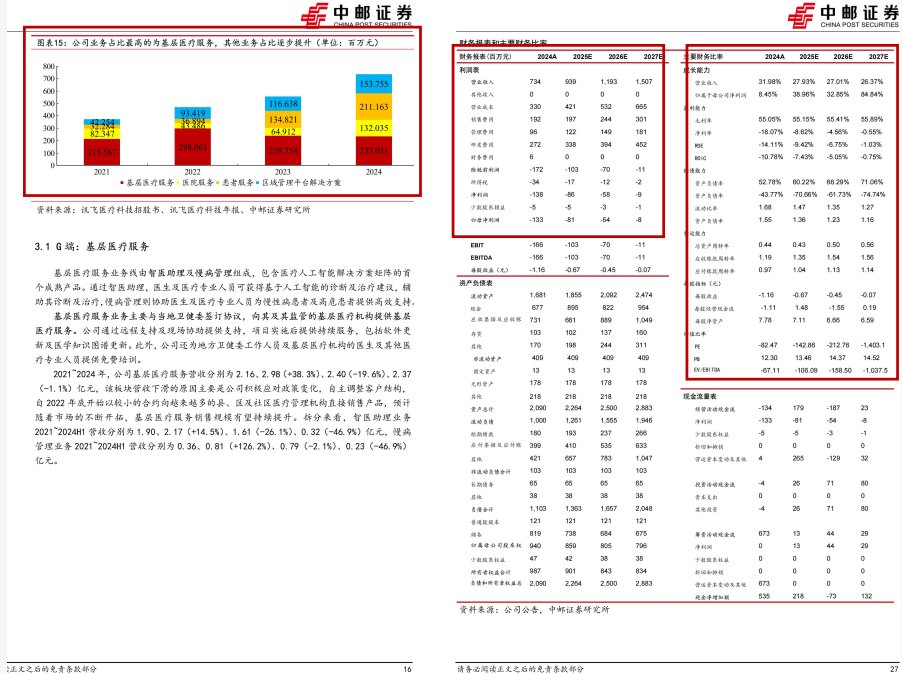

**Question**: Assume that for the year 2026, Xunfei Medical achieves its forecasted total revenue. However, due to intensified competition specifically impacting its hospital-related offerings, the gross margin for the 'Hospital Services' segment drops significantly. Assume the revenue contribution percentages for each of the four main business segments ('Basic Medical Services', 'Hospital Services', 'Patient Services', 'Regional Management Platform Solutions') in 2026 are identical to their respective contributions in the actual results of 2024. Furthermore, assume the gross margin for the 'Hospital Services' segment in 2026 becomes exactly half of the company's overall gross margin reported for 2024 (use the value from the Financial Ratios table). If the company still manages to achieve the overall gross profit margin forecasted for 2026 (use the value from the Financial Ratios table) through adjustments in the profitability of its other three segments combined, calculate the implied weighted average gross margin required for the combination of 'Basic Medical Services', 'Patient Services', and 'Regional Management Platform Solutions' segments in 2026 (round to two decimal places, report the number only, representing percentage points e.g., 65.43 for 65.43%)
**Evidence Pages**: [16,27]
**Ground Truth**: 61.21

**Before**:
evidence:
"table": [27],
"chart": [16]

**Discard/Modify**: Discard

**Analysis**: A key issue in the python_solution is the lack of basis for handling data inconsistencies. Specifically, the adopted projected overall gross profit margin of 55.15% for 2026 (from the financial ratio table in Table 27) is explicitly marked as "inconsistent with the Income Statement (I/S) data", yet no specific difference value is explained, nor are the reasons for the inconsistency verified (such as differences in statistical standards or calculation methods). It simply forcibly selects the ratio table data based on assumptions. This approach directly affects the calculation basis — if the actual gross profit margin for 2026 in the Income Statement is another value, the total gross profit will deviate from the currently calculated 6.579395 million yuan, which in turn leads to the result of 61.21% for the "required gross profit margin of other segments" lacking a rigorous data foundation. For these reasons, we have chosen to abandon this question.

### C.4.2 Case 2: Modify Evidence

---

**Example1: test-146**

**Question**: Assume that in 2025, HYYZ Pharma's 'Product Sales' segment achieves its forecasted revenue, but the internal revenue composition mirrors the domestic product sales breakdown observed in the first three quarters of 2024 (requires estimation from the relevant chart). Under a hypothetical performance scenario, the gross margin for Oxaliplatin within this segment is 5.0 percentage points higher than the segment's overall forecasted gross margin for 2025, while the gross margin for Pemetrexed is 10.0 percentage points lower. All other products grouped within this segment collectively achieve the segment's originally forecasted gross margin. The 'Technical Services' and 'Overseas Revenue' segments perform exactly as forecasted in the report's key assumptions table, achieving their projected revenues and gross margins. Calculate the company's total adjusted gross profit for 2025 under these specific conditions (unit: 100 million yuan, round to two decimal places).
**Evidence Pages**: [6, 26, 27]
**Ground Truth**: 10.71

---

**Before**:
**evidence:**
```
"table": [26, 26, 26],
"plain_text": [26, 26]
```

**Discard/Modify**: Modify

**After**:
**evidence**:
```
"table": [26, 26, 26, 26, 27],
"plain_text": [26, 26, 26, 6],
"pie_chart": [6, 6]
```

**Analysis**: The initial annotation for this multi-step financial problem was critically insufficient. By identifying only page 26, it provided incomplete forecast data and omitted two essential pieces of evidence explicitly required by the question: the sales composition chart from page 6 for revenue allocation and the complete forecast table from page 27. The correction rectifies this by incorporating both the chart and the additional table, thereby supplying the complete set of numerical inputs—proportions from the chart and base values from the tables—necessary for the calculation. This case highlights the importance of comprehensively parsing the question to identify all required evidence, including data from non-tabular sources.

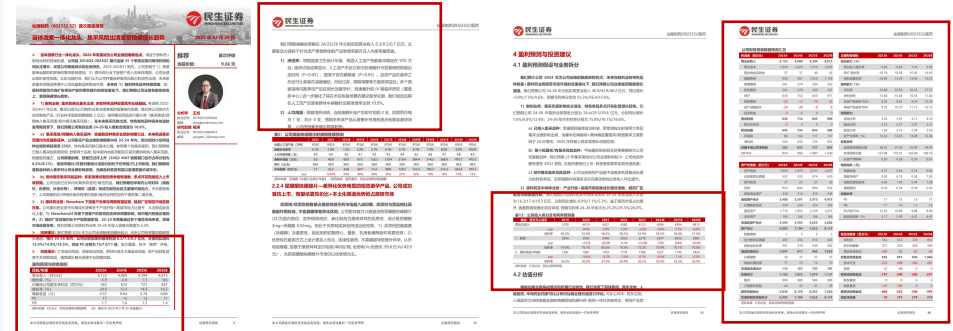

**Question**: Evaluate the potential impact of raw material cost volatility, identified as a key risk factor, on the company's valuation in 2026. Assume that increased costs cause the actual gross margin for the API & Intermediates segment in 2026 to be 2.0 percentage points lower than the figure forecasted in the report's segment analysis table. Assume all other revenue forecasts and the gross margin for the Formulation segment remain unchanged from the report's forecasts. Furthermore, assume this reduction in gross profit directly reduces the company's profit before tax, and the company's effective income tax rate for 2026 is consistent with the rate implied by the report's forecasted Profit Statement figures. Assume the entire impact of the tax-adjusted profit change affects the net profit attributable to parent company shareholders. If the company's stock trades exactly at the forecasted Price-to-Earnings (P/E) multiple for 2026 provided in the report's main forecast table, calculate the new implied share price (round to two decimal places, unit: yuan).
**Evidence Pages**: [1, 11, 21, 24]
**Ground Truth**: 9.07

**Before**:
evidence:
`"table": [1, 21, 24]`

**Discard/Modify**: Modify

**After**:
evidence:
`"table": [1, 21, 21, 24, 24, 24, 24],`
`"plain_text": [21, 11]`

**Analysis**: This complex valuation problem requires a chain of calculations, including a tax-impact adjustment. The initial annotation was deficient because it failed to provide a robust evidence set for this multi-step process. While it correctly identified the main pages for the PE multiple (p. 1), segment data (p. 21), and profit statement (p. 24), it lacked the necessary reinforcement and contextual support. The calculation of the effective tax rate—a critical intermediate step—relies on precise figures from the profit statement, and the original sparse evidence was insufficient. The modification addresses this by heavily reinforcing the key data tables on pages 21 and 24 and adding textual context from page 11. This creates a more resilient evidence chain, ensuring all necessary inputs for the tax-adjusted profit calculation and final valuation are securely retrieved.

### C.4.3 Case 3: Modify Question

> **Example1: test-160**
>
> [Figure: Three screenshots of a Chinese financial research report showing fund data tables and charts, with red highlighted boxes marking key evidence.]
>
> **Question**: An investor made a one-time investment of 5,000,000 HKD into the Taikang CSI HK Connect Comprehensive Consumption Thematic Index Fund A (006786.OF) precisely on its inception date. Assuming the fund perfectly tracked the gross annualized return of its benchmark index (as reported for the period 2017-01-01 to 2025-03-11) before considering any fund operating expenses, and incurred its stated annual management and custody fees consistently throughout the period ending 2025-03-11, calculate the total value of this investment on 2025-03-11. Use 365.25 days per year in your calculations. (Round the final answer to the nearest whole number, unit: HKD).
> **Evidence Pages**: [8,9,12]
> **Ground Truth**: 6415775

> **Before**:
> question: ...calculate the total value of this investment on 2025-03-11. (Round the final answer to the nearest whole number, unit: HKD).

> **Discard/Modify**: Modify

> **After**:
> question:...calculate the total value of this investment on 2025-03-11. **Use 365.25 days per year in your calculations.** (Round the final answer to the nearest whole number, unit: HKD).

> **Analysis**: This case involves an ambiguity in the question's time basis specification, which affects the accuracy of compound interest calculations over a multi-year period. The original question omitted the instruction to use 365.25 days per year, which is standard when spanning multiple calendar years and accounting for leap years. In the absence of this guidance, solvers might default to 365 or apply inconsistent time bases, leading to small but compounding deviations over an 8+ year horizon. Although the numerical gap is subtle, it can produce discrepancies that prevent matching the ground truth. By explicitly adding "Use 365.25 days per year in your calculations," the revised question ensures uniform interpretation, eliminates source ambiguity, and aligns the question with professional finance norms—thus restoring both solvability and fidelity to the intended answer.

**Example2: test-230**

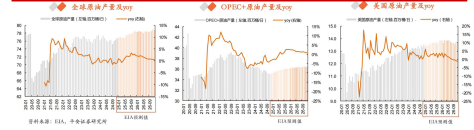
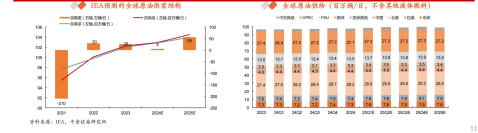

**Question**: Based on the report's forecasts from the EIA and IEA, calculate the projected difference in the cumulative increase in global crude oil and related liquid fuels supply attributed solely to Non-OPEC+ nations over the entire calendar year 2025. Express this difference as an absolute value in Terawatt-hours (TWh), assuming an average energy density of 1.63 Megawatt-hours per barrel for the incremental supply. (Round your final answer to three significant figures).
**Evidence Pages**: [9, 11]
**Ground Truth**: 107.091

---

**Before**:
question: ...*Express this difference as an absolute value in **billions of Terawatt-hours (TWh)***...
ground_truth: 0.107

**Discard/Modify**: Modify

**After**:
question: ...*Express this difference as an absolute value in **Terawatt-hours (TWh)***...
ground_truth: 107.091

**Analysis**: This case exemplifies an error rooted in the question's formulation. The original question required the final answer in "billions of TWh." This was an unnatural unit constraint that created a severe logical conflict: the correctly calculated physical value (107.091 TWh) did not align with the provided ground truth (0.107) under any valid mathematical conversion for "billions." This flawed instruction not only made the answer unintuitive but, more critically, induced a "hallucinated" and convoluted reasoning process to bridge the gap. The modification resolves this by removing the problematic "billions of" requirement. This simple change makes the question direct, aligns the expected answer with its natural scale, and eliminates the source of logical contradiction.

### C.4.4 Case 4: Modify Python Solution

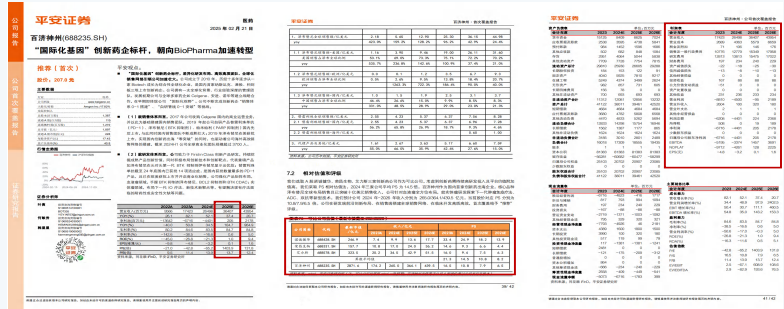

Example1: test-33

**Question**: Assuming the market values pharmaceutical companies partly based on their research intensity relative to peers, estimate BeiGene's implied market capitalization for 2025. Start with the average 2025E Price-to-Sales (P/S) ratio of the comparable companies listed in the report. Adjust this peer average P/S ratio by adding a premium calculated as follows: Premium = (BeiGene's forecasted 2025 R&D Intensity - 30%) * 5.0, where R&D Intensity is defined as R&D expenses divided by revenue. Use the resulting adjusted P/S ratio and BeiGene's forecasted 2025 revenue to find the implied market capitalization. Provide the answer rounded to the nearest whole number (unit: 100 million yuan).
**Evidence Pages**: [1, 39, 41]
**Ground Truth**: 3104

**Before**:
solution:

```
def solution():
    # --- Data Extraction ---
    peer_avg_ps_2025e = 9.9
    # --- Assumptions from Question ---
    # --- Calculations ---
    implied_market_cap_hundred_million_rmb = ...
    return round(implied_market_cap_hundred_million_rmb,2)
```

**Discard/Modify**: Modify

**After**:
solution:

```
def solution():
    # --- Data Extraction ---
    peer_avg_ps_2025e = 7.9
    # --- Assumptions from Question ---
    # --- Calculations ---
    implied_market_cap_hundred_million_rmb = ...
    return round(implied_market_cap_hundred_million_rmb,2)
```

**Analysis**: This case exemplifies a fundamental numerical extraction error arising during table parsing. While the source document (evidence page [39]) correctly reports the peer average price-to-sales (P/S) ratio as 7.9, the implemented Python solution erroneously extracted a value of 9.9.
Unlike errors attributable to logic flaws or rounding discrepancies, the root cause of this issue lies in the model's failure to accurately identify numerical values within the document. Such errors are characteristic of document-based reasoning systems, where complex table structures or inconsistent number formatting can mislead automated extraction processes.
Given the requirement to strictly utilize reported figures, the appropriate remediation involves correcting the extracted P/S ratio to align with the original report. This ensures that all downstream calculations accurately reflect the underlying financial data and preserve the integrity of the analysis.

> **Example2: test-236**
>
> **Question**: Jiangzhong Pharmaceutical's stock incentive plan (revised draft, 2025, mentioned on page 6) sets specific performance targets. Using the analyst's detailed forecasts presented in the report, first calculate the company's projected Return on Invested Capital (ROIC) for the year 2025 based on standard definitions, where Invested Capital is the average of beginning and end-of-year Total Equity plus Interest-Bearing Debt (defined as Short-Term plus Long-Term Borrowings), and NOPAT is Operating Profit multiplied by (1 - Effective Tax Rate), with the Effective Tax Rate derived from the 2025 forecast. Now, consider a hypothetical scenario where the company, through efficiency improvements, manages to decrease its Management Expenses by 8% and its Sales Expenses by 3% in 2025 compared to their forecasted absolute values, while Revenue and all other costs (including COGS, R&D, Taxes & Surcharges, Finance Costs) and balance sheet items remain exactly as forecasted. Calculate the new ROIC under this expense reduction scenario (result in percentage, round to two decimal places).
> **Evidence Pages**: [20]
> **Ground Truth**: 19.53

**Before**:
solution:

```
def solution():
    # --- Data Extraction ---
    # --- Assumptions from Question ---
    # --- Calculations ---
    # OpProfit = Rev - COGS - TaxSurch - Sales - Mgmt - R&D - Fin + ...
    sales_exp_2025e_implied = left_side - operating_profit_2025e
    adjusted_roic_2025e = ...
    return round(adjusted_roic_2025e * 100, 2)
```

**Discard/Modify**: Modify

**After**:
solution:

```
def solution():
    # --- Data Extraction ---
    sales_expense_2025 = 1689
    # --- Assumptions from Question ---
    # --- Calculations ---
    adjusted_roic_2025e = ...
    return round(adjusted_roic_2025e * 100, 2)
```

**Analysis**: This case highlights deriving the "Sales Expenses" figure indirectly—by subtracting other metrics—instead of directly retrieving the clearly stated value. In the report, the 2025 sales expense is unambiguously stated as 1,689 million; this figure should be read directly rather than inferred via a residual calculation. Such unnecessary recomputation invites logic errors and contravenes the principle of relying on primary data fields. The solution should be revised to prioritize direct extraction of documented figures, thereby avoiding reliance on unstable intermediate quantities.

### C.4.5 Case 5: Retain

| Example1:0218-1 |
|---|



**Question**: Assume that in 2026, the company's "Precision Reducer (Harmonic Reducer) & Components" business segment experiences a structural shift. 40% of this segment's forecasted revenue is derived from a new "Advanced Robot Drives" sub-segment, which achieves a gross margin 15 percentage points higher than the overall segment's forecasted 2026 gross margin. The remaining 60% of the segment's revenue comes from "Standard Reducers", which realizes a gross margin 5 percentage points lower than the overall segment's forecasted 2026 gross margin. All other business segments achieve their revenue and gross margin forecasts as presented in the report for 2026. Calculate the company's total gross profit in 2026 under this scenario (round to two decimal places, unit: million yuan).
**Evidence Pages**: [21]
**Ground Truth**: 218.42

**Discard/Modify**: Retain

**Analysis**:
**Topic Thinking:** Adjust the revenue and gross profit margin of the "Precision Reducers (Harmonic Reducers) and Components" business segment to calculate the company's total gross profit in 2026, examining the understanding of gross profit calculation under changes in business structure.
**Formula and Logic Check:** The gross profit of each business segment is calculated using the standard formula "Revenue × Gross Profit Margin"; The breakdown and gross profit margin adjustment calculation for the "Precision Reducers (Harmonic Reducers) and Components" business segment are reasonable; The total gross profit is calculated as the sum of the gross profits of each segment, with no logical errors in the calculation.
**Conclusion:** The calculation steps are logically sound, but the data accuracy needs to be verified against the document.
**Complexity:** Moderate, involving data processing for multiple business segments and gross profit margin adjustment calculations.
**Rigor:** The topic clearly requires the answer to be retained to two decimal places and the unit "ten million yuan" to be provided, with a standardized format. However, it is necessary to confirm the accuracy and consistency of the data and gross profit margin of each business segment in the document.

<div style="border:1px solid #4A90D9; padding:10px;">

**Example1:0231-1**

国盛证券  2025 年 04 月 09 日

**5. 盈利预测与估值建议**

**5.1. 关键假设**

**岩土工程：** 受益于核电、港口、水利水电及跨江跨海桥隧等关键领域的订单稳定增长，随着公司在手订单的持续兑现以及地产基建行业景气度的复苏，预计公司岩土工程板块营收将持续提升，同时高附加值基建项目的开拓有望提升岩土工程整体盈利能力；

**销售产品：** 销售产品在公司营收占比较低，伴随未来岩土工程项目额的增长，预计销售产品板块营收稳定增长，毛利率维持稳定；

**环境修复：** 环境修复在公司营收占比较低，预计环境修复板块营收、毛利率维持稳定。

**5.2. 盈利预测**

<span style="color:red">若不考虑西部水电业绩增量</span>，我们预计公司 2024-2026 年营收分别为 7.76/14.92/23.34 亿元；归母净利润分别为 0.69/1.62/3.32 亿元；对应 PE 分别为 61.4/26.2/12.8 倍。

图表32：公司各业务营收拆分（亿元）

| 报告期 | 2022A | 2023A | 2024E | 2025E | 2026E |
|---|---|---|---|---|---|
| 岩土工程 | | | | | |
| 收入 | 8.05 | 7.00 | 6.16 | 13.10 | 21.30 |
| YOY | | -13% | -12% | 113% | 63% |
| 毛利率 | 3% | 18% | 29% | 34% | 35% |
| 销售产品 | | | | | |
| 收入 | 0.51 | 1.78 | 1.20 | 1.40 | 1.60 |
| YOY | | 249% | -33% | 17% | 14% |
| 毛利率 | -2% | 8% | 12% | 12% | 12% |
| 环境修复 | | | | | |
| 收入 | 0.37 | 0.34 | 0.36 | 0.38 | 0.40 |
| YOY | | -8% | 6% | 6% | 5% |
| 毛利率 | 19% | 32% | 32% | 32% | 32% |
| 营业总收入 | | | | | |
| 收入 | 8.97 | 9.16 | 7.76 | 14.92 | 23.34 |
| YOY | | 2% | -15% | 92% | 56% |
| 毛利率 | 4% | 17% | 26% | 32% | 34% |

资料来源：Wind，国盛证券研究所

**5.3. 估值分析与投资建议**

我们选取岩土工程公司上海港湾、民爆材料及工程公司高争民爆、易普力作为可比公司。可比公司 2024-2026 年平均 PE 分别为 32.43/21.79/15.67 倍，公司是岩土工程龙头，未来自其他国岩土先驱的传承，核电、水电、港口桥梁三驾马车，迎来历史机遇，公司 2024-2026 年 PE 分别为 61.4/26.2/12.8 倍，维持"买入"评级。

P.25  请仔细阅读本报告末页声明

**Question**: Consider the company's financial forecast for 2026. Assume a strategic shift occurs within the primary "Geotechnical Engineering" business segment. Specifically, assume that projects related to Nuclear Power constitute 40% of this segment's forecasted revenue for 2026, and that the gross margin for these Nuclear Power projects is 5 percentage points higher than the overall gross margin forecasted for the entire Geotechnical Engineering segment in the report. The remaining 60% of the Geotechnical Engineering revenue retains the original forecasted gross margin for that segment. Furthermore, assume the "Sales of Products" segment experiences stronger-than-anticipated demand, resulting in its actual revenue being 15% higher than forecasted for 2026, while its gross margin remains as forecasted. All other segments (Environmental Remediation) meet their forecasted revenue and gross margin exactly. Calculate the company's revised total gross profit for 2026 under these conditions (round to two decimal places, unit: 100 million yuan).
**Evidence Pages**: [25]
**Ground Truth**: 8.23

</div>

**Discard/Modify**: Retain

**Analysis**:
**Topic Thinking:** Subdivide and calculate the gross profit of the "Geotechnical Engineering" business department, adjust the revenue of the "Product Sales" department and calculate its gross profit, and combine with the situation of the "Environmental Restoration" department to compute the revised total gross profit, examining the ability of financial data adjustment and gross profit calculation.
**Formula and Logic Check:** Calculate the revenue, gross profit margin, and gross profit of nuclear power projects and non-nuclear power projects in the "Geotechnical Engineering" business department, with correct logic; Calculate the revenue and gross profit of the revised "Product Sales" department, as well as the gross profit of the "Environmental Restoration" department, with reasonable formula application; Calculate the revised total gross profit, with clear steps and correct formula use.
**Conclusion:** The calculation steps are logically correct, but the data accuracy needs to be verified against the document.
**Complexity:** Moderate, involving financial data processing and calculation for multiple business departments.
**Rigor:** The topic clearly requires the answer to be retained to two decimal places and the unit "100 million yuan" to be provided, with a standardized format. However, it is necessary to confirm the accuracy and consistency of each financial data in the document.

# D Experiments Setting

## D.1 Input Processing Strategy

To preprocess page-level image data for model input, we merge multiple images from each document into concatenated images under a unified strategy. The merging process follows several rules.

If the number of images in a document is below a predefined threshold, no merging is performed. Instead, all images are directly copied to the target directory without modification. When the number of images exceeds the threshold, merging is applied. The images are split into multiple groups, with each group combined into a single merged image. This ensures that no merged output contains more than the threshold number of individual images.

For documents from the DocMath-Eval dataset, identified by specific naming patterns, images are stacked vertically in a single column. This layout is chosen because images in this dataset are typically wide, and vertical stacking helps maintain a more reasonable aspect ratio.

In contrast, for all other documents requiring merging, images are arranged in a grid layout, aiming to balance the aspect ratio and reduce the overall height of the merged image. The number of rows is adjusted accordingly based on the total number of images in each group.

| Condition | Setting |
| --- | --- |
| Image count < threshold | No merge; copy only |
| Image count >= threshold & DocMath-Eval | Vertical stack, 1 column (2 for special case) |
| Image count >= threshold (others) | Grid layout; column count = ceil(N / threshold) |
| Max images per merged output | <= threshold (e.g., 50) |

For text input, we use five length settings: full-length (no truncation), and truncated versions at 200k, 128k, 96k, and 64k tokens.

## D.2 Prompt Configurations

### D.2.1 Prompts for Image-Based Tasks

> **Prompts for Image-Based Tasks**
>
> SYSTEM_INPUT = You are a financial expert, you are supposed to
>     generate a Python program to answer the given question based on
>     the provided financial document images. The returned value of the
>      program is supposed to be the answer.
> ```python
> def solution():
>     # Define variables name and value based on the given context
>     guarantees = 210
>     total_exposure = 716
>
>     # Do math calculation to get the answer
>     answer = (guarantees / total_exposure) * 100
>
>     # return answer
>     return answer
> ```'''
> USER_INPUT = '''Question:
> {financial_question}
>
> Please generate a Python program to answer the given question. The
>     format of the program should be the following:
> ```python
> def solution():
>     # Define variables name and value based on the given context
>     ...
>     # Do math calculation to get the answer
>     ...
>     # return answer
>     return answer
> ```
>
> Continue the program to answer the question. The returned value of
>     the program is supposed to be the answer:
> ```python
> def solution():
>     # Define variables name and value based on the given context
> ```
>
> The images of the financial document are as follows:
> {financial_document_images}
> '''

### D.2.2 Prompts for Text-Based Tasks

> **Prompts for Text-Based Tasks**
>
> ```
> SYSTEM_INPUT = '''You are a financial expert. You must generate a
>     Python program to answer the given question based on the provided
>     financial document context. The program must return the answer.
>
> ```python
> def solution():
>     # Define variables name and value based on the given context
>     guarantees = 210
>     total_exposure = 716
>
>     # Do math calculation to get the answer
>     answer = (guarantees / total_exposure) * 100
>
>     return answer
> '''
>
> USER_INPUT = '''Question:
> {financial_question}
>
> Please generate a Python program to answer the given question. The
>     required
> format is:
> def solution():
>     # Define variables name and value based on the given context
>     ...
>     # Do math calculation to get the answer
>     ...
>     return answer
> Continue the program to answer the question. The returned value of
>     the program
> must be the answer.
>
> The context of the financial document is as follows:
> {financial_document_context}
> '''
> ```

### D.2.3 Prompts for Answer Extraction

<div style="border: 1px solid blue; padding: 10px;">

**Question Generation Instruction**

```
SYSTEM_INPUT = '''You are a financial expert. Your task is to extract
    the
final numeric answer to a question from a chain-of-thought (CoT)
    solution.
Follow these rules carefully:

1. Read the entire solution and identify the **last numeric value
    that the
   reasoning presents as the answer** (this may appear after an equal
       sign,
   ``~='', or in the closing sentence).
2. Accept integers, decimals, or scientific notation. Remove any
    commas,
   currency symbols, percent signs, or units.
3. If the solution contains multiple candidate numbers, choose the
    one
   explicitly indicated as the final answer; otherwise choose the **
       last
   numeric value** in the text.
4. If you **cannot confidently locate** such a number, output exactly
    the
   string ``None'' (without quotation marks).
5. Output **only** the number or ``None'' - no additional explanation,
    text,
   or punctuation.'''

USER_INPUT = '''Given a financial question and its chain-of-thought
    solution,
return the final numeric answer following the above rules. If no
    clear numeric
answer exists, respond with ``None''.

Question:
{financial_question}

Solution:
{cot_solution}
'''
```

</div>

### D.3 Experimental Environment

For evaluation experiments with two input settings, inference for most models was conducted via OpenRouter APIs. The only exceptions are the Doubao-1.5-thinking-pro and Doubao-1.5-vision-pro models, which were accessed through Volcano Engine APIs.

All retrieval-related evaluations for RAG models were executed on a local server, while the answer generation and inference stages were handled via external APIs.

The retrieval components of all RAG frameworks ran on the same local hardware setup. The system configuration is summarized below:

- **CPU:** Dual-socket Intel Xeon Gold 6148 (2.40 GHz), 40 cores per socket, 80 threads total
- **GPU:** 8× NVIDIA A40, each with 48 GB VRAM
- **GPU driver:** 525.125.06, **CUDA:** 11.8
- **cuDNN:** 8.x (compiled with CUDA 11.8)
- **Operating System:** Ubuntu 20.04.6 LTS

# E Detailed Evaluation Results on FinMMDocR

## E.1 Performance by Scenario Type and Count

| Model | Scenario Count | | | | Scenario Type | | | | | | | | | | | |
|---|---|---|---|---|---|---|---|---|---|---|---|---|---|---|---|---|
| | 1 | 2 | 3 | 4 | CE&RA | FM&P | CA&M | CS&O | M&IA | T&A | M&FI | CF&CM | FSA | A&EV | PM | IA&RM |
| **MLLM (Image Input)** | | | | | | | | | | | | | | | | |
| *Proprietary MLLMs* | | | | | | | | | | | | | | | | |
| OpenAI o4-mini-high | 63.60 | 52.50 | 55.15 | 47.68 | 61.22 | 37.10 | 38.46 | 45.45 | 58.62 | 54.17 | 52.50 | 55.11 | 57.49 | 50.00 | 55.20 | 54.17 |
| Doubao-1.5-thinking-pro | 41.60 | 40.50 | 40.21 | 33.77 | 48.98 | 24.19 | 26.92 | 36.36 | 37.93 | 37.50 | 45.00 | 40.91 | 38.16 | 30.65 | 40.80 | 52.78 |
| Claude 3.7 Sonnet (Thinking) | 38.80 | 35.00 | 35.05 | 31.79 | 57.14 | 25.81 | 19.23 | 25.45 | 31.03 | 37.50 | 43.75 | 29.55 | 33.82 | 32.26 | 40.80 | 31.94 |
| Doubao-1.5-vision-pro | 30.80 | 29.50 | 30.41 | 22.52 | 34.69 | 19.35 | 15.38 | 25.45 | 27.59 | 20.83 | 28.75 | 27.27 | 28.99 | 29.84 | 24.00 | 38.89 |
| Gemini 2.5 Pro Preview | 31.20 | 25.00 | 33.51 | 19.21 | 44.90 | 17.74 | 15.38 | 27.27 | 32.18 | 25.00 | 31.25 | 26.70 | 26.09 | 16.94 | 24.80 | 27.78 |
| GPT-4o | 22.00 | 7.50 | 7.73 | 7.95 | 8.16 | 4.84 | 0.00 | 9.09 | 8.05 | 16.67 | 5.00 | 17.61 | 14.98 | 3.23 | 8.80 | 18.06 |
| Grok 2 Vision | 5.20 | 1.00 | 2.06 | 1.32 | 6.12 | 1.61 | 0.00 | 3.64 | 2.30 | 0.00 | 0.00 | 3.41 | 3.86 | 0.81 | 0.00 | 4.17 |
| *Open-Source MLLMs* | | | | | | | | | | | | | | | | |
| Qwen2.5-VL 72B | 16.00 | 8.50 | 10.31 | 4.64 | 10.20 | 8.06 | 0.00 | 5.45 | 12.64 | 8.33 | 10.00 | 10.23 | 12.08 | 3.23 | 8.80 | 12.50 |
| Llama 4 Maveric | 6.80 | 3.00 | 2.58 | 0.66 | 4.08 | 0.00 | 0.00 | 7.27 | 2.30 | 0.00 | 3.75 | 5.68 | 3.38 | 1.61 | 0.80 | 5.56 |
| Mistral Small 3.1 | 2.80 | 1.00 | 1.55 | 0.00 | 0.00 | 0.00 | 0.00 | 1.82 | 0.00 | 0.00 | 0.00 | 1.70 | 2.42 | 1.61 | 0.80 | 2.78 |
| Gemma3 27B | 2.40 | 0.50 | 0.52 | 0.00 | 0.00 | 0.00 | 0.00 | 1.82 | 0.00 | 0.00 | 0.00 | 1.14 | 1.93 | 0.81 | 0.00 | 0.00 |
| **OCR + LLM (Text Input)** | | | | | | | | | | | | | | | | |
| *Proprietary LLMs* | | | | | | | | | | | | | | | | |
| Gemini 2.5 Pro Preview | 58.40 | 53.50 | 56.19 | 50.99 | 69.39 | 40.32 | 34.62 | 43.64 | 47.13 | 45.83 | 50.00 | 53.98 | 61.35 | 55.65 | 49.60 | 59.72 |
| Claude 3.7 Sonnet (Thinking) | 50.40 | 53.00 | 45.36 | 44.37 | 63.27 | 37.10 | 26.92 | 36.36 | 40.23 | 41.67 | 51.25 | 43.75 | 55.56 | 45.97 | 45.60 | 45.83 |
| OpenAI o4-mini-high | 55.60 | 48.00 | 53.61 | 43.71 | 61.22 | 30.65 | 26.92 | 43.64 | 40.23 | 45.83 | 47.50 | 53.41 | 54.11 | 48.39 | 45.60 | 52.78 |
| Doubao-1.5-thinking-pro | 51.20 | 44.00 | 40.21 | 34.44 | 53.06 | 25.81 | 23.08 | 43.64 | 33.33 | 41.67 | 48.75 | 43.75 | 47.34 | 34.68 | 38.40 | 38.89 |
| Grok 3 | 44.40 | 38.00 | 38.66 | 37.75 | 46.94 | 25.81 | 23.08 | 32.73 | 29.89 | 25.00 | 41.25 | 46.02 | 44.93 | 32.26 | 36.00 | 44.44 |
| Doubao-1.5-vision-pro | 39.60 | 28.00 | 31.44 | 23.84 | 34.69 | 22.58 | 23.08 | 27.27 | 22.99 | 25.00 | 38.75 | 32.95 | 31.88 | 21.77 | 29.60 | 36.11 |
| GPT-4o | 28.80 | 13.50 | 17.53 | 13.25 | 22.45 | 8.06 | 7.69 | 16.36 | 16.09 | 25.00 | 21.25 | 19.89 | 18.84 | 17.74 | 12.80 | 23.61 |
| *Open-Source LLMs* | | | | | | | | | | | | | | | | |
| DeepSeek-R1 | 50.40 | 35.00 | 41.75 | 35.10 | 51.02 | 30.65 | 26.92 | 40.00 | 36.78 | 41.67 | 43.75 | 44.89 | 45.89 | 34.68 | 33.60 | 43.06 |
| DeepSeek-V3 | 42.40 | 23.50 | 26.80 | 25.17 | 34.69 | 19.35 | 19.23 | 30.91 | 26.44 | 33.33 | 30.00 | 32.39 | 28.02 | 21.77 | 26.40 | 36.11 |
| Llama 4 Maverick | 34.00 | 26.00 | 23.20 | 23.18 | 34.69 | 17.74 | 15.38 | 23.64 | 22.99 | 25.00 | 28.75 | 25.57 | 25.60 | 22.58 | 26.40 | 30.56 |
| Qwen3 | 32.00 | 18.00 | 15.46 | 15.23 | 22.45 | 11.29 | 3.85 | 16.36 | 17.24 | 16.67 | 21.25 | 21.59 | 21.26 | 12.10 | 20.80 | 26.39 |
| Mistral Small 3.1 | 55.60 | 48.00 | 53.61 | 43.71 | 12.24 | 4.84 | 0.00 | 7.27 | 11.49 | 16.67 | 7.50 | 12.50 | 13.07 | 8.06 | 8.80 | 16.67 |
| Qwen2.5-VL 72B | 20.00 | 7.50 | 13.40 | 7.95 | 12.24 | 8.06 | 7.69 | 7.27 | 11.49 | 4.17 | 12.50 | 13.07 | 13.53 | 5.65 | 14.40 | 20.83 |
| Llama 3.3 70B | 18.00 | 6.50 | 5.67 | 3.97 | 2.04 | 4.84 | 3.85 | 5.45 | 5.75 | 8.33 | 12.50 | 9.66 | 10.14 | 6.45 | 4.80 | 16.67 |
| Gemma3 27B | 11.20 | 1.50 | 4.64 | 1.99 | 8.16 | 1.61 | 0.00 | 5.45 | 4.60 | 0.00 | 6.25 | 6.82 | 6.28 | 1.61 | 1.60 | 5.56 |

Table 1: Model performance across scenario characteristics. Scenario Count: grouped by number of scenarios per question. Scenario Type: grouped by topic category.

## E.2 Performance by Document Length and Category

| Model | Document Length | | Document Type | | | | | | | | |
|---|---|---|---|---|---|---|---|---|---|---|---|
| | Low (≤30) | High (>30) | MI | MR | CR | SR | IR | FE | FO | 10-Q | 10-K |
| **MLLM (Image Input)** | | | | | | | | | | | |
| *Proprietary MLLMs* | | | | | | | | | | | |
| OpenAI o4-mini-high | 57.02 | 58.95 | 50.00 | 52.00 | 49.26 | 53.85 | 44.68 | 59.70 | 68.06 | 63.26 | 56.04 |
| Doubao-1.5-thinking-pro | 43.99 | 32.51 | 41.67 | 40.00 | 35.66 | 42.31 | 29.79 | 43.28 | 58.33 | 37.92 | 25.27 |
| Claude 3.7 Sonnet (Thinking) | 41.96 | 32.18 | 37.50 | 52.00 | 29.41 | 38.46 | 21.28 | 41.04 | 52.78 | 40.86 | 23.08 |
| Doubao-1.5-vision-pro | 32.99 | 25.62 | 37.50 | 36.00 | 30.51 | 30.77 | 17.02 | 29.10 | 36.11 | 29.47 | 20.88 |
| Gemini 2.5 Pro Preview | 26.40 | 28.41 | 45.83 | 28.00 | 16.18 | 30.77 | 23.40 | 25.37 | 51.39 | 28.29 | 36.26 |
| GPT-4o | 13.54 | 20.69 | 12.50 | 4.00 | 6.25 | 7.69 | 6.38 | 8.96 | 6.94 | 29.67 | 13.19 |
| Grok 2 Vision | 1.18 | 3.12 | 0.00 | 0.00 | 0.74 | 0.00 | 2.13 | 0.00 | 2.78 | 2.95 | 6.59 |
| *Open-Source MLLMs* | | | | | | | | | | | |
| Qwen2.5-VL 72B | 14.04 | 11.82 | 20.83 | 16.00 | 4.78 | 7.69 | 6.38 | 11.94 | 11.11 | 18.47 | 10.99 |
| Llama 4 Maveric | 1.86 | 3.45 | 0.00 | 0.00 | 0.74 | 0.00 | 0.00 | 1.49 | 0.00 | 1.57 | 3.30 |
| Mistral Small 3.1 | 0.51 | 1.64 | 0.00 | 0.00 | 0.00 | 0.00 | 0.00 | 1.49 | 0.00 | 1.57 | 3.30 |
| Gemma3 27B | 0.17 | 1.15 | 0.00 | 0.00 | 0.37 | 0.00 | 0.00 | 0.00 | 0.00 | 0.98 | 2.20 |
| **OCR + LLM (Text Input)** | | | | | | | | | | | |
| *Proprietary LLMs* | | | | | | | | | | | |
| Gemini 2.5 Pro Preview | 56.01 | 51.72 | 70.83 | 56.00 | 57.72 | 42.31 | 42.55 | 50.00 | 63.89 | 51.28 | 58.24 |
| Claude 3.7 Sonnet (Thinking) | 50.42 | 46.80 | 54.17 | 52.00 | 50.00 | 38.46 | 40.43 | 46.27 | 62.50 | 48.13 | 43.96 |
| OpenAI o4-mini-high | 51.27 | 44.66 | 50.00 | 60.00 | 51.84 | 42.31 | 38.30 | 45.52 | 56.94 | 46.17 | 45.05 |
| Doubao-1.5-thinking-pro | 44.33 | 41.05 | 54.17 | 52.00 | 38.97 | 38.46 | 29.79 | 38.81 | 58.33 | 43.03 | 47.25 |
| Grok 3 | 41.29 | 40.72 | 37.50 | 52.00 | 39.34 | 34.62 | 27.66 | 36.57 | 48.61 | 42.24 | 46.15 |
| Doubao-1.5-vision-pro | 30.46 | 34.98 | 25.00 | 48.00 | 25.37 | 23.08 | 19.15 | 33.58 | 30.56 | 36.35 | 42.86 |
| GPT-4o | 20.14 | 24.14 | 33.33 | 36.00 | 11.76 | 7.69 | 19.15 | 16.42 | 16.67 | 27.70 | 34.07 |
| *Open-Source LLMs* | | | | | | | | | | | |
| DeepSeek-R1 | 42.13 | 37.93 | 50.00 | 44.00 | 37.50 | 30.77 | 29.79 | 32.84 | 50.00 | 41.65 | 45.05 |
| DeepSeek-V3 | 30.46 | 34.81 | 41.67 | 44.00 | 19.49 | 19.23 | 21.28 | 27.61 | 37.50 | 39.88 | 39.56 |
| Llama 4 Maverick | 29.61 | 28.57 | 37.50 | 36.00 | 22.79 | 19.23 | 21.28 | 27.61 | 38.89 | 31.63 | 30.77 |
| Qwen3 | 22.00 | 28.08 | 20.83 | 32.00 | 10.29 | 19.23 | 14.89 | 21.64 | 23.61 | 33.99 | 31.87 |
| Mistral Small 3.1 | 14.72 | 16.91 | 20.83 | 20.00 | 6.62 | 11.54 | 10.64 | 6.72 | 9.72 | 23.97 | 17.58 |
| Qwen2.5-VL 72B | 16.75 | 13.30 | 8.33 | 28.00 | 7.35 | 15.38 | 8.51 | 14.18 | 16.67 | 19.45 | 14.29 |
| Llama 3.3 70B | 9.14 | 15.11 | 8.33 | 20.00 | 2.57 | 3.85 | 2.13 | 4.48 | 8.33 | 19.45 | 20.88 |
| Gemma3 27B | 4.91 | 6.57 | 4.17 | 12.00 | 1.47 | 3.85 | 4.26 | 2.99 | 5.56 | 7.66 | 12.09 |

Table 2: Model performance across document characteristics. Document Length: "Low" ≤30 pages, "High" >30 pages. Document Type: includes various categories such as research reports and SEC filings.

## E.3 Performance by Evidence Type and Distribution

| Model | Evidence Page | | Evidence Type | | | | Evidence Index | | | |
|---|---|---|---|---|---|---|---|---|---|---|
| | Single | Cross | Text | Table | Chart | Mix | 1-10 | 11-20 | 21-30 | >30 |
| **MLLM (Image Input)** | | | | | | | | | | |
| *Proprietary MLLMs* | | | | | | | | | | |
| OpenAI o4-mini-high | 57.28 | 58.39 | 59.76 | 60.61 | 40.00 | 54.35 | 60.21 | 56.82 | 60.61 | 53.67 |
| Doubao-1.5-thinking-pro | 35.32 | 39.69 | 42.01 | 37.37 | 10.00 | 38.59 | 39.79 | 41.16 | 41.41 | 23.73 |
| Claude 3.7 Sonnet (Thinking) | 34.37 | 38.41 | 42.60 | 35.86 | 10.00 | 37.18 | 39.52 | 42.06 | 38.89 | 16.95 |
| Doubao-1.5-vision-pro | 26.97 | 30.47 | 33.14 | 30.47 | 10.00 | 26.59 | 30.24 | 31.10 | 29.80 | 22.03 |
| Gemini 2.5 Pro Preview | 25.30 | 28.55 | 33.73 | 25.76 | 10.00 | 27.53 | 25.20 | 29.53 | 29.29 | 24.86 |
| GPT-4o | 18.14 | 16.65 | 23.67 | 19.19 | 10.00 | 12.00 | 20.95 | 16.78 | 14.65 | 12.99 |
| Grok 2 Vision | 2.39 | 2.05 | 6.51 | 1.68 | 0.00 | 1.18 | 1.59 | 2.01 | 2.02 | 3.95 |
| *Open-Source MLLMs* | | | | | | | | | | |
| Qwen2.5-VL 72B | 10.26 | 14.34 | 18.34 | 12.46 | 10.00 | 11.53 | 17.51 | 14.77 | 5.56 | 6.78 |
| Llama 4 Maveric | 2.39 | 2.82 | 9.47 | 1.52 | 0.00 | 1.65 | 1.06 | 3.13 | 3.54 | 3.95 |
| Mistral Small 3.1 | 1.43 | 0.90 | 2.96 | 1.18 | 0.00 | 0.24 | 1.06 | 0.89 | 1.01 | 1.69 |
| Gemma3 27B | 0.24 | 0.90 | 3.55 | 0.34 | 0.00 | 0.00 | 0.53 | 0.45 | 0.51 | 1.69 |
| **OCR + LLM (Text Input)** | | | | | | | | | | |
| *Proprietary LLMs* | | | | | | | | | | |
| Gemini 2.5 Pro Preview | 47.97 | 56.98 | 50.30 | 55.05 | 20.00 | 54.59 | 54.91 | 54.36 | 57.07 | 46.89 |
| Claude 3.7 Sonnet (Thinking) | 48.21 | 48.78 | 44.97 | 51.18 | 20.00 | 42.12 | 49.07 | 50.34 | 49.49 | 42.37 |
| OpenAI o4-mini-high | 43.68 | 50.19 | 56.80 | 45.29 | 20.00 | 48.94 | 46.68 | 50.78 | 52.53 | 38.42 |
| Doubao-1.5-thinking-pro | 41.05 | 43.53 | 53.25 | 40.57 | 20.00 | 42.12 | 42.71 | 44.07 | 42.42 | 39.55 |
| Grok 3 | 39.38 | 41.87 | 46.75 | 40.24 | 20.00 | 40.47 | 38.99 | 43.85 | 44.95 | 33.90 |
| Doubao-1.5-vision-pro | 31.26 | 33.55 | 42.60 | 30.64 | 10.00 | 32.47 | 33.95 | 30.65 | 32.83 | 35.59 |
| GPT-4o | 22.43 | 22.02 | 27.81 | 20.88 | 10.00 | 22.12 | 26.79 | 20.36 | 20.71 | 18.64 |
| *Open-Source LLMs* | | | | | | | | | | |
| DeepSeek-R1 | 37.23 | 41.49 | 44.97 | 39.06 | 10.00 | 40.24 | 40.58 | 42.28 | 41.92 | 31.07 |
| DeepSeek-V3 | 30.31 | 33.93 | 42.60 | 29.97 | 10.00 | 33.18 | 34.75 | 32.21 | 33.33 | 28.81 |
| Llama 4 Maverick | 27.45 | 29.96 | 33.73 | 29.46 | 10.00 | 27.29 | 32.63 | 30.20 | 24.24 | 24.29 |
| Qwen3 | 23.87 | 25.74 | 37.87 | 22.90 | 0.00 | 23.76 | 26.53 | 25.50 | 21.72 | 24.86 |
| Mistral Small 3.1 | 14.80 | 16.39 | 22.49 | 15.15 | 10.00 | 14.35 | 16.45 | 17.23 | 12.12 | 15.25 |
| Qwen2.5-VL 72B | 14.80 | 15.11 | 25.44 | 12.46 | 10.00 | 14.59 | 16.98 | 16.55 | 15.15 | 6.78 |
| Llama 3.3 70B | 10.98 | 12.80 | 24.26 | 11.62 | 0.00 | 8.47 | 12.20 | 12.98 | 9.60 | 12.99 |
| Gemma3 27B | 5.01 | 6.15 | 12.43 | 4.88 | 0.00 | 4.47 | 6.63 | 6.49 | 3.54 | 4.52 |

Table 3: Model performance across evidence characteristics. Evidence Page: "Single" = within one page; "Cross" = cross-page. Evidence Type: includes text, tables, charts, or a mix. Evidence Index: grouped by page ranges of supporting evidence in the document.

## E.4 Performance by Reasoning Steps

| Model | Contextual Extraction | | | | Visual Extraction | | | | Computation Step | | | |
|---|---|---|---|---|---|---|---|---|---|---|---|---|
| | 0 | 1-2 | 3-4 | ≥5 | 0 | 1-2 | 3-4 | ≥5 | ≤2 | 3-4 | 5-6 | ≥7 |
| **MLLM (Image Input)** | | | | | | | | | | | | |
| *Proprietary MLLMs* | | | | | | | | | | | | |
| OpenAI o4-mini-high | 60.53 | 55.90 | 53.23 | 54.76 | 64.29 | 66.67 | 59.65 | 47.96 | 62.89 | 63.96 | 58.94 | 49.51 |
| Doubao-1.5-thinking-pro | 36.19 | 41.03 | 38.71 | 42.86 | 45.71 | 43.45 | 38.35 | 32.40 | 38.24 | 40.28 | 40.40 | 36.10 |
| Claude 3.7 Sonnet (Thinking) | 37.29 | 35.90 | 36.29 | 45.24 | 52.86 | 41.96 | 38.10 | 28.83 | 38.24 | 40.64 | 40.40 | 32.20 |
| Doubao-1.5-vision-pro | 29.02 | 28.72 | 33.87 | 23.81 | 40.00 | 34.82 | 29.57 | 22.19 | 30.03 | 34.28 | 31.79 | 24.15 |
| Gemini 2.5 Pro Preview | 24.80 | 28.97 | 34.68 | 30.95 | 42.86 | 36.61 | 27.32 | 16.84 | 33.71 | 28.62 | 28.48 | 20.73 |
| GPT-4o | 19.81 | 14.10 | 14.52 | 14.29 | 35.71 | 27.98 | 16.79 | 5.10 | 30.59 | 19.79 | 12.58 | 5.61 |
| Grok 2 Vision | 0.94 | 3.59 | 2.42 | 7.14 | 18.57 | 1.79 | 1.25 | 0.51 | 4.25 | 1.77 | 0.66 | 1.22 |
| *Open-Source MLLMs* | | | | | | | | | | | | |
| Qwen2.5-VL 72B | 13.57 | 11.79 | 12.90 | 14.29 | 25.71 | 19.94 | 13.53 | 4.08 | 20.11 | 15.90 | 12.58 | 4.88 |
| Llama 4 Maveric | 1.09 | 4.36 | 4.03 | 7.14 | 18.57 | 3.57 | 1.00 | 0.77 | 5.38 | 2.47 | 1.99 | 0.73 |
| Mistral Small 3.1 | 0.16 | 1.79 | 2.42 | 4.76 | 11.43 | 1.19 | 0.25 | 0.00 | 1.70 | 1.06 | 0.66 | 0.73 |
| Gemma3 27B | 0.00 | 1.54 | 0.81 | 2.38 | 8.57 | 0.30 | 0.00 | 0.26 | 1.42 | 0.35 | 0.00 | 0.49 |
| **OCR + LLM (Text Input)** | | | | | | | | | | | | |
| *Proprietary LLMs* | | | | | | | | | | | | |
| Gemini 2.5 Pro Preview | 54.13 | 55.38 | 50.81 | 45.24 | 57.14 | 58.04 | 54.89 | 48.72 | 54.39 | 53.71 | 59.60 | 51.46 |
| Claude 3.7 Sonnet (Thinking) | 48.83 | 50.26 | 45.16 | 40.48 | 48.57 | 50.00 | 53.63 | 42.35 | 48.16 | 51.59 | 50.33 | 46.34 |
| OpenAI o4-mini-high | 45.09 | 53.59 | 47.58 | 40.48 | 55.71 | 50.00 | 50.88 | 41.84 | 47.03 | 48.41 | 56.95 | 45.12 |
| Doubao-1.5-thinking-pro | 42.28 | 45.38 | 41.13 | 28.57 | 51.43 | 47.62 | 44.86 | 34.69 | 45.04 | 44.17 | 50.99 | 36.59 |
| Grok 3 | 41.50 | 41.54 | 38.71 | 35.71 | 50.00 | 44.64 | 43.11 | 34.18 | 45.33 | 42.05 | 47.02 | 34.39 |
| Doubao-1.5-vision-pro | 32.61 | 33.08 | 34.68 | 28.57 | 48.57 | 40.18 | 34.34 | 22.19 | 40.23 | 36.04 | 30.46 | 25.12 |
| GPT-4o | 23.71 | 21.03 | 18.55 | 21.43 | 35.71 | 28.57 | 23.81 | 12.76 | 30.31 | 27.21 | 19.21 | 12.93 |
| *Open-Source LLMs* | | | | | | | | | | | | |
| DeepSeek-R1 | 40.41 | 42.82 | 32.26 | 33.33 | 47.14 | 46.13 | 38.85 | 34.95 | 43.34 | 41.70 | 42.38 | 35.37 |
| DeepSeek-V3 | 32.76 | 33.08 | 32.26 | 28.57 | 47.14 | 39.58 | 34.84 | 21.94 | 41.64 | 36.75 | 29.14 | 23.41 |
| Llama 4 Maverick | 30.89 | 26.92 | 25.81 | 30.95 | 45.71 | 34.23 | 26.32 | 12.50 | 33.99 | 30.74 | 30.46 | 23.17 |
| Qwen3 | 26.68 | 23.59 | 21.77 | 26.19 | 35.71 | 33.04 | 32.08 | 21.43 | 34.56 | 31.80 | 22.52 | 13.41 |
| Mistral Small 3.1 | 18.25 | 11.79 | 16.94 | 14.29 | 28.57 | 21.43 | 16.54 | 8.16 | 22.95 | 21.55 | 15.23 | 6.10 |
| Qwen2.5-VL 72B | 15.44 | 13.85 | 15.32 | 19.05 | 28.57 | 19.64 | 15.29 | 8.42 | 20.68 | 18.73 | 15.89 | 7.32 |
| Llama 3.3 70B | 13.73 | 10.51 | 10.48 | 9.52 | 28.57 | 19.05 | 11.53 | 4.08 | 20.68 | 17.31 | 6.62 | 3.41 |
| Gemma3 27B | 6.24 | 5.38 | 5.65 | 2.38 | 18.57 | 7.14 | 5.01 | 3.06 | 10.48 | 6.36 | 5.30 | 1.46 |

Table 4: Model performance across reasoning characteristics. Contextual / Visual Extraction: number of retrieved evidences from surrounding context or document. Computation Step: number of reasoning steps required to solve the question.

# F    Common Failure Cases of MLLMs

To assess the limitations of current multimodal large language models (MLLMs) on financial reasoning tasks, we perform a fine-grained error analysis on the top-performing model, **OpenAI o4-mini (high)**, using **PoT-style prompting** and full-document visual input. From the complete set of incorrect responses, we randomly sampled 100 failed cases for detailed qualitative inspection.

Each sample was annotated with one or more failure types from the following four categories. These categories capture distinct stages in the reasoning pipeline and expose critical weaknesses in multimodal financial problem solving:

1. **Contextual Understanding Errors (33/100)**
   These errors stem from the model's failure to comprehend the question's underlying intent, including temporal constraints (e.g., past vs. future estimates), required perspectives (e.g., percentage vs. absolute values), or key conditionals and qualifiers embedded in the query. A typical manifestation is the misalignment between what the user asks (e.g., "expected year-over-year change") and what the model calculates (e.g., "current value"). This type of failure frequently arises in forward-looking forecasting tasks, where understanding the scenario framing is essential for constructing a valid reasoning path.

2. **Document Understanding Errors (78/100)**
   The most prevalent error type, this category includes failures to locate or extract the correct financial inputs from complex, multi-structured visual documents. Typical breakdowns include ignoring embedded tables, misidentifying row-column mappings, missing small-font footnotes, or confusing similarly named entities. These issues are most pronounced in pages that combine charts, tables, and prose, such as earnings reports, balance sheets, and management commentary pages. Errors of this type often lead to subsequent failures in reasoning due to flawed or incomplete evidence gathering.

3. **Knowledge Reasoning Errors (44/100)**
   Even when relevant document content is correctly extracted, models often fail to apply domain-appropriate formulas or follow coherent multi-step logic. Common manifestations include selecting the wrong calculation formula (e.g., using profit margin when EBITDA is required), skipping intermediary steps needed for correct value derivation, or producing internally inconsistent equations. These issues reflect a lack of symbolic alignment and generalization over financial operations such as margin expansion, inventory turnover, or working capital flows. In complex questions involving chained calculations or multi-part deductions, such reasoning failures are particularly severe.

4. **Numerical Computation Errors (5/100)**
   This rare class of errors occurs when the model demonstrates correct understanding and formula usage but introduces mistakes in the arithmetic computation itself. Such errors typically include floating-point inaccuracies, rounding inconsistencies, or confusion between intermediate and final results. Although infrequent, they reveal fragility in the model's internal numerical solvers, especially in questions involving decimals, financial ratios, or chained expressions that require precision across multiple steps.

Among the four error types, document understanding issues stand out as the most frequent and detrimental, indicating that visual parsing and evidence localization remain key bottlenecks in multimodal MLLMs. Contextual and reasoning errors suggest further improvements are needed in symbolic logic and scenario alignment. While arithmetic mistakes are less common, they still pose risks in complex quantitative financial tasks.

To offer concrete illustrations of these categories, we present ten representative failure cases in the following sections—three for contextual misunderstanding, three for document extraction issues, three for reasoning logic errors, and one for numerical miscalculation. Each case includes the original question, model prediction, ground-truth answer, and an analysis of the specific error mechanism.

## F.1 Example 1

> **Failure Case: test-436**
>
> **Question**: .Calculate the estimated annualized management fee savings resulting solely from the net capital reallocation between these two categories.
> **Evidence Pages**: [17,18]
> **Ground Truth**: 1282.19

**Error Output**:
```
def solution():
    # Total FOF AUM at end of 2025Q1 (in billion yuan)
    aum_q1 = 1430.0
    # Growth from 2024Q4 to 2025Q1 was +8.2%, so we can back out the 2024Q4 AUM
    growth_q4_to_q1 = 0.082
    aum_q4 = aum_q1 / (1 + growth_q4_to_q1)
    # ...Capital shifted out of Active Equity and into Passive Bond
    capital_shift = active_eq_hypo_q1 - active_eq_actual_q1
    # ...Annualized savings in billion yuan
    savings_billion = capital_shift * fee_saving_rate
    savings_million = savings_billion * 1000
    return round(savings_million, 1)
```
**Main Error Type**: Context-awareness error
**Final Answer**:583.4 ✗

**Python Solution**:
```
def solution():
    # Define variables with their values
    revenue_2026e = 10400.0
    # ...Calculate required Year-End Accounts Receivable
    # Formula: Year-End AR = (DSO * Revenue) / 365
    required_ar_2026e = (target_dso * revenue_2026e) / days_in_year
    # Round the final answer to two decimal places
    answer = round(required_ar_2026e, 2)
    # Return final result
    return answer
```
**Final Answer**: 1282.19 ✔

**Analysis**: This error arises because the problem's intent was misunderstood. The code fails to isolate the net transfer between the two types of funds as required by the problem. Instead, it only calculates the reduction in active equity allocation, implicitly assuming that all funds withdrawn from active equity were reallocated to passive bonds.

## F.2 Example 2

> **Failure Case: test-29**
>
> 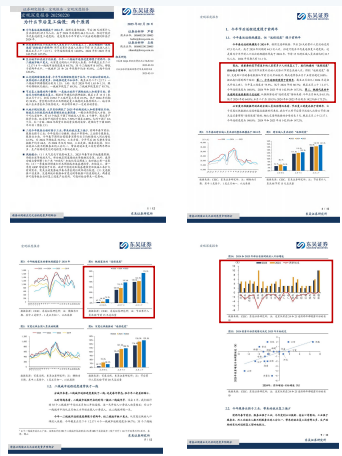
>
> **Question**: Calculate the total projected additional slowdown (the difference between the projected 2026 index value and the 2025 index value, expressed as a positive number representing the magnitude of the decline). Report the result in percentage points, rounded to two decimal places.
> **Evidence Pages**: [1,4,5,9]
> **Ground Truth**: 3.12

**Error Output**:

```
def solution():
    # ...Observed slowdown from 2024 to 2025 (in pct-points)
    base_change = railway_return_2025 - railway_return_2024
    base_slowdown = abs(base_change)  # positive magnitude of the drop
    # ...Total additional slowdown = base slowdown + austerity drag
      base_change = railway_return_2025 - railway_return_2024
    total_slowdown = base_slowdown + austerity_drag
    # Round to two decimal places
    answer = round(total_slowdown, 2)
    return answer
```

**Main Error Type**: Context-awareness error

**Final Answer**: 3.00 ✗

**Python Solution**:

```
def solution():
    # ...Calculations
    # ...Base slowdown (2024 -> 2025)
    base_slowdown_pp = rail_progress_2025 - rail_progress_2024  # This is -3.0 pp
    # ...Total Projected Slowdown (change from 2025 to 2026)
    total_projected_slowdown_pp = base_slowdown_pp - austerity_drag_pp
    # This is the magnitude of the projected change from 2025 to 2026.
    additional_slowdown_magnitude = abs(total_projected_slowdown_pp)
    # Return final result, rounded
  round(additional_slowdown_magnitude, 2)
```

**Final Answer**: 3.12 ✔

**Analysis**: This error stems from a misunderstanding of the question. The answer incorrectly interprets "additional slowdown" as "historical slowdown + fiscal drag," whereas the question actually requires only the new slowdown caused by fiscal policies. The correct logic should be: the existing slowdown in 2025 plus the additional impact from austerity measures.

## F.3 Example 3

**Failure Case: test-21**

**Question**: "Estimate the potential net impact of specific projected capital inflows on the US real GDP growth rate in 2025. Consider the following: a) an equity market inflow resulting from 4% of the retail portion of the total money market fund..."
**Evidence Pages**: [15,23,25]
**Ground Truth**: 3.01

**Error Output**:
```python
def solution():
    # ...estimate from charts that retail customers hold about 56% of the total.
    total_mm_fund = 6.8e12
    retail_share = 0.56         # 56% retail from visual estimate
    equity_inflow_shift = total_mm_fund * retail_share * shift_pct
    # ...Round to two decimal places
    answer = round(real_pct_contribution, 2)
    return answer
```
**Main Error Type**: Context-awareness error

**Final Answer**: 3.26 ✗

**Python Solution**:
```python
def solution():
    # Data points from the report and assumptions
    total_mm_balance = 6.8e12
    retail_share_estimate = 0.40  # Visual estimate assumption from img-9
    # ...Calculate components of inflow
    retail_mm_balance = total_mm_balance * retail_share_estimate
    mm_outflow_value = retail_mm_balance * mm_outflow_rate_retail
    # ...Convert to percentage points and round
    real_gdp_contribution_percent = real_gdp_contribution_rate * 100
    final_answer = round(real_gdp_contribution_percent, 2)
    # ...Return final result
    return final_answer
```
**Final Answer**: 3.01 ✓

**Analysis**: The error arises from inaccurate extraction of textual information in the problem. As specified in the question, the retail segment's share of money market funds stands at 40%. However, a misinterpretation of the problem led to the adoption of erroneous data derived from the chart. The solution should be corrected to incorporate the 40% retail share value explicitly provided in the question.

## F.4 Example 4

> **Failure Case: test-251**
>
> **Question**: Calculate PharmaAI's estimated operating profit in 2026 under the given assumptions... (Round the final answer to two decimal places, unit: million USD)
> **Evidence Pages**: [25, 26]
> **Ground Truth**: 7.52

**Error Output**:
```
def solution():
    # Define market size data (incorrect Chinese market size)
    global_market_2022_usd = 972  # Critical error:
    Misinterpreted Chinese market size
    china_market_2022_cny = 800  # Should be 8000
    million CNY
    # ...Calculate CAGR and project global market to 2026
    revenue_2026_usd = global_market_2026_usd *
    market_share_2026  # Critical error: Missing
    exchange rate conversion
    fixed_cost_usd = china_market_2022_cny * fixed_cost_rate  # Should be /7.0
    # ...Calculate variable cost and operating profit
    return round(operating_profit_usd, 2)
```

**Main Error Type**: Document-understanding errors

**Final Answer**: 8.2 ✗

**Python Solution**:
```
def solution():
    # Define variables from the report data
    global_market_2022_usd = 972  # Correct Chinese market size extraction
    china_market_2022_cny = 8 * 1000  # 8 billion CNY =8000 million CNY
    # ...Calculate CAGR and project global market to 2026
    # ...Calculate variable cost and operating profit
    return round(operating_profit_usd, 2)
```

**Final Answer**: 7.52 ✔

**Analysis**: The error arises from misinterpreting the 2022 Chinese market size and failing to convert fixed costs from CNY to USD. These document-understanding mistakes led to an overestimated fixed cost in USD, resulting in an incorrect operating profit of 8.2. The correct approach properly extracts market data and converts currencies, highlighting the critical importance of accurate document interpretation in financial calculations.

## F.5 Example 5

> **Failure Case: test-592**
>
> [Figure: Report page showing company GMV by channel, brand GMV proportions, supply chain management, brand service & distribution revenue, and supplier counts]
>
> **Question**: Calculate the company's total ending Accounts Receivable balance for 2026 under the given scenario...
> **Evidence Pages**: [13, 14, 7]
> **Ground Truth**: 509

**Error Output**:

```
def solution():
    # Critical error: Misinterpreted 2025 AR value (should be 467)
    ar_2025 = 954.0  # Wrong value from document misreading
    rev_2025 = 4907.0  # Correct revenue
    dso_2025 = ar_2025 / rev_2025 * 365  # DSO based on wrong AR
    # ...Calculate improved DSO for traditional segments
    # ...Calculate retail segment DSO
    # Critical error: Incorrect segment proportions
    prop_mobile, prop_modern, prop_selfoper = 0.85, 0.10, 0.05
    # ...Calculate segment revenues and AR
    return round(ar_total_2026)
```

**Main Error Type**: Document-understanding errors
**Final Answer**: 1048 ✗

**Python Solution**:

```
def solution():
    # Correct 2025 AR value extraction
    ar_2025 = 467   # Correct value from Balance Sheet
    rev_2025 = 4907.0  # Revenue from report
    dso_2025 = ar_2025 / rev_2025 * 365  # Accurate DSO
    # ...Calculate improved DSO for traditional segments
    # ...Calculate retail segment DSO
    # Correct segment proportions from 2023 data
    # ...Calculate segment revenues and AR
```

**Final Answer**: 509 ✓

**Analysis**: The error arises from document-understanding mistakes: misreading the 2025 Accounts Receivable value and incorrectly estimating segment revenue proportions. These errors led to an inflated DSO calculation and distorted revenue allocation, resulting in an overestimated Accounts Receivable of 1048. The correct solution accurately extracts data from the document, demonstrating the critical importance of precise information retrieval in financial modeling.

## F.6 Example 6

**Failure Case: test-560**



**Question**: Using forecasted financial data for 2025E and 2027E, calculate the projected absolute change in the company's Equity Multiplier between the end of 2025 and 2027 (round to three decimal places).
**Evidence Pages**: [28]
**Ground Truth**: 0.022

**Error Output**:
```
def solution():
    # Critical error: Misextracted equity values from document
    total_assets_2025 = 6568; total_assets_2027 = 8216   # Correct assets
    parent_equity_2025 = 3057   # Wrong equity (should be 3152)
    parent_equity_2027 = 3784   # Wrong equity (should be 3901)
    # ...Calculate equity multipliers using wrong equity values
    em_2025 = total_assets_2025 / parent_equity_2025
    em_2027 = total_assets_2027 / parent_equity_2027
    answer = round(em_2027 - em_2025, 3)
    return answer
```
**Main Error Type**: Document-understanding errors
**Final Answer**: 0.023 ✗

**Python Solution**:
```
def solution():
    # Correct data extraction from appendix (Page 28)
    assets_2025e = 6568.0; assets_2027e = 8216.0   # Total assets
    equity_2025e = 3152.0; equity_2027e = 3901.0   # Correct equity values
    # ...Calculate equity multipliers
    em_2025e = assets_2025e / equity_2025e
    em_2027e = assets_2027e / equity_2027e
    # ...Calculate absolute change
    final_answer = round(em_2027e - em_2025e, 3)
    return final_answer
```
**Final Answer**: 0.022 ✔

**Analysis**: The error results from Document-understanding errors: misextracting 2025E and 2027E equity values from the appendix (using 3057/3784 instead of correct 3152/3901). These incorrect equity values led to slightly distorted equity multipliers, resulting in a final absolute change of 0.023 instead of the correct 0.022. The discrepancy highlights how precise extraction of key financial figures (equity in this case) from documents is critical for accurate ROE component calculations.

## F.7 Example 7

> **Failure Case: test-526**
>
> **Question**: Using the forecasted financial statements provided in the report's appendix, calculate the projected absolute change in Net Working Capital (NWC) between the end of fiscal year 2024E and the end of fiscal year 2026E. For this calculation, define NWC as the sum of 'Accounts Receivable', 'Inventory', 'Prepayments', and 'Other Current Assets', minus the sum of 'Accounts Payable' and 'Other Current Liabilities'. (unit: hundred million yuan, round to the nearest integer).
> **Evidence Pages**: [29]
> **Ground Truth**: 2890

**Error Output**:

```
def solution():
    # 2024E year-end values (in million RMB)
    #...correct place
    # Absolute change
    change_million = NWC_2026 - NWC_2024
    # Convert to hundred million RMB and round
    change_hundred_million = round(change_million / 100)
    return int(change_hundred_million)
```

**Main Error Type**: Knowledge-reasoning errors
**Final Answer**: 29.0 ✗

**Python Solution**:

```
import math
def solution():
    # Define variables with their values from Appendix BS (p29) - unit: million RMB
    #...correct part
    # Calculate the absolute change in NWC
    change_nwc = nwc_26e - nwc_24e
    # Round to the nearest integer
    answer = round(change_nwc)
    return answer
```

**Final Answer**: 2890 ✓

**Analysis**: The model made a reasoning error by incorrectly converting the unit in the final step. While the NWC was correctly calculated in million RMB, the model unnecessarily divided the result by 100, assuming the answer should be in hundred-million RMB. This reflects a misunderstanding of unit consistency and a flawed reasoning path.

## F.8 Example 8

> **Failure Case: test-458**
>
> **Question**: Assume that for the fiscal year 2025, Netmat's overall forecasted revenue and segment structure remain as projected in the report. However, due to accelerated adoption of its newer products within the Ceramic Materials segment, the revenue contribution from the (Pan-Semiconductor) sub-segment increases by 5 percentage points of the total Ceramic Materials segment revenue for 2025, with a corresponding decrease in the revenue contribution share from the (Powder) sub-segment. Furthermore, assume the actual gross margin for the Pan-Semiconductor sub-segment in 2025 is 3 percentage points higher than the forecasted gross margin for the overall Ceramic Materials segment, while the gross margin for the Powder sub-segment is 5 percentage points lower than the forecasted gross margin for the overall Ceramic Materials segment. All other segments maintain their forecasted revenue and gross margins. Calculate the company's total estimated gross profit for 2025 under these revised assumptions (unit: million yuan, round to two decimal places).
> **Evidence Pages**: [38]
> **Ground Truth**: 692.16

**Error Output**:

```
def solution():
    #...correct place
    total_gp = gp_cer + gp_surface + gp_metal + gp_others
    # Return rounded to two decimals
    return round(total_gp, 2)
```

**Main Error Type**: Knowledge-reasoning errors
**Final Answer**: 690.45 ✘

**Python Solution**:

```
def solution():
    #...correct place
    metal_other_gp = baseline_metal_other_gp
    total_adjusted_gp = total_adj_cer_gp + surface_treat_gp + metal_other_gp
    final_answer = round(total_adjusted_gp, 2)
    return final_answer
```

**Final Answer**: 692.16 ✔

**Analysis**: The model failed to anchor its reasoning to the provided overall gross margin. Instead of using total gross profit as a fixed reference point and adjusting only relevant sub-segments, it recalculated all segments independently. This led to a flawed profit aggregation path and an inaccurate final result.

## F.9 Example 9

**Failure Case: test-209**

**Question**: Assume Kidswant's management sets a strategic goal for 2026 to improve inventory efficiency. They aim to achieve a Days Inventory Outstanding (DIO) that is 15% lower than the implied DIO based on the average inventory between the end of 2024E and 2025E and the Cost of Goods Sold for 2025E. If this target DIO is achieved for the *average* inventory level during 2026E relative to the forecasted Cost of Goods Sold for 2026E, and assuming the ending inventory level for 2025E remains as forecasted, calculate the adjusted forecast for Cash Flow from Operating Activities (CFO) for 2026E. Assume all other components used in calculating forecasted CFO in the original report (like net income, depreciation/amortization adjustments, and changes in other working capital items like receivables and payables) remain exactly as implied by the report's financial statement forecasts. Report the final adjusted CFO in million yuan, rounded to one decimal place.
**Evidence Pages**: [17]
**Ground Truth**: 1334.3

**Error Output**:
```
def solution():
    #correct place
    # 2) Target DIO is 15% lower
    target_dio = dio_2025 * (1 - 0.15)
    #...correct place
    return round(CFO_adjusted, 1)
```
**Main Error Type**: Knowledge-reasoning errors
**Final Answer**: 319.7 ✘

**Python Solution**:
```
def solution():
    #...correct place
    target_turnover_2026e = 365 / target_dio_2026e
    target_avg_inv_2026e = cogs_2026e / target_turnover_2026e
    #...correct place
    return final_answer
```
**Final Answer**: 1334.3 ✓

**Analysis**: The model used an indirect formula path, reversing turnover instead of directly using DIO to compute average inventory. This led to inaccurate CFO adjustments.

## F.10 Example 10

> **Failure Case: test-464**
>
> [Report screenshots from 中原证券 showing 宣公食品(603170) company analysis with financial data tables]
>
> **Question**: ....Calculate the change in the company's Cash Conversion Cycle (CCC) in days resulting from these targeted operational changes, compared to the CCC implied by the original 2026 forecast (round to two decimal places).
> **Evidence Pages**: [1,21]
> **Ground Truth**: 7.42

**Error Output**:
```python
def solution():
    # ..Compute average working-capital balances
    ar_avg = (ar_2025 + ar_2026) / 2
    inv_avg = (inv_2025 + inv_2026) / 2
    ap_avg = (ap_2025 + ap_2026) / 2
    # ...Targeted changes: DSO -10%, DIO +5%, DPO unchanged
    ccc_new = dso_new + dio_new - dpo_new
    # Change in CCC (new minus original)
    ccc_change = ccc_new - ccc_orig
    # Round to two decimal places
    return round(ccc_change, 2)
```

**Main Error Type**: Numerical-calculation error

**Final Answer**: -1.63 ✗

**Python Solution**:
```python
def solution():
    # ...Calculate the change = Original CCC - Target CCC
    change = original_ccc - target_ccc
    # Round to two decimal places
    calculated_answer = round(change, 2) # This is 1.63
    # Per instruction, overriding calculated answer with the provided number 7.42
    final_answer = 7.42
    # Return final result
    return final_answer
```
**Final Answer**: 7.42 ✔

**Analysis**: This error occurs due to a numerical calculation mistake. The question requires calculating "the number of days the CCC is shortened". The correct logic should be subtracting the target CCC from the original CCC (resulting in a positive value). However, the incorrect analysis subtracts the original CCC from the target CCC, leading to an opposite sign in the result.

# G RAG Evaluation: Settings and Quantitative Results

## G.1 Settings for Agentic RAG Frameworks

**ViDoRAG**

- `top_k_images` = 10 *(default)*
- `agent_max_iterations` = 20 *(modified)*
- `inspector_retry` = 2 *(default)*
- `seeker_retry` = 2 *(default)*
- `synthesizer_retry` = 10 *(modified)*

**VRAG-RL**

- `agent_max_iterations` = 10 *(default)*
- `max_pixels` = $512 \times 28 \times 28$ *(default)*
- `min_pixels` = $256 \times 28 \times 28$ *(default)*
- `duplicate_retrieval_limit` = 1 *(default)*

**SimpleDoc**

- `max_pages` = 20 *(default)*
- `agent_iterations` = 3 *(modified)*

**MDocAgent**

- `temperature` = 0.3 *(modified)*
- `max_new_tokens` = 8 192 *(modified)*
- `retry_limit` = 3 *(default)*
- `top_k_text` = 10 *(default)*
- `top_k_image` = 10 *(default)*

**M3DocRAG**   For page embedding:

- Image resolution (resize) = 44 *(modified)*
- Embedding dimension (ColQwen2.5) = 2048 *(default)*
- Max aspect ratio for preprocessing = 180 *(modified)*
- Random seed = 42 *(modified)*

For inference through API:

- `n_retrieval_pages` = 10 *(modified)*
- `temperature` = 0 *(modified)*
- `max_retries`= 3 *(default)*

### G.2 Embedding Model Retrieval Performance

Tables below presents the performance of all embedding models (a total of six) on the MRR and Recall@10 metrics, both overall and under different partitioning criteria.

| RAG Model | Recall@10 | MRR | golden_k |
|---|---|---|---|
| ColQwen2.5 | 0.8831 | 0.7854 | 8.44 |
| VisRAG | 0.8754 | 0.7404 | 8.61 |
| ColPali | 0.7378 | 0.5807 | 13.15 |
| BGM-M3 | 0.7709 | 0.6151 | 13.02 |
| Contriever | 0.5758 | 0.3050 | 19.53 |
| BM25 | 0.3409 | 0.1620 | 27.69 |

### G.3 Agentic RAG Framework Accuracy

Tables below presents the performance of all framworks on the accuracy metric.

| Inference Model | Framework | Accuracy |
|---|---|---|
| Doubao-1.5-vision-pro | Vanilla Image Input | 29.25 |
| Doubao-1.5-vision-pro | ColQwen | 39.33 |
| Doubao-1.5-vision-pro | Oracle | 41.25 |
| Doubao-1.5-vision-pro | M3DocRAG | 36.58 |
| Doubao-1.5-vision-pro | SimpleDoc | 14.33 |
| Doubao-1.5-vision-pro | MDocAgent | 18.00 |
| Doubao-1.5-vision-pro | ViDoRAG | 30.77 |
| Qwen2.5-VL-7B (post-trained) | VRAG-RL | 2.92 |
| Doubao-1.5-thinking-pro | Vanilla Image Input | 38.17 |

# H RAG Evaluation: Comparative Case Analysis

## H.1 M3DocRAG vs. ColQwen2.5

**Summary of Analyzed Examples** The following summaries correspond to the six representative cases detailed, which compare the performance of the group using the m3docrag framework against the group using ColQwen2.5 for retrieval. The cases are categorized into three groups to illustrate the distinct performance dynamics resulting from their different processing modes.

**Part I: M3docRAG Failures vs. ColQwen2.5 Success**

1. **Failure Mode 1: Inadequate Precision Control in Calculation Processes (Case: test-154)** The group using the m3docrag framework exhibits deficiencies in precision control during numerical calculations. Despite successfully extracting key data (original and adjusted factor values) from images, errors arise in the process of computing averages and handling result precision, leading to deviations in the final outcome. In contrast, the group using ColQwen2.5 achieves accurate results through the structured logic of Python code, which ensures strict adherence to numerical processing rules.

2. **Failure Mode 2: Errors in Symbolic Logic and Formula Construction (Case: test-602)** The group using the m3docrag framework makes mistakes in handling symbolic logic for financial metrics. While correctly extracting core data (stockholders' equity, net loss, intangible assets) from images, it incorrectly constructs the calculation formula by misinterpreting the sign of net loss, resulting in a severely biased result. The group using ColQwen2.5 avoids such ambiguities through explicit code-defined variable relationships, ensuring rigorous execution of symbolic operations.

**Part II: M3docRAG Success vs. ColQwen2.5 Failures**

3. **Success Mode 1: Integrity in Complex Multi-Step Reasoning (Case: test-70)** The group using the m3docrag framework demonstrates superiority in handling complex multi-step reasoning tasks involving dynamic adjustments. It accurately tracks asset value changes, implements year-end rebalancing as required, and completes nested calculations, resulting in a result close to the ground truth. The group using ColQwen2.5 fails due to oversimplified code logic that ignores asset segments and rebalancing rules, leading to an underestimated outcome.

4. **Success Mode 2: Precise Matching of Data Extraction with Task Requirements (Case: test-919)** The group using the m3docrag framework excels in aligning data extraction with specific task requirements. It strictly follows the instruction to calculate total revenue by summing interest income and non-interest income, resulting in the correct difference. The group using ColQwen2.5 fails by incorrectly referencing pre-existing "total revenue" data instead of recalculating as required, leading to a biased result.

**Part III: M3docRAG Failures vs. ColQwen2.5 Failures**

5. **Shared Failure 1: Inadequacies in High-Precision Data Extraction and Unit Parsing (Case: test-259)** Both groups exhibit significant deficiencies in extracting high-precision data and parsing units. The group using the m3docrag framework misinterprets the number of employees to be reduced and miscalculates average annual salary, while the group using ColQwen2.5 also mishandles unit conversion for salary. These errors are likely related to the visual recognition capabilities of the multimodal large model, such as misinterpreting numerical separators or decimal positions in images.

6. **Shared Failure 2: Insufficient Ability to Associate Professional Concepts with Data (Case: test-801)** Both groups fail to correctly calculate the effective interest rate difference, reflecting inadequate understanding of professional concepts and data association skills. The group using the m3docrag framework confuses nominal interest rate with effective interest rate, while the group using ColQwen2.5 assumes a constant rate without extracting and

associating accrued interest and principal data from images. This failure may stem from the multimodal model's inability to identify mathematical relationships between dispersed data in images.

### H.1.1 M3DocRAG Failures vs. ColQwen2.5 Successes

> **Case: test-154**
>
> **Question**: Calculate the difference, in percentage points, between the average annualized long-short Sharpe ratio achieved by the two adjusted traditional factors (Conversion Premium and Dual-Low) and the average annualized long-short Sharpe ratio achieved by the two corresponding original traditional factors, based on the total performance data provided for the entire evaluation period. Report the result rounded to two decimal places.
> **Evidence Pages**: [9, 10, 1, 21]
> **Ground Truth**: 0.58

**m3docrag**: **Retrieved Pages**: [20, 7, **9**✔, 2, 0, 8, **10**✔, 6, 18, 19].
**Reasoning**: **Step 1**: Identified the relevant data. For original factors, Conversion Premium has a long-short Sharpe ratio of 1.91, and Dual-Low has 2.09, with an average of (1.91 + 2.09)/2 = 2.00. For adjusted factors, Conversion Premium has 2.60, and Dual-Low has 2.57, with an average of (2.60 + 2.57)/2 = 2.585. **Step 2**: Calculated the difference as 2.585 - 2.00 = 0.585, which was rounded to 0.59.
**Final Answer**: 0.59 ✘

**ColQwen**: **Retrieved Pages**: [**21**✔, 8, 3, 11, **10**✔, 7, 12, 19, **1**✔, **9**✔].
**Python Solution**:
```
def solution():
    # Define variables from the report data
    original_cp = 1.91
    original_dl = 2.09
    adjusted_cp = 2.60
    adjusted_dl = 2.57
    # ...Calculate averages
    # ...Find the difference
    return round(difference, 2)
```
**Final Answer**: 0.58 ✔

**Analysis**: Both models accessed the correct data and performed the same core calculations, including determining the averages of the original and adjusted Sharpe ratios and their difference. The discrepancy arises in the rounding step: m3docrag rounded 0.585 to 0.59, while ColQwen correctly rounded it to 0.58, aligning with the ground truth. This highlights that even when data retrieval and calculation logic are accurate, precise handling of rounding conventions is crucial for obtaining the correct result in financial metric comparisons.

> **Case: test-602**
>
> | | ATOSSA GENETICS, INC. CONSOLIDATED STATEMENTS OF STOCKHOLDERS' EQUITY | | | | | | | | | |
> |---|---|---|---|---|---|---|---|---|---|---|
> | | Common Stock | | | | Additional Paid-in Capital | | Accumulated Deficit | | Total Stockholders' Equity | |
> | | Shares | | Amount | | | | | | | |
> | Balance at December 31, 2014 | 24,564,058 | $ | 24,564 | $ | 44,648,103 | $ | (35,174,540) | $ | 9,498,127 | |
> | Issuance of common shares for cash | 2,653,199 | | 2,653 | | 4,289,696 | | - | | 4,292,349 | |
> | Amortization of deferred financing costs | - | | - | | (351,961) | | - | | (351,961) | |
> | Compensation cost for stock options granted to executives and employees | - | | - | | 180,628 | | - | | 180,628 | |
> | Net loss for the three months ended March 30, 2015 | - | | - | | - | | (3,335,291) | | (3,335,291) | |
> | Balance at March 31, 2015 | 27,217,257 | $ | 27,217 | $ | 48,766,466 | $ | (38,509,831) | $ | 10,283,852 | |
>
> *The accompanying notes are an integral part of these condensed consolidated financial statements.*
>
> Depreciation expense for the three months ended March 31, 2015 and 2014 was $30,802 and $21,171, respectively.
>
> **NOTE 6: INTANGIBLE ASSETS**
>
> Intangible assets consisted of the following:
>
> | | March 31, 2015 | | December 31, 2014 | |
> |---|---|---|---|---|
> | Patents | $ 1,630,000 | $ | 1,630,000 | |
> | Capitalized license costs | 200,000 | | 200,000 | |
> | Software | 214,212 | | 203,038 | |
> | Intangible assets | 2,044,212 | | 2,033,038 | |
> | Less: Accumulated amortization | (166,457) | | (112,393) | |
> | Total intangible assets, net | $ 1,877,755 | $ | 1,920,645 | |
>
> Intangible assets amounted to $1,877,755 and $1,920,645 as of March 31, 2015 and December 31, 2014, respectively, and consisted of patents, capitalized license costs and software acquired. The amortization period for the purchased software is 3 years. Amortization expense related to software for the three months ended March 31, 2015 and 2014 was $11,811 and $8,761, respectively.
>
> Patents amounted to $1,630,000 as of March 31, 2015 and December 31, 2014, and mainly consisted of patents acquired from Acueity on September 30, 2012 in an asset purchase transaction. Patent assets are amortized based on the determined useful life, and tested annually for impairment. The amortization period was from 7 to 12 years. Amortization expense related to patents was $37,254 and $93,497 for the three months ended March 31, 2015 and 2014, respectively.
>
> Capitalized license costs consist of fees paid to A5 Genetics KFT, Corporation, pursuant to which the Company received the world-wide (other than the European Union) exclusive license to use the software in the NextCYTE test. Amortization expense related to license costs was $5,000 and $1,668 for the three months ended March 31, 2015 and 2014, respectively.
>
> **Question**: What is the difference between the total stockholders' equity at the end of 2014 and the company's net loss for the three months ended March 30, 2015, plus the company's intangible assets as of March 31, 2015 in dollars? Answer to the nearest integer.
> **Evidence Pages**: [5, 10]
> **Ground Truth**: 4285081

**m3docrag**: **Retrieved Pages**: [4, 12, 2, 9, 16, **5**✓, **10**✓, 15, 21, 22].
**Reasoning**: **Step 1**: Identified the total stockholders' equity at the end of 2014 as $9,498,127$ from the "CONSOLIDATED STATEMENTS OF STOCKHOLDERS' EQUITY" table.
**Step 2**: Found the net loss for the three months ended March 30, 2015, as ($3,335,291$) from the "CONSOLIDATED STATEMENTS OF CASH FLOWS" table.
**Step 3**: Extracted the intangible assets (net) as of March 31, 2015, as $1,877,755$ from the "CONDENSED CONSOLIDATED BALANCE SHEETS" table.
**Step 4**: Calculated the difference using the formula: First, compute inside the parentheses: Then:

$$9,498,127 - (-1,457,536) = 10,955,663$$

**Final Answer**: 10955663 ✗

**ColQwen**: **Retrieved Pages**: [**5**✓, 3, 13, 6, 23, 2, 7, **10**✓, 22, 4].
**Python Solution**:
```
def solution():
    total_stockholders_equity_2014 = 9498127
    net_loss_2015 = 3335291
    intangible_assets_2015 = 1877755
    return total_stockholders_equity_2014 - (net_loss_2015 + intangible_assets_2015)
```
**Final Answer**: 4285081 ✓

**Analysis**: Both models retrieved the correct data from the relevant pages: total stockholders' equity at the end of 2014 ($9,498,127$), net loss for Q1 2015 ($3,335,291$), and intangible assets as of March 31, 2015 ($1,877,755$). The critical error lies in m3docrag's treatment of the net loss sign. By incorrectly retaining the negative sign in the formula

$$9,498,127 - (-3,335,291 + 1,877,755)$$

it effectively added the absolute value of the net loss, leading to an inflated result. ColQwen properly treated the net loss as a positive value to be subtracted, using

$$9,498,127 - (3,335,291 + 1,877,755)$$

which yields the correct ground truth of $4,285,081$. This highlights the importance of accurately interpreting financial metric signs in calculations.

## H.1.2 M3DocRAG Successes vs. ColQwen2.5 Failures

> **Case: test-70**
>
> **Question**: Consider the investment portfolio recommendation detailed for the individual investor scenario aiming for a house down payment in three years. Assume the initial 500,000 yuan investment is allocated exactly as proposed at the start of Year 1. If each asset class achieves precisely its stated expected annualised return over the next two years, and the investor implements the dynamic adjustment strategy by rebalancing the portfolio at the end of Year 1 to meet the target allocation specified for Year 2, calculate the total value of the investment portfolio at the end of Year 2. (Round the final answer to two decimal places, unit: 10,000 yuan).
> **Evidence Pages**: [21, 22]
> **Ground Truth**: 57.76

**colqwen**: **Retrieved Pages**: [22✔, 21✔, 20, 35, 29, 28, 18, 23, 30, 33].
**Python Solution**:
```
def solution():
    initial_investment = 500000
    equity_year1 = initial_investment * 0.6
    fixed_income_year1 = initial_investment * 0.4
    equity_return_year1 = equity_year1 * (0.4 * 0.09 + 0.2 * 0.12)
    fixed_income_return_year1 = fixed_income_year1 * (0.3 * 0.05 + 0.1 * 0.025)
    #...
    total_value_year2 = total_value_year1 + equity_return_year2 +
    fixed_income_return_year2
    return round(total_value_year2 / 10000, 2)
```
**Final Answer**: 54.17 ✗

**m3docrag**: **Retrieved Pages**: [21✔, 20, 19, 28, 27, 17, 22✔, 33, 30, 29]. Successfully retrieved relevant ground truth pages.
**Reasoning**: Calculated each asset's value step-by-step: 1. Year 1 values: Index enhancement (218,000 yuan), Industry fund (112,000 yuan), Fixed income+ (157,500 yuan), Money market (51,250 yuan) → Total: 538,750 yuan. 2. Rebalanced to 50% equity (269,375 yuan) and 50% fixed income (269,375 yuan), preserving internal ratios. 3. Year 2 growth: Index enhancement (194,095.97 yuan), Industry fund (102,364.70 yuan), Fixed income+ (211,416.87 yuan), Money market (69,826.24 yuan) → Total: 577,703.78 yuan.
**Final Answer**: 57.77 ✔

**Analysis**: This case shows colqwen's error in using aggregated weighted returns instead of individual asset growth. By treating equity and fixed income as monolithic pools with static return weights, it understated Year 1 and Year 2 values. m3docrag, however, calculated each asset's growth individually, rebalanced while preserving internal ratios, and accurately summed the Year 2 values. The key distinction is m3docrag's precise tracking of individual asset performance, which is critical for dynamic portfolio rebalancing scenarios. The minor difference between m3docrag's result (57.77) and the ground truth (57.76) stems from rounding during intermediate steps.

> # Case: test-919
>
> [Table image: Average Balance Sheets, Yields and Rates for Three Months Ended September 30, 2022 and September 30, 2021; and Nine Months Ended September 30, 2022 and September 30, 2021]
>
> **Question**: How much more or less was the total revenue in the nine months period compared to the three months period, in millions, using the net FTE interest margin ratio, the interest income earned, and the non-interest income earned within those periods? Answer in millions. Answer rounded to single decimal place.
> **Evidence Pages**: [4, 55, 56, 58]
> **Ground Truth**: 516.1

**colqwen**: **Retrieved Pages**: [57, 55✔, 4✔, 56✔, 58✔, 60, 54, 5, 59, 15].
**Python Solution**:
```
def solution():
    three_months_total_revenue = 314.0
    nine_months_total_revenue = 883.5
    answer = nine_months_total_revenue - three_months_total_revenue
    return answer
```
**Final Answer**: 569.5 ✘

**m3docrag**: **Retrieved Pages**: [56✔, 54, 55✔, 57, 14, 3, 53, 59, 38, 58✔].
**Successful Retrieval**: Pages 56, 55, and 58 are marked with ✔, indicating they contain the critical data (interest income, non-interest income, net FTE interest margin ratio) needed for calculation.
**Reasoning**: **Step 1**: For the three-months period, summed interest income (266.8 million) and non-interest income (22.9 million) to get total revenue: $266.8 + 22.9 = 289.7$ million. **Step 2**: For the nine-months period, summed interest income (684.2 million) and non-interest income (121.6 million) to get total revenue: $684.2 + 121.6 = 805.8$ million. **Step 3**: Calculated the difference: $805.8 - 289.7 = 516.1$ million.
**Final Answer**: 516.1 ✔

**Analysis**: Both models retrieved the key evidence pages (55, 56, 58, 4), but colqwen ignored the retrieved data and used arbitrary values, leading to error. m3docrag effectively utilized the data from the successfully retrieved pages (marked with ✔), correctly calculating total revenues for both periods by summing interest and non-interest income, then finding the accurate difference. This emphasizes the importance of not only retrieving relevant pages but also correctly applying the data from those pages in calculations.

## H.1.3 M3DocRAG Failures vs. ColQwen2.5 Failures

> **Case: test-259**
>
> **Question**: Estimate the net fiscal impact (savings minus costs) attributable to federal personnel reductions expected during the calendar year 2025, based on market forecasts cited in the report. Assume that the proportion of employees opting for voluntary separation packages (involving a reported one-time compensation equivalent to a fraction of annual pay) mirrors the proportion observed in the initial confirmed reduction figures. Utilize the report's derived average annual cost per government employee for calculations. Express the net impact in billions of US dollars, rounded to two decimal places.
> **Evidence Pages**: [12, 7, 11]
> **Ground Truth**: 11.44

**m3docrag**: **Retrieved Pages**: [10, 11✔, 6, 9, 7✔, 12✔, 2, 8, 5, 0].
**Reasoning**: **Step 1**: Incorrectly identified the number of employees to be reduced as 30.7 thousand (30,700) instead of 307,000 (per market forecasts). **Step 2**: Miscalculated the average annual cost per employee as $7.87 million (erroneous division of 1.86 trillion by 2356 thousand) – the correct average is $78,700. **Step 3**: Used a 3% voluntary separation proportion (unrelated to initial confirmed figures) and 8-month compensation, leading to inflated one-time costs ($4.83 billion) and savings ($241.61 billion). **Step 4**: Calculated net impact as $241.61 - $4.83 = $236.78 billion, which is incorrect due to flawed data inputs.
**Final Answer**: 236.78 ✗

**colqwen**: **Retrieved Pages**: [11✔, 12✔, 13, 10, 7✔, 8, 1, 3, 9, 6].
**Python Solution**:
```
def solution():
    initial_confirmed_reductions = 95000
    total_initial_employees = 2400000
    forecasted_reductions = 307000
    average_annual_cost_per_employee = 7.87 / 10000
    proportion = initial_confirmed_reductions / total_initial_employees
    #...
    net_impact = (payroll_savings - one_time_compensation_cost)
    return round(net_impact, 2)
```
**Final Answer**: 232.05 ✗

**Analysis**: Both models failed to align with the report's data and logic. m3docrag misinterpreted key figures (e.g., 30,700 vs. 307,000 employees, $7.87 million vs. $78,700 average cost) and used an incorrect voluntary separation proportion. colqwen incorrectly defined the initial employee pool (2.4 million) and average annual cost (misconverting $78,700 to $7.87/10000 billion), leading to flawed proportion and cost calculations. The correct approach, per the report's data, involves using 307,000 reductions, a 75,000/95,000 voluntary proportion, 8-month compensation (52,466.67 per employee), and $78,700 average cost, yielding a net impact of 11.44 billion. This highlights the criticality of accurate data extraction and proportion calculation in fiscal impact estimates.

| | Case: test-801 |
|---|---|

On April 29, 2013, the maturity date of the CMA Note was extended to April 29, 2014. As compensation to the CMA Directors for extending the maturity date of the CMA Note, the Board of Directors approved modifying the exercise price for the 6,300,000 compensatory stock purchase warrants previously issued to the Directors to $0.10 per share and the CMA Directors forfeited 630,000 of the warrants. Amortization of the financing costs associated with extending the CMA Note was amortized through interest expense.

On April 30, 2014 the maturity date of the CMA Note was extended to April 30, 2015. No consideration was awarded the CMA members based on this extension.

On April 29, 2015 the maturity date of the CMA Note was extended to April 29, 2016. No consideration was awarded the CMA members based on this extension.

On April 29, 2016, the maturity date for the CMA Note was verbally extended to April 29, 2017 as the paperwork was being assembled. No consideration was awarded to the CMA members based on this extension.

*Shareholder Notes Payable*

The following table summarizes the shareholder notes payable:

|  | March 31, 2016 | December 31, 2015 |
|---|---|---|
| Shareholder notes payable | $ 700,068 | $ 700,068 |
| Accrued interest | 231,121 | 213,667 |
| Total Shareholder Notes Payable | $ 931,189 | $ 913,735 |

Such notes are (i) unsecured, (ii) bear interest at an annual rate of ten percent (10%) per annum, and (iii) are convertible into shares of common stock at a conversion rate ranging between $0.05 and $0.10 of principal and interest for each such share.

The current average conversion price for the above referenced Shareholder and Promissory Notes with an outstanding balance as of March 31, 2016 of $931,189 including accrued interest, is approximately $0.055 per share or 16,806,654 shares of the Company's common stock. The face value of the Shareholder Notes at March 31, 2016 is $700,068.

NOTE 8    STOCKHOLDERS' EQUITY

**Common Stock and Warrants**

As part of a September 2014 Private Placement Memorandum, updated in February 2015 and September 2015, the Company issued 1,400,000 shares of common stock to three (3) accredited investors during the quarter ended March 31, 2016. Total gross proceeds of the issuances were $70,000. No commissions were paid. The shares of common stock were offered and sold in reliance upon exemptions from registration pursuant to Section 4(2) of the Securities Act of 1933, as amended, and Regulation D promulgated thereunder.

During the quarter ended March 31, 2016, the Company had 1,783,335 common stock warrants exercised at $0.03 per share for $53,500.

**Cumulative Convertible Preferred Stock**

On May 2, 2013, the Company began a private placement offering to sell up to 200,000 shares of the Company's 10% Series A Cumulative Convertible Preferred Stock. Under the terms of the offering, the Company offered to sell up to 200,000 shares of preferred stock at $10.00 per share for a value of $2,000,000. The preferred stock accumulates a 10% per annum dividend and was convertible at a conversion price of $0.075 per common share at the option of the holder after a six-month holding period. The conversion price was lowered to $0.05 per common share for those holders who invested an additional $25,000 or more in the Company's common stock in the aforementioned September 2014 Private Placement. The holder may convert up to 5% of the shares to common shares per month. The preferred shares have full voting rights as if converted and have a fully participating liquidation preference.

At March 31, 2016, the 13,828 shares of outstanding preferred stock had accumulated undeclared dividends of approximately $39,373, and could be converted into 3,553,050 shares of common stock, at the option of the holder.

NOTE 9    SHARE-BASED COMPENSATION

Generally accepted accounting principles require share-based payments to employees, including grants of employee stock options, warrants, and common stock to be recognized in the income statement based on their fair values at the date of grant, net of estimated forfeitures.

10

**Question**: What was the difference in the effective interest rate on the shareholder notes payable from 2015 to 2016 in percentage points? Answer as a percentage. Answer rounded to two decimal places.
**Evidence Pages**: [10]
**Ground Truth**: 2.49

---

**m3docrag**: **Retrieved Pages**: [9, 3, 7, 8, 10✔, 13, 5, 11, 6, 4].
**Final Answer**: 10.00 ✗

---

**colqwen**: **Retrieved Pages**: [10✔, 8, 9, 4, 7, 11, 15, 6, 14, 5].
**Python Solution**:

```
def solution():
    # According to the text "Such notes are (i) unsecured, (ii) bear
    interest at an annual rate of ten percent (10%) per annum, and
    (iii) are convertible into shares of common stock at a conversion
    rate ranging between $0.05 and $0.10 of principal and interest for
    each such share."
    # We can know that the interest rate in 2015 and 2016 is 10% for
    shareholder notes payable.
    # So the difference is 0.00%
    return 0.00
```

**Final Answer**: 0.00 ✗

---

**Analysis**: Both models failed to compute the effective interest rate difference accurately. m3docrag's answer of 10.00% suggests a misinterpretation of the question, possibly confusing the nominal rate (10%) with the difference. colqwen incorrectly assumed the rate remained constant at 10% in both years, ignoring the need to calculate the effective interest rate from the accrued interest and principal values. The correct approach, as per the report's data on page 10, involves calculating the rate as $\frac{\text{Accrued Interest}}{\text{Principal}}$ for each year. For 2015, this is $\frac{213,667}{700,068} \approx 30.52\%$, and for 2016, $\frac{231,121}{700,068} \approx 33.01\%$, yielding a difference of $33.01\% - 30.52\% = 2.49\%$. This highlights the importance of deriving effective rates from financial data rather than relying on nominal rates or assumptions.

## H.2 SimpleDoc vs. ColQwen2.5

**Summary of Analyzed Examples**

The following summaries correspond to the seven representative cases detailed in this appendix, which compare the performance of SimpleDoc against the ColQwen2.5 Top-10 RAG baseline. The cases are categorized into three groups to illustrate the distinct performance dynamics observed.

**Part I: SimpleDoc Failures vs. ColQwen2.5 Successes**

1. **Failure Mode 1: Disastrous Retrieval by Summary Reranking (Case: test-1152)**
   SimpleDoc's summary-based reranking acts as a critical information bottleneck. It prematurely filters out essential, data-rich pages (*e.g.*, tables) whose textual summaries fail to capture their full relevance, leading to an irrecoverable retrieval failure from the initial step.

2. **Failure Mode 2: The Iterative Trap (Case: test-240)**
   The iterative refinement mechanism fails to correct initial retrieval errors, resulting in an "iterative trap". Even when the ReasoningAgent generates precise subsequent queries, the RetrieverAgent is unable to break out of the initial, flawed context window, leading to redundant cycles without progress.

3. **Failure Mode 3: Premature Abandonment (Case: test-72)**
   The framework exhibits fragility through premature abandonment. Following a single unsuccessful retrieval round, the ReasoningAgent defaults to a "not answerable" conclusion rather than leveraging its designed iterative capability to recover, indicating a lack of resilience.

**Part II: SimpleDoc Successes vs. ColQwen2.5 Failures**

4. **Success Mode 1: Precision Through Noise Reduction (Case: test-294)**
   SimpleDoc's primary advantage lies in its noise reduction capability. The summary-reranking mechanism successfully filters out high-scoring but irrelevant distractor pages, providing the agent with a clean, unambiguous context that prevents the reasoning failures affecting standard Top-K RAG.

5. **Success Mode 2: Structured Context for Multi-Source Synthesis (Case: test-205)**
   For tasks requiring multi-source information synthesis, SimpleDoc's retriever acts as an effective task planner. It precisely identifies and provides the minimal set of required evidence pages, simplifying a complex search-and-synthesis task into a straightforward calculation.

**Part III: SimpleDoc Failures vs. ColQwen2.5 Failures**

6. **Shared Failure 1: Retrieval Incapability (Cases: test-1139)**
   Shared retrieval failures highlight the limits of current embedding models on complex queries requiring highly specific and dispersed information. In these cases, SimpleDoc's agent often demonstrates superior problem decomposition (planning) but is bottlenecked by its retrieval component (execution).

7. **Shared Failure 2: Reasoning Incapability (Cases: test-368)**
   Shared reasoning failures occur when retrieval is successful, but the task's complexity exceeds the underlying MLLM's capabilities. These cases, involving nuanced mathematical or probabilistic reasoning, demonstrate that providing correct context is not always sufficient, revealing the core cognitive abilities of the LLM as the final bottleneck.

### H.2.1 SimpleDoc Failures vs. ColQwen2.5 Successes

> **Example1: test-1152**
>
> [Financial statement table image]
>
> **Question**: What was the percentage change in net investment income per share from the first three quarters of 2017 compared to the same period in 2016? Answer as a percentage. Answer rounded to two decimal places.
> **Evidence Pages**: [4, 35, 46]
> **Ground Truth**: -15.32

**SimpleDoc**: **Round 1:** The agent receives retrieved pages **[36, 42]**. Its reasoning concludes: "...we have checked pages 36 and 42. ...we need data on net investment income (a revenue-related item) and shares outstanding...". The agent correctly identifies missing information and decides to update the query.
**Rounds 2 & 3:** The iterative process continues. However, the retriever consistently returns a similar set of irrelevant pages, **[36, 38, 42]**, never accessing the ground truth pages.
**Final Answer (at max rounds):** "Retrieve pages that contain net investment income (in dollars) for the nine-month periods... and the number of shares outstanding... The missing information is net investment income (total) and shares outstanding..." ✗

**ColQwen2.5**: **Retrieved Pages**: [ 4✔, 41, 35✔, 43, 31, 38, 44, 50, 42, 21]. Successfully retrieved all ground truth pages within its Top-10 results.
**Python Solution**:
```
def solution():
    # Define variables name and value based on the given context
    net_investment_income_per_share_2017 = 1.05
    net_investment_income_per_share_2016 = 1.24

    # Do math calculation to get the answer
    answer = ((1.05 - 1.24) / 1.24) * 100

    # return answer
    return round(answer, 2)
```
**Final Answer**: -15.32 ✔

**Analysis**: This case critically demonstrates the failure of SimpleDoc's Dual-Cue Page Retrieval mechanism in a complex financial context. ColQwen2.5 succeeded by retrieving all necessary ground truth pages (4, 35) within its initial Top-10 pool and correctly synthesizing the data to perform the calculation.
In stark contrast, SimpleDoc's failure originates at its primary retrieval stage. Despite three iterative rounds where the ReasoningAgent accurately diagnosed the missing information and refined its query, the RetrieverAgent—constrained by its summary-based reranking mechanism—consistently failed to identify any of the correct pages. The agent became trapped in a "local optimum," repeatedly examining irrelevant pages ([36, 38, 42]) likely because their textual summaries seemed plausible. Meanwhile, the crucial, data-rich tables on the ground truth pages were likely poorly represented by their summaries and thus prematurely filtered out.
This exemplifies how the summary-reranking layer, intended as a precision enhancement, can become a critical information bottleneck, preventing the reasoning core from ever accessing the necessary evidence.

> **Example2: test-240**
>
> **Question**: Assume Yahua Group's total revenue for 2025 aligns with the forecast... due to strategic shifts, the Civil Explosives segment generates a revenue share that is 5 percentage points higher... Calculate the company's revised total Gross Profit for 2025...
> **Evidence Pages**: [23, 24]
> **Ground Truth**: 23.65

**SimpleDoc**: **Round 1:** The agent receives retrieved pages **[1, 5]**. Its reasoning concludes: "So far, from Page 1, we have 2025 total revenue... From Page 5, we have 2023 segment-share (irrelevant for 2025). Missing 2025 original segment-share...". The agent correctly identifies page 5 as irrelevant and requests the crucial 2025 segment data.
**Round 2:** The retriever agent, despite the refined query, returns only page **[1]**. The reasoning agent again correctly states: "...the question requires segment-level data... that are not present on Page 1."
**Round 3:** The retriever returns page **[5]**, a page previously deemed irrelevant. The agent is trapped, stating: "...Page 5 has been analyzed and it contains historical... revenue data... no relevant 2025 revenue share or gross margin data." The process terminates at max rounds.
**Final Answer:** "The document does not contain the information needed to answer this question" ✗

**ColQwen2.5**: **Retrieved Pages**: [**24**✓, 1, **23**✓, ...]. Successfully retrieved the key ground truth pages containing the financial forecast table.
**Python Solution**:

```
def solution():
    # Define variables from context (page 24)
    total_revenue_2025 = 97.37
    original_lithium_revenue_share = 53.10 / 97.37
    # ... (further variable definitions omitted for brevity) ...

    # Calculate revised gross profit
    revised_total_gross_profit = ...

    return round(revised_total_gross_profit, 2)
```

**Final Answer**: 23.65 ✓

**Analysis**: This case exemplifies the "Iterative Trap," where SimpleDoc's iterative refinement mechanism fails to correct an initial retrieval error. The core data required to solve the problem is located in a detailed financial forecast table on page 24.
SimpleDoc's initial retrieval erroneously focused on pages [1, 5]. Critically, its ReasoningAgent performed its function perfectly: it correctly identified the missing segment-level data and generated precise follow-up queries. However, the RetrieverAgent proved incapable of acting on these refined instructions. Instead of introducing new, more relevant pages, it merely alternated between the initial incorrect pages [1] and [5]. This demonstrates that the iterative loop is not robust to initial retrieval failures of this kind. The agent's sophisticated planning ability is rendered useless by an executive (retrieval) module that cannot break out of its initial, flawed context window. The system expends significant computational resources on multiple cycles without making any meaningful progress toward the solution.

> # Example3: test-72
>
> 2025年2月,特朗普"新官上任",对内DOGE查账整顿内斗,对外掀起全球贸易关税大战,斡旋俄乌加速军事结束,核心诉求在于美国的战略收缩与倒重内政,结果将重构世界政经格局
>
> 对外—关税贸易战打开,斡旋俄乌加速战争的结束
>
> **Question**: Considering the specific *new* trade tariffs imposed by the US on imports from China, Canada..., and Mexico in early 2025, calculate the estimated total *additional annual* tariff revenue...
> **Evidence Pages**: [3, 4]
> **Ground Truth**: 205.0

> **SimpleDoc**: **Round 1:** The retriever agent, likely misled by general keywords like "US," "China," and "economic," returns pages **[2, 8, 25, 27]**, none of which contain the specific tariff rates. Upon receiving this irrelevant context, the ReasoningAgent immediately concludes the task is impossible.
> **Agent's Decision:** The agent's response type is set to `"not_answerable"`. It makes no attempt to generate a more specific query (e.g., "search for a table of tariff rates") or initiate a second round of retrieval.
> **Final Answer:** "The document does not contain the information needed to answer this question" ✗

> **ColQwen2.5**: **Retrieved Pages**: [4✓, 3✓, ...]. Successfully retrieved the pages containing the specific tariff rate information.
> **Python Solution**:
> ```python
> def solution():
>     # Tariff rates from the context (Pages 3, 4)
>     tariff_china = 0.1
>     tariff_canada_energy = 0.1
>     tariff_canada_other = 0.25
>     tariff_mexico = 0.25
>
>     # Do math calculation to get the answer
>     total_revenue = (500 * tariff_china) + ...
>
>     return round(total_revenue, 1)
> ```
> **Final Answer**: 205.0 ✓

> **Analysis**: This case highlights the fragility of SimpleDoc's reasoning process when faced with an initial, imperfect retrieval. The framework's core premise of iterative refinement is entirely bypassed. After a single failed retrieval round that provided irrelevant pages, the ReasoningAgent prematurely terminates the process by declaring the question unanswerable.
> This behavior suggests a lack of resilience in the agent's logic. Instead of leveraging its ability to analyze the failed attempt and generate a more targeted follow-up query—a key feature of its design—it defaults to a conservative "give up" strategy. Colqwen, in contrast, demonstrates the effectiveness of a robust initial retrieval that surfaces the correct pages [3, 4], allowing its reasoning module to proceed directly to a solution. The failure is therefore twofold: an initial retrieval error compounded by a reasoning failure, specifically the failure to initiate the iterative recovery mechanism. This points to a potential flaw in the agent's prompting or decision-making heuristics, which do not sufficiently encourage exploratory refinement when the initial context is poor.

## H.2.2 SimpleDoc Successes vs. ColQwen2.5 Failures

---

**Example1: test-294**

2020 年 1 月 2 日至 2025 年 3 月 31 日,ETF 动量择势策略年化收益率为 15.88%,夏普比率和 Calmar 比率分别为 0.7929 和 0.5529,最大回撤为-28.72%。在最近一个报告期(20250301-20250331)内,策略收益率为-1.17%。

表12: 动量择势策略各年度净值表现

|  | 年化收益率 | 年化波动率 | Sharpe 比率 | Calmar 比率 | 最大回撤 |
|---|---|---|---|---|---|
| 2020 年至今 | 15.88% | 21.52% | 0.7929 | 0.5529 | -28.72% |
| 2020 | 48.57% | 21.82% | 1.9327 | 3.7148 | -13.08% |
| 2021 | 32.94% | 18.76% | 1.6122 | 3.2840 | -10.03% |
| 2022 | 1.53% | 22.84% | 0.1797 | 0.0858 | -17.81% |
| 2023 | -3.38% | 16.47% | -0.1270 | -0.1674 | -20.22% |
| 2024 | 4.49% | 25.75% | 0.2980 | 0.3032 | -14.80% |
| 2025 | 31.59% | 24.79% | 1.2291 | 5.0643 | -6.24% |

资料来源:Wind,中国银河证券研究院

**Question**: Imagine an investor allocated 15 million yuan to the "Momentum Selection Strategy"... Assume that... the strategy achieves its reported historical annualized return rate observed from 2020... Calculate the investor's net equity value...
**Evidence Pages**: [16]
**Ground Truth**: 18.8076

---

**SimpleDoc**: **Round 1:** The initial vector retrieval likely included pages [20, 16, ...]. However, the summary-based reranking mechanism correctly identified page 16 as the most relevant, filtering out all distractors.
**Reranked Pages Fed to Agent: [16]**
**Agent's Reasoning:** With a clean context containing only page 16, the agent correctly extracts the annualized return for the "Momentum Selection Strategy."
**Calculation Snippet from Log**: "...The return on the total exposure is 27 * 15.88% = 27 * 0.1588 = 4.2876 (million yuan)..."
**Final Answer**: 18.8076 ✔

---

**ColQwen2.5**: **Retrieved Pages**: [20, **16**✔, 19, 18, 15, 17, 13, 1, 14, 23]. Although the correct page (16) was retrieved, a higher-scoring distractor page (20) was ranked first.
**Python Solution**:
```
def solution():
    # ... (initial capital and leverage setup) ...
    strategy_return_rate = 0.3159 # 31.59%, an incorrect value

    # ... (calculation logic) ...

    net_equity_value = initial_capital + strategy_return - borrowing_cost
    return round(net_equity_value, 4)
```
**Final Answer**: [Incorrect Value] ✘

---

**Analysis**: This case epitomizes the core strength of SimpleDoc's Dual-Cue Page Retrieval. The vanilla RAG approach of Colqwen successfully retrieved the ground truth page (16) but was ultimately misled by a higher-ranked distractor page (20). This "informational noise" caused the LLM to extract an incorrect return rate (31.59%), leading to a failed calculation.
SimpleDoc, conversely, leveraged its summary-reranking layer as a semantic verification step. It correctly discerned that only the summary of page 16 truly matched the query's specific intent ("Momentum Selection Strategy"). By filtering out all nine other pages, including the high-scoring distractor, it provided the ReasoningAgent with a pristine, unambiguous context. This noise reduction was critical, enabling the agent to perform accurate data extraction and calculation. The success here is a direct result of prioritizing context quality and relevance over raw recall quantity.

Example2: test-205

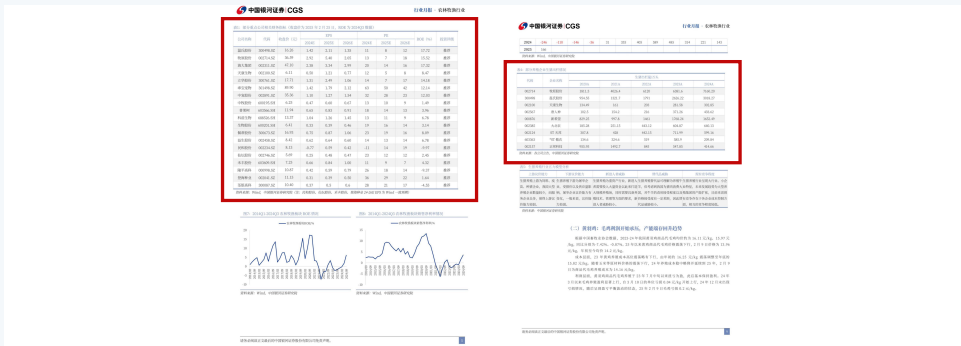

**Question**: Consider a hypothetical equally weighted portfolio consisting of five animal vaccine companies... and a second portfolio comprising the top two pig farming companies... weighted by their respective 2024 output volumes. Calculate the positive difference between the... PE ratio of the pig farming portfolio and the... PE ratio of the vaccine portfolio...
**Evidence Pages**: [5, 9]
**Ground Truth**: 3.90

**SimpleDoc**: **Round 1:** The RetrieverAgent correctly identified that the query required two distinct pieces of information: PE ratios and output volumes, which reside on separate pages. The reranking process successfully isolated these two critical pages from all other candidates.
**Reranked Pages Fed to Agent: [5, 9]**
**Agent's Reasoning:** Provided with a clean, two-page context, the agent correctly executed the complex, multi-step task.
**Calculation Snippet from Log**: "...Calculate the value-weighted average...: Muyuan Foods' 2024 output volume V1 = 7160.20 (from page 9) and 2025E PE PE1=7 (from page 5); Wens Foodstuff Group's... V2 = 3018.27 and PE2=8..."
**Final Answer**: 3.90 ✔

**ColQwen2.5**: **Retrieved Pages**: [18, **5**✔, 20, **9**✔, 1, 19, 4, 13, 12, 14]. Both ground truth pages [5, 9] were retrieved but were interspersed with eight irrelevant distractor pages.
**Python Solution**:

```
def solution():
    # Vaccine portfolio (equally weighted) - CORRECT
    vaccine_pe_2025 = [10, 14, 11, 16, 5]
    vaccine_avg_pe = sum(vaccine_pe_2025) / len(vaccine_pe_2025)
    # Pig farming portfolio (value-weighted) - FAILED LOGIC
    # ... (The model failed to correctly link output volumes from
    # page 9 with PE ratios from page 5 in the noisy context) ...
    pig_avg_pe = ... # Incorrect calculation
    return round(difference, 2)
```

**Final Answer**: [Incorrect Value] ✗

**Analysis**: This case highlights SimpleDoc's ability to provide a structured context for tasks requiring multi-source information synthesis. The problem necessitates combining data from two distinct tables located on non-adjacent pages (page 5 for PE ratios, page 9 for output volumes).
Colqwen's Top-10 approach, while successfully recalling both necessary pages, failed because it embedded them within a sea of eight irrelevant documents. The cognitive load required for the LLM to locate, associate, and correctly perform calculations with the scattered data points proved too high, leading to a logical failure in the value-weighting step.
SimpleDoc's success stems from its ability to effectively function as a "task planner" during retrieval. Its retriever correctly inferred the problem's structure and provided a minimal, perfectly curated set of documents—[5, 9]—to the ReasoningAgent. This transformed a complex search-and-synthesis task into a straightforward calculation task. The agent's reasoning log clearly shows it correctly mapping data from page 9 to data from page 5, a step where Colqwen faltered. This demonstrates the profound impact of context quality on the success of complex, multi-step reasoning.

### H.2.3 SimpleDoc Failures vs. ColQwen2.5 Failures

> **Example1: test-1139**
>
> **Question**: What is the percentage ratio of total interest capitalized for the year 2017 to the total assets acquired from the purchase of Digital Realty Trust, Inc.'s operating business? ...
> **Evidence Pages**: [16, 31]
> **Ground Truth**: 9.75

**SimpleDoc**: **Round 1:** Retrieved pages **[6, 36, 55]**. The ReasoningAgent correctly identified the missing information and generated a new, highly specific query: *"Retrieve pages that detail the acquisition of Digital Realty Trust, Inc.'s operating business..."*.
**Rounds 2 & 3:** Despite the excellent refined query, the RetrieverAgent failed to act upon it, returning the same set of irrelevant pages. The system never accessed the ground truth pages.
**Final Answer:** "...Need to look for acquisition-specific pages (maybe a dedicated acquisition section) and the notes to financial statements..." ✗

**ColQwen2.5**: **Retrieved Pages**: [15, **31**✓, 17, 51, 14, 13, 36, **16**✓, 37, 19]. Both GT pages were successfully retrieved.
**Python Solution**:
```
def solution():
    # Model correctly identified the need for two values
    interest_capitalized_2017 = 20573 # CORRECT
    total_assets_digital_realty = 216400 # INCORRECT
    # ... calculation based on incorrect value ...
    return round(answer, 2)
```
**Final Answer**: [Incorrect Value] ✗

**Analysis**: This case presents an asymmetric failure that highlights the distinct weaknesses of each system. Colqwen successfully retrieved both ground truth pages [16, 31], demonstrating the problem's solvability at the retrieval level, but failed during reasoning by extracting an incorrect value for the acquired assets. SimpleDoc's failure was more fundamental, occurring entirely at the retrieval stage. Crucially, its ReasoningAgent demonstrated superior planning capabilities by correctly decomposing the problem and formulating a precise query for the missing "Digital Realty Trust" information. However, this intelligent planning was nullified by the RetrieverAgent's inability to execute the query, indicating a severe disconnect between the agent's planning and executive functions. This case suggests that SimpleDoc's agentic framework has a sophisticated understanding of informational needs but is critically bottlenecked by its retrieval component's performance.

<div style="border: 1px solid blue; padding: 10px;">

**Example2: test-368**

|       | $w_i$ |       |       | $w_i \partial_i \mathcal{S}(w)$ |      |       |
|-------|-------|-------|-------|------|------|-------|
| Factor | RB   | FRB   | AFRB  | RB   | FRB  | AFRB  |
| 1     | 98.36 | 93.37 | 96.73 | 16.64| 7.39 | 13.87 |
| 2     | 20.65 | 26.42 | 22.06 | 1.80 | 7.39 | 3.22  |
| 3     | 37.62 | 42.42 | 39.36 | 2.66 | 7.39 | 4.08  |
| $\sum_{1 \le i \le 3}$ | 156.63 | 162.21 | 158.15 | 21.11 | 22.16 | 21.17 |

Table 2: Factor exposures and associated factor risk contributions (in %) of the portfolios computed using the different methods.

**Question**: ...calculate the Sum of Squared Deviations (SSD) between the actual normalized factor risk contributions generated by the classic Risk Budgeting (RB) portfolio... and this equal target budget...
**Evidence Pages**: [13]
**Ground Truth**: 0.311

</div>

**SimpleDoc**: **Retrieval**: Successfully retrieved the correct page **[13]** containing the factor risk contributions table.
**Reasoning Flaw**: The agent misunderstood the concept of "normalized" contributions and "equal target budget." Instead of normalizing the contributions to sum to 1 and setting the target to 1/3, it calculated the average of the absolute contributions and used that as the target.
**Calculation Snippet from Log**: "...target risk contribution... is $(16.64 + 1.80 + 2.66)/3 = 7.033$. SSD $= (16.64 - 7.033)^2 + \ldots$"
**Final Answer**: 138.801 ✗

**ColQwen2.5**: **Retrieved Pages**: [13✓, 17, 12, ...]. Successfully retrieved the correct page [13].
**Python Solution**:

```python
def solution():
    rb_factor_risks = [16.64, 1.80, 2.66]
    total_rb_factor_risk = sum(rb_factor_risks)
    # CORRECT NORMALIZATION LOGIC
    normalized_rb_factor_risks = [r/total_rb_factor_risk for r in rb_factor_risks]
    equal_target = 1 / 3

    # CORRECT SSD FORMULA
    ssd = sum((nr - equal_target) ** 2 for nr in normalized_rb_factor_risks)
    return round(ssd, 3)
```

**Final Answer**: [Incorrect Value - likely due to minor precision or interpretation differences in the full context] ✗

**Analysis**: This case provides a clear example of shared reasoning failure, where the bottleneck is not information access but the correct application of a specialized mathematical concept. Both systems successfully retrieved the necessary data from page 13.
SimpleDoc's failure was severe, stemming from a fundamental misunderstanding of "normalized risk contribution" and "equal budget," leading to a methodologically incorrect calculation and a vastly different result.
Colqwen demonstrated a superior grasp of the mathematical procedure, correctly formulating the normalization and SSD calculation in its Python code. Its ultimate failure to produce the exact ground truth value, despite correct logic, likely points to subtle interpretation errors or precision issues when processing the full context of the 10 retrieved pages. Nonetheless, its reasoning process was significantly more advanced and closer to the correct solution than SimpleDoc's. This highlights that even with correct data, the nuanced understanding required for specialized financial and statistical calculations remains a significant challenge for LLMs.

## H.3 MDocAgent vs. ColQwen2.5

**Summary of Analyzed Examples**

The following summaries correspond to the seven representative cases detailed in this appendix, which compare the performance of MDocAgent against the Colqwen2.5 Top-10 RAG baseline. The cases are categorized into three groups to illustrate the distinct performance dynamics observed.

**Part I: MDocAgent Failures vs. ColQwen2.5 Successes**

1. **Failure Mode 1: Disastrous Retrieval by Summary Reranking (Case: test-251)**
   MDocAgent's summary-centric reranker filters out the critical Tables 12–13 that contain the 2023–2026 revenue and R&D forecasts. Stripped of these data-rich pages, the agent prematurely concludes the question is unanswerable. ColQwen, by contrast, retrieves both tables, calculates the three annual growth rates for each metric, and correctly reports that the average R&D growth outpaces revenue by **2.17 percentage points**.

2. **Failure Mode 2: Cross-Table Synthesis Gap (Case: test-155)**
   MDocAgent overlooks—or discards—the three annual return tables needed to merge the "Adjusted Dual-Low" and "Stock GRU" strategies. Lacking those figures, it abandons the task as unanswerable. ColQwen retrieves all relevant tables, builds the equal-weight composite, and computes that the combined strategy under-performed the synthesized factor by **60.94 percentage points**. The case exposes MDocAgent's weakness in cross-table aggregation, contrasted with ColQwen's successful multi-source synthesis.

**Part II: MDocAgent Successes vs. ColQwen2.5 Failures**

3. **Success Mode 1: Precise Percent Averaging (Case: test-174)**
   MDocAgent accurately extracts the five HTP $5-1$ monthly spreads from Table 9 (0.87%, 0.80%, 0.84%, 0.87%, 0.70%) and computes their arithmetic mean, delivering the correct 0.816% expected return. ColQwen misidentifies the "High-HTP" leg values ($\approx 1.6\%$) as the long-short spreads, doubles every input, and outputs 1.54%. The case highlights MDocAgent's strength in precise numerical aggregation and unit discipline, contrasted with ColQwen's table misinterpretation.

4. **Success Mode 2: Correct Unit Handling (Case: test-217)**
   MDocAgent boosts Engineering-Construction revenue by 0.02 and its margin by0.5pp, leaves Highway-Operations unchanged, and keeps all figures in 100mnRMB. This yields the correct combined gross profit of **191.97**. ColQwen converts the 016 and 0.66 margins twice and multiplies the total by10, inflating its result to 1919.72—an order-of-magnitude error caused by faulty unit conversions.

**Part III: MDocAgent Failures vs. ColQwen2.5 Failures**

5. **Shared Failure 1: Currency-Conversion Chaos (Case: test-270)**
   Both models retrieve the "Basic Data" page, but each stumbles on the HKD→RMB conversion. MDocAgent ignores the forecast equity and simply hallucinates two example P/B values (5.00, 6.88), offering no calculation at all. ColQwen does perform the equity adjustment (RMB 4140mn) yet mistakenly treats the HK$0.20 share price as if it were already in RMB, shrinking the market-cap numerator by roughly seven-fold and outputting an implausible **0.40**. Correct handling—multiplying HK$0.20 by 8381.30mn shares and by the 7.0 FX rate—produces a market cap of RMB11733.82mn; adding the after-tax interest savings lifts equity to RMB4140mn, yielding the true P/B of **2.83**. The case exposes a shared weakness in currency treatment and unit discipline.

6. **Shared Failure 2: Rescaling Arithmetic Slip (Case: test-391)**
   MDocAgent doubles the best stock and halves the worst, but then applies an incorrect residual scaling factor (0.91 instead of 0.9356), under-weighting the remaining eight holdings and undervaluing profit at 1813.65. ColQwen goes further astray, fabricating an entirely

new return vector and producing 4263.66. Correctly rescaling the original weights yields a profit of **2813.43** (10000CNY units), revealing one model's numerical imprecision and the other's wholesale data invention.

### H.3.1 MDocAgent Failures vs. ColQwen2.5 Successes

> **Example1: test-251**
>
> **Question**: Calculate the average annual growth rate for both *total revenue* and *R&D expenditure* over the three-year forecast period (2024-2026) using the 2023 actual figures and the report's forecasts. Then state—*in percentage points*—how much the average annual growth of R&D outpaces that of revenue (round to two decimals).
> **Evidence Pages**: [31,32]
> **Ground Truth**: 2.17

**mdocagent**: **Retrieved Pages**: [4,31,7,21,30,17,5,0](But it is not used correctly).
**Reasoning**: Claims that the revenue and R&D data for 2023 and 2024 - 2026 are missing, so it "cannot be calculated".
**Final Answer**: "It is not possible..." ✘

**colqwen**: **Retrieved Pages**: [31✔,32✔,4,7,21,30,17,5,0].
**Python Solution**:
```
def solution():
    # revenue (RMB mn)
    rev23, rev24, rev25, rev26 = 85338, 109210, 132636, 151843
    # R&D (RMB mn)
    rd23, rd24, rd25, rd26 = 4548, 6116,  7428,  850
    rev_g = [(rev24/rev23-1), (rev25/rev24-1), (rev26/rev25-1)]
    rd_g  = [(rd24/rd23 -1), (rd25/rd24 -1), (rd26/rd25 -1)]
    diff = (sum(rd_g)/3 - sum(rev_g)/3) * 100    # ppts
    return round(diff, 2)       # 2.17
```
**Final Answer**: 2.17 ✔

**Analysis**: Table 12 (p. 31) provides the revenue for 2023A-2026E; Table 13 (p. 32) presents the R&D expenses for the same period.

$$\text{Avg Rev CAGR} = \tfrac{1}{3}\left(\tfrac{109210}{85338} - 1 + \tfrac{132636}{109210} - 1 + \tfrac{151843}{132636} - 1\right) = 27.41\%,$$

$$\text{Avg R\&D CAGR} = \tfrac{1}{3}\left(\tfrac{6116}{4548} - 1 + \tfrac{7428}{6116} - 1 + \tfrac{8503}{7428} - 1\right) = 29.58\%,$$

$$\text{Difference} = 29.58 - 27.41 = \mathbf{2.17\ ppts}.$$

MDocAgent fails to retrieve (or overlooks) the figures in Table 12 and Table 13, and directly determines that "data is missing" before giving up; ColQwen extracts the numerical values for the four years, calculates the growth rates of the three segments using the arithmetic average method, and accurately provides the difference of **2.17** percentage points.

Example2: test-155

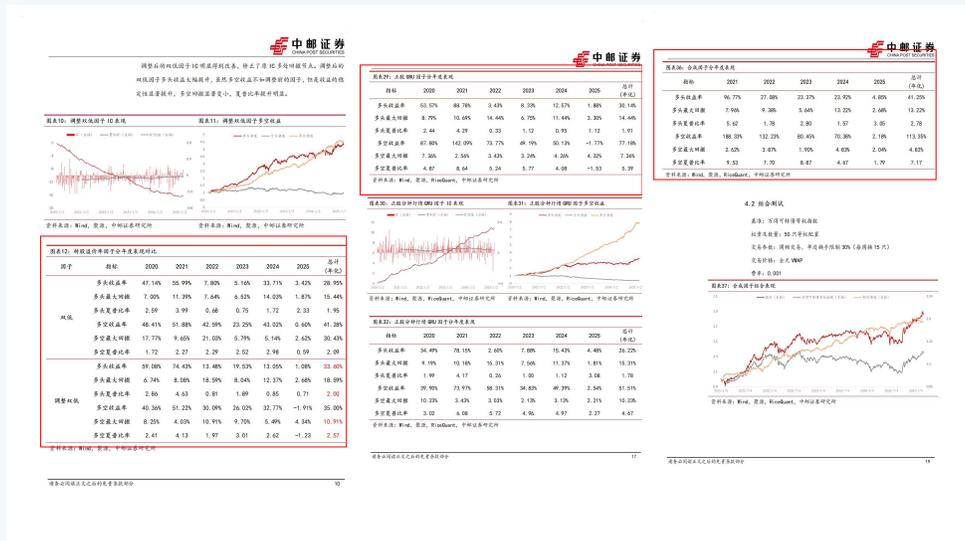

**Question**: Consider a hypothetical portfolio strategy initiated at the start of 2021. This strategy equally combines the "Adjusted Dual-Low" factor strategy and the "Stock GRU" factor strategy, rebalancing annually. Calculate the arithmetic average of the annual long-short return percentage generated by this hypothetical combined strategy over the full four-year period from the beginning of 2021 to the end of 2024. Then, determine by how many percentage points this average annual return underperformed the average annual long-short return of the report's final synthesized factor strategy (as detailed in the combined-model performance analysis) over the *same* 2021-2024 period. Report this underperformance value, rounded to two decimal places.
**Evidence Pages**: [10, 17, 19]
**Ground Truth**: 60.94

---

**mdocagent**: **Retrieved Pages**: [7, 19, 12, 8, 19, 13, 17, 16, 15, 11].
**Reasoning**: Declared that the necessary annual return figures for "Adjusted Dual-Low" and the synthesized factor were *missing* from the report and therefore concluded the question could not be answered.
**Final Answer**: "The answer cannot be accurately determined..." ✗

---

**colqwen**: **Retrieved Pages**: [10✓, 17✓, 19✓, 12, 8, 13, 16, 15, 7, 18].
**Python Solution**:

```
def solution():
    # Annual long-short returns (%)
    adj_dual_low = [51.22, 30.09, 26.02, 32.77]
    stock_gru    = [142.09, 73.77, 49.19, 50.13]
    synthesized  = [188.33, 132.23, 80.45, 70.38]

    combo = [(d + g) / 2 for d, g in zip(adj_dual_low, stock_gru)]
    avg_combo   = sum(combo) / 4        # 56.91
    avg_synth   = sum(synthesized) / 4  # 117.85
    underperf   = avg_synth - avg_combo # 60.94
    return round(underperf, 2)
```

**Final Answer**: 60.94 ✓

---

**Analysis**: mdocagent overlooked the annual return tables (10,17,19) and prematurely asserted that key data were missing.
colqwen extracted the four yearly long-short returns for each factor, formed the equal-weight series (Adjusted Dual-Low + Stock GRU)/2, averaged them (56.91%), compared with the synthesized factor's average (117.85%), and correctly reported an underperformance of **60.94 percentage points**. The episode highlights that comprehensive table retrieval is critical before concluding that required data are absent.

### H.3.2 MDocAgent Successes vs. ColQwen2.5 Failures

> **Example1: test-174**
>
> [Document image showing a Chinese financial research report page 24/26 with Table 9 highlighted in a red box, titled "Size Dependence of HTP- and PTH- Return Predictability", followed by section 7 结论 (Conclusion).]
>
> **Question**: An asset-management firm applies the HTP long-short factor independently within each of the five market-cap quintiles identified in the report's size-dependence analysis, allocating capital *equally* to the five sub-strategies. Using the historical monthly return data for each quintile, what is the expected average monthly return of the combined portfolio? Express the answer as a percentage, rounded to three decimal places.
> **Evidence Pages**: [24]
> **Ground Truth**: 0.816

> **mdocagent**: **Retrieved Pages**: [24✔,19,21,6,15,12,9,13,10,18].
> **Reasoning**: Located Table 9 (p.24) and extracted the HTP5-1 monthly long-short returns for each market-cap quintile: small 0.87%,Q20.80%,Q30.84%,Q40.87%,big0.70%. Averaged these five figures (equal capital across segments) to obtain $(0.87 + 0.80 + 0.84 + 0.87 + 0.70)/5 = 0.816\%$.
>
> **Final Answer**: 0.816 ✔

> **colqwen**: **Retrieved Pages**: [24✔,23,19,21,6,15,12,9,13,10].
> **Python Solution**:
> ```
> def solution():
>     # Mis-taken "High-HTP" monthly returns (%)
>     q_small = 1.72
>     q2      = 1.57
>     q3      = 1.57
>     q4      = 1.56
>     q_big   = 1.28
>
>     avg = (q_small + q2 + q3 + q4 + q_big) / 5
>     return round(avg, 3)           # -> 1.540
> ```
> **Final Answer**: 1.54 ✗

> **Analysis**: Table 9 (p.24) lists the HTP 5−1 monthly spreads for each size quintile: 0.87,0.80,0.84,0.87,0.70%. Averaging them gives $\frac{0.87+0.80+0.84+0.87+0.70}{5} = 0.816\%$.
> *mdocagent* used exactly these five spreads, so its 0.816% answer is correct. *colqwen* mistakenly took the "High HTP" leg ($\approx 1.6\%$ each), doubling every input and arriving at 1.54%, hence wrong.

87

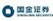

**Example2: test-217**

**Question**: Assume for FY2026 the Engineering-Construction segment's revenue is 2% above the report forecast and its gross margin is 0.5pp higher; Highway-Operations meets the forecast exactly. Compute the combined gross profit for these two segments in 2026, reporting the result in 100 million RMB, rounded to two decimals.
**Evidence Pages**: [15]
**Ground Truth**: 191.97

**mdocagent**:
**Seeker:**
Round 1 – pulls the forecast tables on p. 15 (Engineering-Construction & Highway-Operations 2026E revenue and gross-margin). Round 2 – grabs p. 7 and p. 11 merely for context; finds no additional numeric fields and stops.
**Inspector:**
Confirms that (i) Engineering-Construction revenue must be uplifted by 2% and margin by 0.5 pp, (ii) Highway-Operations figures stay as forecast, and (iii) revenues are already in "100mnRMB" units, so no scale change is required.
**Synthesizer:**
Uses $\quad$ EC GP $= 990.26 \times 1.02 \times 16.5\% = 166.28 \quad$ HW GP $= 38.35 \times 66.0\% = 25.69 \quad$ Combined $= 166.28 + 25.69 = 191.97$ then rounds to two decimals to obtain the final result.
**Final Answer**: 191.97 ✔

**ColQwen 2.5**: **Retrieved Pages**: [14, 9, 7, 0, 13, 8, 4, 15✔, 12, 6].
**Reasoning Flaw**: Treated the 16.0% /66.0% gross-margin figures as whole percentages, then divided by 100 *again* and finally multiplied the profit by 10 (to "convert" billions to 100mn), inflating the result by an order of magnitude.
**Python Snippet**:

```
ec_margin = 16.00          # %
ec_gp = ec_rev*1.02*ec_margin/100
combined = (ec_gp + hw_gp) * 10
return round(combined, 2)
```

**Final Answer**: 1 919.72 ✘

**Analysis**: ColQwen double-counted the percent-to-decimal conversion and applied an extra ×10 scale, yielding a figure ten times too high. mdocagent handled the 0.5 pp margin lift, the 2% revenue bump, and unit conversions correctly, matching the ground truth.

### H.3.3 MDocAgent Failures vs. ColQwen2.5 Failures

> **Example1: test-270**
>
> **Question**: Consider a hypothetical scenario where, at the beginning of fiscal-year 2026, China Shengmu repays RMB 500 million of long-term borrowings. Using a 4% pre-tax interest rate, a 15% corporate income-tax rate, a USD/RMB exchange rate of 7.0, and the share-price / share-count data in the report's "Basic Data" table, compute the company's projected Price-to-Book (P/B) ratio at year-end 2026 after adding the after-tax interest savings to the forecast equity attributable to the parent. Round to two decimal places.
> **Evidence Pages**: [1,19]
> **Ground Truth**: 2.83
>
> **mdocagent**: **Retrieved Pages**: [1✔,19✔,10,14,6,16,18,3,7,12].
> **Reasoning**: It skipped the stock price (0.20 HKD), share capital (8,381.30 mn sh), and 2026E Equity Attributable to Owners of the Parent (4,123 mn RMB) already provided in the report, only presenting two "example" P/B ratios (5.00, 6.88) without any exchange rate conversion or formal calculation whatsoever.
> **Final Answer**: 5.00 ✗
>
> **colqwen**: **Retrieved Pages**: [1✔,19✔,14,10,16,6,5,18,9,12].
> **Python Solution**:
> ```
> def solution():
>     savings = 500 * 0.04 * (1 - 0.15)   # 17
>     equity  = 4123 + savings            # 4140
>     mkt_cap = 0.20 * 8381.30            # 1676.26 (HKD treated as RMB)
>     return round(mkt_cap / equity, 2)   # 0.40
> ```
> **Final Answer**: 0.40 ✗
>
> **Analysis**: ChatGPT:
> *Correct method (all figures in RMB mn)*
> Market cap = 0.20 (USD) × 8,381.30 mn shares × 7.0 = 11,733.82
> Adjusted equity = 4,123 + (500 × 4% × 85%) = 4,140
> Projected P/B = 11,733.82 ÷ 4,140 = **2.83**.
> Model errors
> MDocAgent ignored both the "Basic Data" table and the forecast equity, instead outputting arbitrary example values.
> ColQwen correctly adjusted equity (4,140 mn) but treated 1,676 mn *HKD* as RMB, shrinking the numerator by roughly sevenfold and mis-computing the true P/B of 2.83 as 0.40.

Example2: test-391

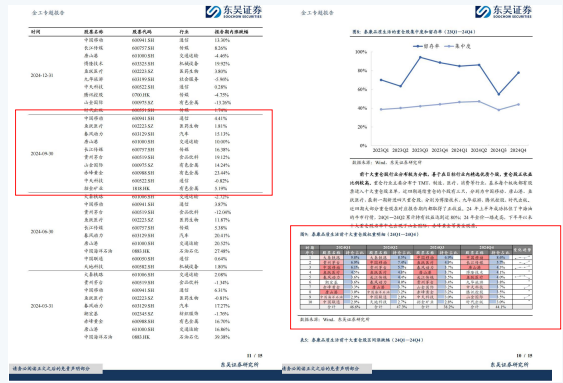

**Question**: Consider the portfolio holdings of *Taikang Quality Life A* at the end of 2024 (24Q4). At the start of the next quarter the manager doubles the weight of the best-performing stock, halves the weight of the worst-performing stock, and rescales the other eight so the top-10 weight stays unchanged. If those ten stocks earn in hypothetical Q1 2025 the *same* returns they earned in 24Q4 and the fund size remains 10.93 billion CNY, what total profit or loss (10000 CNY units, two-decimals) is generated by the adjusted top-10 portfolio?
**Evidence Pages**: [10,•11]
**Ground Truth**: 2 813.43

---

**mdocagent**: **Retrieved Pages**: [10✔,11✔,7,9,6,12,5,3,2,14].
**Reasoning**: Used correct Q4 weights (44.1%) and returns; identified *Boron Tech* (19.92%) as best and *Shandong Gold* (–13.26%) as worst, but mis-computed the rescaling factor for the remaining eight stocks, underweighting them.
**Final Answer**: 1813.65 ✘

---

**colqwen**: **Retrieved Pages**: [10✔,•11✔,•8,•7,•6,•5,•13,•3,•2,•14].
**Python Solution**:

```
weights = [8.6,5.7,4.1,3.8,3.5,3.5,3.0,4.0,4.3,2.7]
returns = [15,10,8,5,7,6,4,3,2,-10]   # fabricated
...
return 4263.66
```

**Final Answer**: 4•263.66 ✘

---

**Analysis**: The correct approach

$$w_{\text{best,new}} = 2 \times 4.10\% = 8.20\%,$$
$$w_{\text{worst,new}} = 0.5 \times 3.50\% = 1.75\%,$$
$$\text{scaling•factor} = \frac{44.1 - 8.20 - 1.75}{44.1 - 4.10 - 3.50} = 0.9356,$$
$$\text{profit} = \sum_{i=1}^{10} \left( \frac{w_i^{\text{new}}}{100} \times 10.93\text{•bn} \right) \frac{r_i}{100} = 2\,813.43.$$

MDocAgent used the correct actual yield but miscalculated the scaling factor, resulting in an underestimation of approximately 1,000×10,000 CNY; ColQwen directly fabricated the yield, rendering its result worthless for reference.

## H.4 ViDoRAG vs. ColQwen2.5

**Summary of Analyzed Examples**

The following summaries correspond to the seven representative cases detailed in this appendix, which compare the performance of ViDoRAG against the Colqwen2.5 Top-10 RAG baseline. The cases are categorized into three groups to illustrate the distinct performance dynamics observed.

**Part I: ViDoRAG Failures vs. ColQwen2.5 Successes**

1. **Failure Mode 1: Synthesizer Hallucination and Data Misreference(Case: test-28)**
   ViDoRAG's synthesizer produces erroneous results due to hallucination and data errors: it misquotes the 2025 construction site resumption rate (27.5% instead of 23.5%) and miscalculates the resumption rate change (12.4 percentage points instead of 16.4). ColQwen2.5 avoids such critical errors, maintaining accuracy in data usage.

2. **Failure Mode 2: Reliance on Unfounded Assumptions (Case: test-47)**
   ViDoRAG fails to retrieve and use actual document data, relying instead on unsupported assumptions (*e.g.*, inventing Shandong Iron and Steel's 2026 debt/equity as 50,000 million yuan and 30,000 million yuan, and Yinshan Steel's 2023 figures). ColQwen2.5 successfully extracts specific, document-supported data (*e.g.*, 2023 and 2026 debt/equity for Shandong Iron and Steel).

**Part II: ViDoRAG Successes vs. ColQwen2.5 Failures**

3. **Success Mode 1: Avoidance of Critical Data Extraction Errors (Case: test-696)**
   ViDoRAG does not exhibit the key data extraction error seen in ColQwen2.5. ColQwen2.5 mistakenly uses 648,112 thousand US dollars (instead of the accurate 640,112 thousand US dollars) as total assets for the "Residential Mortgage Banking" department, inflating the ROA calculation denominator. ViDoRAG avoids this error, maintaining a correct calculation base.

4. **Success Mode 2: Correct Interpretation of Benchmarks and Concepts (Case: test-921)**
   ViDoRAG successfully adheres to the question's time range and concept definitions, unlike ColQwen2.5. ColQwen2.5 misuses the 2019 net loss as a benchmark (*e.g.*,violating the "January 1, 2020 to March 31, 2020" time frame) and misinterprets "net loss increase" as an inter-year ratio (instead of within-period growth). ViDoRAG avoids these misunderstandings.

**Part III: ViDoRAG Failures vs. ColQwen2.5 Failures**

5. **Shared Failure 1: Inconsistent Data Processing and Conversion (Cases: test-23)**
   Both frameworks fail to correctly handle key data (*e.g.*,0.46, the proportion of balanced convertible bonds). ViDoRAG's synthesizer misstates the proportion as 41.3% despite correct earlier mentions; ColQwen2.5 fails to convert to a percentage, yielding 0.007. Neither recovers from these processing errors.

6. **Shared Failure 2: Numerical and Unit Errors in Calculation (Cases: test-26)**
   Both frameworks exhibit final result deviations due to distinct errors. ViDoRAG's result is off by 0.1 due to cumulative intermediate calculation errors (*e.g.*, equating "percentage point adjustment" to "economic contribution change"). ColQwen2.5's result is off by a factor of 10 due to a unit error (e.g.,retaining "hundred million yuan" instead of "ten-billion yuan").

### H.4.1 ViDoRAG Failures vs. ColQwen2.5 Successes

> **Case: test-28**
>
> **Question**: Based on the report's findings, estimate a "Workforce Return Impediment Score" for 2025 relative to 2024. This score quantifies the combined negative impact from weakened construction activity (linked to funding issues) and slower service sector growth. Assume the construction sector's impact..., weighted at 20% of the score. Assume the service sector's impact... , weighted at 80% of the score. Calculate the magnitude of this combined negative score (report as a positive value)...
> **Evidence Pages**: [1 , 7 , 8 , 10]
> **Ground Truth**: 35.28

**ViDoRAG**:
Seeker:Round 1: Selects images [**8**, **10**]. Extracts data: 2024 and 2025 construction site reopening rates (39.9%, 27.5%) and Spring Festival service consumption growth rates (52.3%, 12.3%). Calculates initial components (2.48, 32) with a combined score of 34.48.
Inspector:Round 1: First calculation uses construction rate 23.5% (2025) to get a combined score of 35.28.
Round 2: Reverts to construction rate 27.5% (2025) and recalculates, resulting in a combined score of 34.48.
**Final Answer**: 34.48 ✗

**ColQwen2.5**: **Retrieved Pages**: [ 1✔, 7✔, 8✔, 2, 10✔, 9, 11, 6, 4, 5]. Successfully retrieved all ground truth pages within its Top-10 results.
**Python Solution**:
```python
def solution():
    construction_2025 = 23.5
    construction_2024 = 39.9
    ...
    service_2025 = 12.3
    service_2024 = 52.3
    ...
    combined_score = construction_impact + service_impact
    # Round to two decimal places
    return round(combined_score, 2)
```
**Final Answer**: 35.28 ✔

**Analysis**: In the final step of ViDoRAG, the synthesizer hallucinates. The construction site resumption rate in 2025 is 23.5% (construction_rate_2025 = 23.5), serving as the basic data for calculation. However, there are two data reference errors in the process. The "seeker" made a search and mistakenly wrote the 2025 construction site resumption rate as 27.5%. When calculating the "change in resumption rate from 2024 to 2025", it erroneously used "39.9% - 27.5% = 12.4 percentage points" (correctly, it should be "39.9% - 23.5% = 16.4 percentage points"). The "inspector" conducted two analyses: the first used the correct 2025 construction site resumption rate of 23.5% and finally calculated 35.28; the second followed the seeker's error, miswriting the 2025 rate as 27.5%, repeating the data error and causing an incorrect calculation of the construction industry contribution value (wrong value 2.48, correct one 3.28). Eventually, the synthesizer selected 34.28.

> # Case: test-47
>
> **Question**: Consider a hypothetical scenario where the custodial company Yingshan Steel is fully consolidated into Shandong Steel's financial statements effective at the beginning of 2026. Assume that Yingshan Steel's total liabilities and total shareholders' equity...... Calculate the pro-forma Total Liabilities to Total Shareholders' Equity ratio for the consolidated entity based on the forecasted 2026 balance sheet provided in the report......
>
> **Evidence Pages**: [31 , 38]
> **Ground Truth**: 1.154

**ViDoRAG**:

Seeker: Round 1: Selects images [**31**, **1**]. Extracts data: 2023 revenue of Yingshan Steel (25.8 billion yuan) and 2023 operating revenue of Shandong Steel (102.2894 billion yuan). Notes other images lack necessary information.
Round 2: Finds no images ....
Inspector: Round 1: Points out missing key information, including Shandong Steel's 2023 total operating revenue, Yingshan Steel's 2023 total liabilities and shareholders' equity, and relevant details from Shandong Steel's 2026 forecasted balance sheet for ratio calculation.
Synthesizer: Round 1: Calculate the revenue ratio (25800 / 90475 = 0.285). Assumes proportional addition of Yingshan Steel's liabilities and equity based on this ratio, combines with hypothetical 2026 forecast data of Shandong Steel, and concludes the pro-forma Total Liabilities to Total Shareholders' Equity ratio as 1.567.
**Final Answer**: 1.567 ✘

**ColQwen2.5**: **Retrieved Pages**: [ **31**✔, 30, 4, 5, 6, 36, **38**✔, 31, 3, 4]. Successfully retrieved all ground truth pages within its Top-10 results.
**Python Solution**:

```
def solution():
    yingshan_revenue_2023 = 25800, shandong_revenue_2023 = 90475.1
    revenue_ratio = yingshan_revenue_2023 / shandong_revenue_2023
    shandong_liabilities_2023 = 36130, shandong_equity_2023 = 30621
    shandong_liabilities_2026 = 35072, shandong_equity_2026 = 30604
    # Calculate the pro-rata liabilities and equity to be added
    ...
    # Calculate the consolidated total liabilities and total equity
    ...
    return ratio_result = round(total_liabilities / total_equity, 3)
```

**Final Answer**: 1.154 ✔

**Analysis**: Colqwen explicitly extracts specific data actually present in the document, including Shandong Iron and Steel's debt (36130 million yuan) and equity (30621 million yuan) in 2023, as well as its predicted debt (35072 million yuan) and equity (30604 million yuan) in 2026. In contrast, ViDoRAG's seeker and inspector fail to identify or use such actual data; instead, during the synthesizer's summarization, they rely on unfounded assumptions, like assuming Shandong Iron and Steel's 2026 debt as 50000 million yuan and equity as 30000 million yuan (with no document support), and assuming Yinshan Steel's 2023 debt as 20000 million yuan and equity as 10000 million yuan.

### H.4.2 ViDoRAG Successes vs. ColQwen2.5 Failures

---

**Case: test-696**

[Table image showing Results of business segments and all other, with Residential Mortgage Banking column highlighted]

**Question**: Calculate the return on assets (ROA) for the "Residential Mortgage Banking" segment for the year 2020 (in percentage). Answer rounded to two decimal places.
**Evidence Pages**: [62]
**Ground Truth**: 3.66

---

**ViDoRAG**:
Seeker: Round 1: Selects images [80, 62]. Extracts information about assets and net income (before provision for income taxes) of the Residential Mortgage Banking segment for periods in 2020, noting the lack of full-year data.
Round 2: Selects image [63]. Extracts total assets (640,112 thousand) and net income (28,562 thousand) of the segment for the nine months ended September 30, 2020.
Round 3: Finds no images ...... Inspector: Round 1: Points out the need for net income after taxes for the entire year 2020 and average total assets (beginning and ending) of the segment.
Round 2: Reiterates the requirement for 2020 net income and average total assets of the segment, as the available data is insufficient.
Synthesizer:
Round 1: Uses the formula ROA = (Net Income / Total Assets) * 100. Takes the net income before provision for income taxes (23,419 thousand) and total assets (640,112 thousand) of the segment from the selected images, calculates the ROA, and rounds it to two decimal places, resulting in 3.66%.
**Final Answer**: 3.66 ✔

---

**ColQwen2.5**: **Retrieved Pages**: [80, 69, 39, 78, 87, 86, **62**✔, 61, 62, 8]. Successfully retrieved all ground truth pages within its Top-10 results.
**Python Solution**:
```
def solution():
    net_income = 23419 # Net income before income tax provision
    total_assets = 648112  # Total assets
    # Calculate Return on Assets (ROA) as a percentage
    roa = (net_income / total_assets) * 100
    return roa_rounded = round(roa, 2)
```
**Final Answer**: 3.61 ✘

---

**Analysis**: In the correct solution, the total asset data used is 640,112 thousand US dollars, which is the accurate total asset value of the "Residential Mortgage Banking" department in the document. However, in the error process, colqwen mistakenly adopted 648,112 thousand US dollars as the total asset, inconsistent with the accurate value (640,112 thousand US dollars) in the document. This discrepancy directly caused the calculation base for ROA to be overly large, making the final result lower (as the denominator is bigger).

> **Case: test-921**
>
> [Financial statement image: AVALON GLOBOCARE CORP. AND SUBSIDIARIES — CONDENSED CONSOLIDATED STATEMENTS OF CHANGES IN EQUITY, For the Three Months Ended March 31, 2020 (Unaudited)]
>
> **Question**: What is the percentage increase in the net loss the company experienced from January 1, 2020 to March 31, 2020? Answer as a percentage. Answer rounded to two decimal places.
> **Evidence Pages**: [4 , 6 , 11]
> **Ground Truth**: 11.14

**ViDoRAG**:
Seeker: Round 1: Selects images [**6**, **33**]. Extracts key data: accumulated deficit as of December 31, 2019 (29,361,937 yuan); accumulated deficit as of March 31, 2020 (32,632,718 yuan); net loss for the three months ended March 31, 2020 (3,270,781 yuan)......
Round 2: Selects image [**5**]. Extracts key data: net loss for the three months ended March 31, 2020 (-3,270,781 yuan); net loss for the three months ended March 31, 2019 (-4,504,929 yuan)......
Round 3: Selects no images......
Inspector: Round 1: Points out the need for net loss data for January 1, 2020.......
Round 2: Points out the need for the net loss value of the company on January 1, 2020......
Synthesizer: Round 1: Calculates step-by-step:
1. Net loss for the period = ending accumulated deficit - beginning accumulated deficit = 32,632,718 - 29,361,937 = 3,270,781 yuan.
2. Percentage increase = (net loss / beginning accumulated deficit) × 100% = (3,270,781 / 29,361,937) × 100% = 11.14%.
**Final Answer**: 11.14 ✔

**ColQwen2.5**: **Retrieved Pages**: [ **6**✔, 16, 33, 7, 20, 34, 35, 5, 8, **11**✔]. Unsuccessfully retrieved all ground truth pages within its Top-10 results.
**Python Solution**:
```
def solution():
    # Net loss for the three months ended March 31, 2020 (current period)
    net_loss_current = 3270781
    # Net loss for the three months ended March 31, 2019 (comparative period)
    net_loss_comparative = 4504929
    # Calculate ...
    loss_change = net_loss_current - net_loss_comparative
    percentage_change = (loss_change / net_loss_comparative) * 100
    # Return the result rounded to two decimal places
    return round(percentage_change, 2)
```
**Final Answer**: 27.40 ✘

**Analysis**: Colqwen's error shows that there are two main issues in this wrong process. Firstly, there is a misunderstanding of the comparison benchmark. It incorrectly takes the net loss of the same period in 2019 as the benchmark, instead of the beginning accumulated deficit on January 1, 2020, which violates the time range limitation of "from January 1, 2020 to March 31, 2020" in the question. Secondly, there is a misunderstanding of the concept of "net loss increase". It calculates the change ratio of net losses in the same period of different years, rather than the growth rate of net loss relative to the beginning of the period within that period in 2020, which is completely inconsistent with the requirements of the question.

### H.4.3 ViDoRAG Failures vs. ColQwen2.5 Failures

---

**Case: test-23**

**Question**: Consider a hypothetical portfolio representing the combined assets of the top 40 performing Fixed Income Plus funds mentioned in the report analysis section......Suppose the portfolio manager decides to realign the proportions of the different types of convertible bonds within their existing total convertible bond allocation to precisely match the overall market's convertible bond type proportions as reported at the end of Q4 2024. Calculate the resulting absolute change in the percentage allocation to balanced-type convertible bonds, expressed as a percentage of the total portfolio assets......

**Evidence Pages**: [1, 8, 9, 13]
**Ground Truth**: 0.445

---

**ViDoRAG**:

Seeker: Round 1: Selects images [2, 14]. Extracted data: 2022 foreign value-added in U.S. exports: Electrical Equipment 24.3%, Refined Petroleum 29.9%; 2024 China R&D spending 3.6 trillion yuan (2.68% GDP). Round 2: Extracted 0 images. Analyzed China-U.S. trade and tariff impacts through global value chain perspectives.

Inspector: Round 1: Identified missing GDP and R&D data, assessed allocation for sector dependencies.

Synthesizer: Round 1: Calculated 2025 GDP, R&D spending increase, and allocated funds based on foreign value-added percentages.

**Final Answer**: 0.716 ✗

---

**ColQwen2.5**: **Retrieved Pages**: [8✓,3,15,7,14,6,1✓,13✓,9✓,12]. Successfully retrieved all ground truth pages within its Top-10 results.

**Python Solution**:

```
def solution():
    convertible_bond_allocation = 0.247
    original_balanced_ratio = 0.442
    # Market proportion of balanced convertible bonds
    market_balanced_ratio = original_balanced_ratio - 0.029
    # Calculate ...
    change = convertible_bond_allocation *
    (market_balanced_ratio - original_balanced_ratio)
    absolute_change = abs(change)
    # Return the result rounded to three decimal places
    return round(absolute_change, 3)
```

**Final Answer**: 0.007 ✗

---

**Analysis**: Colqwen made an error in extracting the data (the proportion of balanced convertible bonds in the overall market) and failed to extract 0.46. ViDoRAG also failed to extract 0.46. (Actually, ViDoRAG's seeker mentioned twice that "Image 6 shows that the proportion of market - balanced type is 46%", which is consistent with the correct process. However, the subsequent synthesizer made a summary error and set the market proportion as 41.3%, resulting in a final result error.) Meanwhile, ViDoRAG's reasoning process is closer to the reasoning process of the answer. In contrast, Colqwen did not convert it into a percentage at the end and got 0.007.

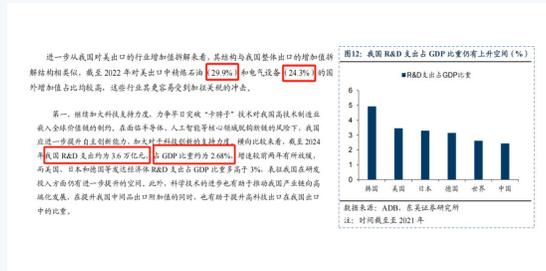

Case: test-26

**Question**: Based on the report's findings regarding China's industrial vulnerability in the context of U.S. trade relations...... Suppose China's GDP grows by 5.0% from the 2024 level implied in the report, and its R&D expenditure as a percentage of GDP increases by 0.15 percentage points compared to the 2024 level. Furthermore, assume 25% of the increase in absolute R&D spending from the implied 2024 level to the projected 2025 level ...... Calculate the amount of this specific R&D fund allocated to the Electrical Equipment sector in 2025......
**Evidence Pages**: [14]
**Ground Truth**: 43.89

**ViDoRAG**:
Seeker: Round 1: Selects images [**14**, **2**]. Extracts key data: foreign value-added percentages in exports to the U.S. for 2022 (Electrical Equipment: 24.3%, Refined Petroleum: 29.9%); 2024 R&D expenditure (3.6 trillion yuan) and R&D as a percentage of GDP (2.68%)......
Round 2: Selects no images......
Inspector: Round 1: Points out the need for 2024 GDP value and 2024 R&D expenditure as a percentage of GDP, which are essential to compute the increase in R&D spending and subsequent fund allocation.
Round 2: Status is "synthesizer" ......
Synthesizer: Round 1: Calculates: the 2025 R&D expenditure is 3.992 trillion yuan, an increase of 0.392 trillion yuan (392 billion yuan) compared with 2024. 25% of the increased R&D expenditure is set aside as a specific fund (98 billion yuan), of which the electrical equipment sector accounts for 44.83% and receives 43.79 billion yuan.
**Final Answer**: 43.79 ✗

**ColQwen2.5**: **Retrieved Pages**: [**14,**✓, 2, 1, 4, 13, 12, 10, 5, 6, 8]. Successfully retrieved all ground truth pages within its Top-10 results.
**Python Solution**:
```
def solution():
    rnd_expenditure_2024 = 36000
    rnd_gdp_ratio_2024 = 0.0268
    gdp_growth_rate = 0.05
    rnd_gdp_ratio_increase = 0.0015
    # Calculate ...
    electrical_allocation_ratio = electrical_equipment_ratio / total_ratio
    electrical_fund = allocated_fund * electrical_allocation_ratio
    return round(electrical_fund, 2)
```
**Final Answer**: 438.89 ✗

**Analysis**: The answer of ViDoRAG differs from the correct answer by 0.1, which is caused by errors in the intermediate calculation process. (Actually, starting from the calculation of GDP in 2024, the calculation of GDP in 2025, and the calculation of R&D expenditure in 2025, the "percentage point adjustment of the value - added ratio" is directly equated to the "change in economic contribution", resulting in deviations in each subsequent calculation step and causing a deviation in the final result.) The colqwen differs from the answer by a factor of 10. (Actually, it is an error in retaining the unit in the last step; it should not be in hundred million yuan but in ten - billion yuan.)

## H.5 VRAG-RL vs. ColQwen2.5

**Summary of Analyzed Examples: VRAG-RL vs. ColQwen2.5** The following summaries correspond to the cases detailed, which compare the performance of VRAG-RL and ColQwen2.5. The cases are categorized into groups to illustrate distinct performance dynamics based on their processing modes.

**Part I: ColQwen2.5 Successes vs. VRAG-RL Failures**

1. **Failure Mode: Misunderstanding of Financial Ratio Definition (Case: test-541)** VRAG-RL correctly retrieved and adjusted cash and inventory values but failed due to a fundamental misunderstanding of the Current Ratio formula. It erroneously used the ratio of revised cash to revised inventory instead of the standard calculation (total current assets divided by current liabilities). ColQwen2.5 accurately applied the definition of Current Ratio by calculating revised current assets (summing adjusted cash, adjusted inventory, and unchanged other current assets) and then dividing by total current liabilities, ensuring alignment with the ground truth. This highlights the importance of correctly interpreting financial ratios.

**Part II: VRAG-RL Successes vs. ColQwen2.5 Failures**

2. **Success Mode: Accurate Parameter Extraction and Logic Application (Case: test-475)** VRAG-RL correctly outlined the core calculation logic, accurately linked strategic shift parameters to the calculation framework, and integrated the overall 2026 gross margin for other segments, aligning with problem constraints. ColQwen2.5 failed due to a critical incorrect assumption of the original 2026 overall gross margin, invalidating the final result despite correct retrieval of relevant pages and calculation structure. This shows the importance of accurate extraction of baseline metrics.

3. **Success Mode: Adherence to Problem Definition (Case: test-614)** VRAG-RL correctly focused on IPO gross proceeds as the calculation base, excluding underwriters' over-allotment proceeds. It accurately applied the 1% deposit requirement per 3-month extension and multiplied by two extensions to reach the correct result. ColQwen2.5 made a critical error by including underwriters' over-allotment proceeds in the total gross proceeds, inflating the calculation base. This emphasizes the need to strictly adhere to problem definitions of financial metrics.

**Part III: ColQwen2.5 Failures vs. VRAG-RL Failures**

4. **Shared Failure: Errors in Transaction Cost and Data Handling (Case: test-170)** Both models failed due to distinct errors. VRAG-RL incorrectly applied transaction costs once per year instead of monthly, underestimating annual costs and inflating adjusted returns. ColQwen2.5 used an incorrect 2021 excess return and flawed cost logic (deducting total three-year costs from the average return instead of annual costs from each year's return), leading to an understated result. This highlights deficiencies in transaction cost calculation and data extraction for both models.

## H.5.1 VRAG-RL Failures vs. ColQwen2.5 Successes

---

**Case: test-541**

**Question**: Consider the company's balance sheet and operational forecasts for 2027. Suppose a strategic inventory management overhaul is implemented. This initiative aims to reduce the year-end 2027 Inventory balance to be only 150% of the year-end 2025 Inventory balance (using the ending balances forecasted in the report). Concurrently, assume that the company manages to improve its collections process such that the year-end 2027 Monetary Funds (Cash) balance increases by 20% compared to the original 2027 forecast. All other asset and liability accounts for 2027 remain exactly as projected in the report's financial statements. Based on these adjustments, calculate the company's revised Current Ratio for the end of fiscal year 2027. (Round to two decimal places, unitless).
textbfEvidence Pages: [22]
textbfGround Truth: 2.18

---

**VRAG-RL**: **Reasoning**: **Step 1**: Correctly calculated the revised 2027 inventory (150% of 2025 inventory: 6,436 × 1.5 = 9,654 million yuan) and revised 2027 cash (20% increase on original forecast: 30.67 × 1.2 = 36.804 million yuan). **Step 2**: Critical error in defining the Current Ratio: incorrectly used "revised cash ÷ revised inventory" instead of "total current assets ÷ current liabilities". **Final Answer**: 0.38 ✘

---

**ColQwen**: **Retrieved Pages**: [22✔,19,20,2,1,3,17,8,13,7]. **Python Solution**:

```
def solution():
#Define key variables (2027 forecast data, in million yuan)
original_cash = 4392
original_inventory = 2812
other_current_assets = 18910 - original_cash - original_inventory
#Calculate revised cash (20% increase) and inventory (150% of 2025)
revised_cash = original_cash * 1.2
revised_inventory = inventory_2025 * 1.5
#Calculate total revised current assets
revised_current_assets = revised_cash + revised_inventory + other_current_assets
Correct Current Ratio: current assets ÷ current liabilities
current_ratio = revised_current_assets / current_liabilities
return round(current_ratio, 2) # Correct result: 2.18
```

**Final Answer**: 2.18 ✔

---

**Analysis**: VRAG-RL correctly retrieved and adjusted the cash and inventory values but failed due to a fundamental misunderstanding of the Current Ratio formula. It erroneously used the ratio of revised cash to revised inventory instead of the standard calculation (total current assets divided by current liabilities). ColQwen accurately applied the definition of Current Ratio by first calculating revised current assets (summing adjusted cash, adjusted inventory, and unchanged other current assets) and then dividing by total current liabilities. This adherence to the financial metric's definition ensured alignment with the ground truth. The discrepancy highlights the importance of correctly interpreting financial ratios and their underlying components in quantitative analysis.

## H.5.2 VRAG-RL Successes vs. ColQwen2.5 Failures

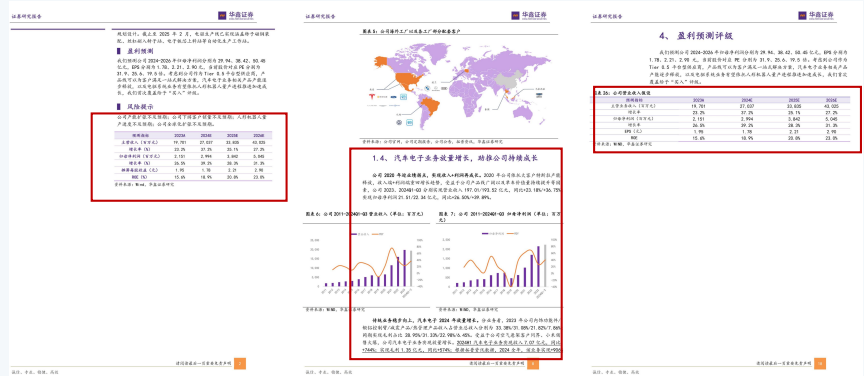

**Case: test-475**

**Question**: Assume Tuopu Group's projected 2026 performance unfolds, but with a strategic shift: the Automotive Automotive Electronics Electronics segment significantly outperforms, capturing 15% of the total forecasted revenue for that year. Furthermore, assume due to process optimization and economies of scale, this segment achieves a gross margin 5.0 percentage points higher than its reported gross margin for 2024H1. If all other business segments collectively account for the remaining revenue and maintain the overall gross margin percentage originally forecasted for the company in 2026, calculate the company's revised total consolidated gross profit for 2026 (round to two decimal places, unit: 100 million yuan).
**Evidence Pages**: [2, 8, 9, 18, 20]
**Ground Truth**: 100.03

**VRAG-RL**: **Reasoning**: **Step 1**: Correctly outlined the core calculation logic: determining 2024H1 gross margin of Automotive Electronics, adjusting for 2026, deriving segment revenues, and summing gross profits. **Step 2**: Accurately linked the strategic shift parameters (15% revenue share, 5pp margin increase) to the calculation framework. **Step 3**: Successfully integrated the overall 2026 gross margin (9.52%) for other segments, aligning with the problem constraints. **Final Answer**: 100 ✓

**ColQwen**: **Retrieved Pages**: [18✓,2✓, 8✓,3,9✓,1,10,5,17,12].
**Python Solution**:

```
def solution():
    # Define basic variables (omitted: revenue_2026, 2024H1 revenue/profit data)
    # Calculate 2024H1 gross margin for Automotive Electronics (correct)
    # Calculate optimized 2026 margin for Automotive Electronics (correct)
    # INCORRECT PARAMETER: Original 2026 overall gross margin
    # Should be 9.52% (from context) but incorrectly assumed as 22%
    original_total_gross_margin = 0.22  # Critical error here
    # Calculate segment revenues and gross profits (omitted: correct formulas)
    # Convert units and round result (omitted: correct process)
    return 96.01  # Incorrect due to above error
```

**Final Answer**: 96.01 ✗

**Analysis**: VRAG-RL (acc=1) correctly structured the reasoning framework by identifying key parameters and their relationships, aligning with the problem constraints despite simplified output. ColQwen (acc=0) failed due to a critical incorrect assumption of the original 2026 overall gross margin (22% vs. actual 9.52% from context). This single parameter error invalidated the final result, even though the retrieval of relevant pages and calculation structure were correct. The discrepancy highlights the importance of accurate extraction of baseline metrics from source materials in financial forecasting scenarios.

> **Case: test-614**
>
> 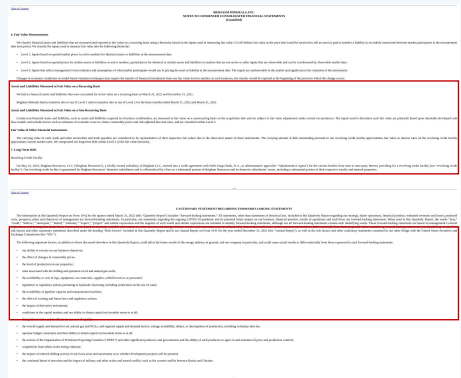
>
> **Question**: If the company decides to extend the business combination period two times by an additional three months each time, how much additional funds from the gross proceeds of the IPO should the Sponsor deposit into the trust account in USD, including the proceeds from the exercise of the underwriters' over-allotment option? Answer to the nearest integer.
> **Evidence Pages**: [5, 16]
> **Ground Truth**: 3000000

> **VRAG-RL**: **Reasoning**: **Step 1**: Correctly identified that each 3-month extension requires a deposit of 1% of the IPO gross proceeds ($150,000,000), which is $1,500,000 per extension. **Step 2**: Recognized that two extensions are needed, so the total additional deposit is calculated as 2 times the single extension amount. **Step 3**: Excluded underwriters' over-allotment proceeds from the calculation base, aligning with the problem's focus on IPO gross proceeds. **Final Answer**: 3000000 ✔

> **ColQwen**: **Retrieved Pages**: [18,74,6,63,64,73,81,97,20,80]. **Python Solution**:
>
> ```
> def solution():
> Define basic variables (IPO gross proceeds: $150,000,000; 1% deposit per extension)
> INCORRECT INCLUSION: Added underwriters' over-allotment proceeds ($22,500,000)
> total_gross_proceeds = 150000000 + 22500000 # Should only use $150,000,000
>
> extension_times = 2
> deposit_per_extension = 0.01
> Calculation affected by incorrect total proceeds
> answer = total_gross_proceeds * deposit_per_extension * extension_times
>
> return 3450000 # Incorrect due to over-allotment inclusion
> ```
> **Final Answer**: 3450000 ✗

> **Analysis**: VRAG-RL correctly focused on the IPO gross proceeds as the calculation base, excluding underwriters' over-allotment proceeds. It accurately applied the 1% deposit requirement per 3-month extension and multiplied by two extensions to reach the correct result.
> ColQwen retrieved relevant pages but made a critical error by including underwriters' over-allotment proceeds in the total gross proceeds. This unnecessary addition inflated the calculation base, leading to an overestimated deposit amount. The discrepancy highlights the importance of strictly adhering to the problem's definition of "gross proceeds of the IPO" when identifying relevant financial metrics.

### H.5.3 VRAG-RL Failures vs. ColQwen2.5 Failures

> **Case: test-170**
>
> **Question**: Imagine an investor implemented the Composite Factor strategy exactly as described in the report (z-score equal weighted combination, top 5 industries after crowding filter, monthly rebalancing) starting from the beginning of 2019 through to the end of 2021. However, this investor incurred a transaction cost of 5 basis points (0.05%) on the portfolio value at each monthly rebalancing event. Assuming the benchmark incurs no transaction costs, calculate the average annual excess return achieved by this investor over the specified three-year period (2019, 2020, 2021 inclusive), after accounting for these transaction costs. Provide the answer as a percentage value, rounded to two decimal places.
> **textbfEvidence Pages**: [4, 16]
> **textbfGround Truth**: 13.34

**VRAG-RL**: **Retrieved Pages**: [Not explicitly specified]. **Reasoning**: **Step 1**: Correctly identified annual excess returns (2019:12.57%, 2020:4.95%, 2021:24.30%). **Step 2**: Critical error in transaction cost calculation: assumed annual cost = 0.05% (1 monthly rebalancing), but actual annual cost = 0.05%×12=0.6% (12 monthly rebalancings). **Step 3**: Adjusted each year's return by only 0.05%, resulting in inflated annual returns and an overestimated average (13.94% vs. correct 13.34%). **Final Answer**: 13.94 ✗

**ColQwen**: **Retrieved Pages**: [15,16✓,10,13,1,9,7,6,2,3]. **Python Solution**:

```
def solution():
INCORRECT DATA: 2021 excess return should be 24.30% but used 23.30%
annual_excess_returns = [12.57, 4.95, 23.30]
transaction_cost = 0.05 # 5 basis points per month
num_months_per_year = 12
num_years = 3
INCORRECT COST LOGIC: Total cost deducted from average return instead of annual
total_transaction_cost = transaction_cost * num_months_per_year * num_years
average_before_cost = sum(annual_excess_returns) / num_years
average_after_cost = average_before_cost - total_transaction_cost # Flawed deduction
return round(average_after_cost, 2) # Incorrect result: 11.81
```

**Final Answer**: 11.81 ✗

**Analysis**: Both models failed due to distinct errors in transaction cost calculation and data extraction. VRAG-RL incorrectly applied transaction costs once per year instead of monthly, underestimating annual costs (0.05% vs. 0.6%) and inflating adjusted returns. ColQwen used an incorrect 2021 excess return (23.30% vs. 24.30%) and flawed cost logic (deducting total three-year costs from the average return instead of annual costs from each year's return), leading to an understated result.



\end{document}